\documentclass[runningheads]{llncs}
\usepackage{graphicx}

\usepackage{tikz}
\usepackage{comment}
\usepackage{amsmath,amssymb} 
\usepackage{color}

\usepackage[width=122mm,left=12mm,paperwidth=146mm,height=193mm,top=12mm,paperheight=217mm]{geometry}

\usepackage[misc]{ifsym} 
\definecolor{linkcolor}{RGB}{255,0,0}
\definecolor{urlcolor}{RGB}{255,105,180}
\definecolor{citecolor}{RGB}{66,168,235}
\usepackage{hyperref}
\hypersetup{colorlinks=true,linkcolor=linkcolor,urlcolor=urlcolor,citecolor=citecolor}
\usepackage{subcaption}
\newcommand{\concat}[0]{\mathbin\Vert}
\usepackage{graphicx}
\usepackage{amsmath}
\usepackage{amssymb}
\usepackage{booktabs}
\usepackage{threeparttable}
\usepackage{subfloat}
\usepackage{tabularx}
\usepackage{adjustbox}
\usepackage{colortbl}

\newcommand{\polyphonic}{\includegraphics[height=0.8em]{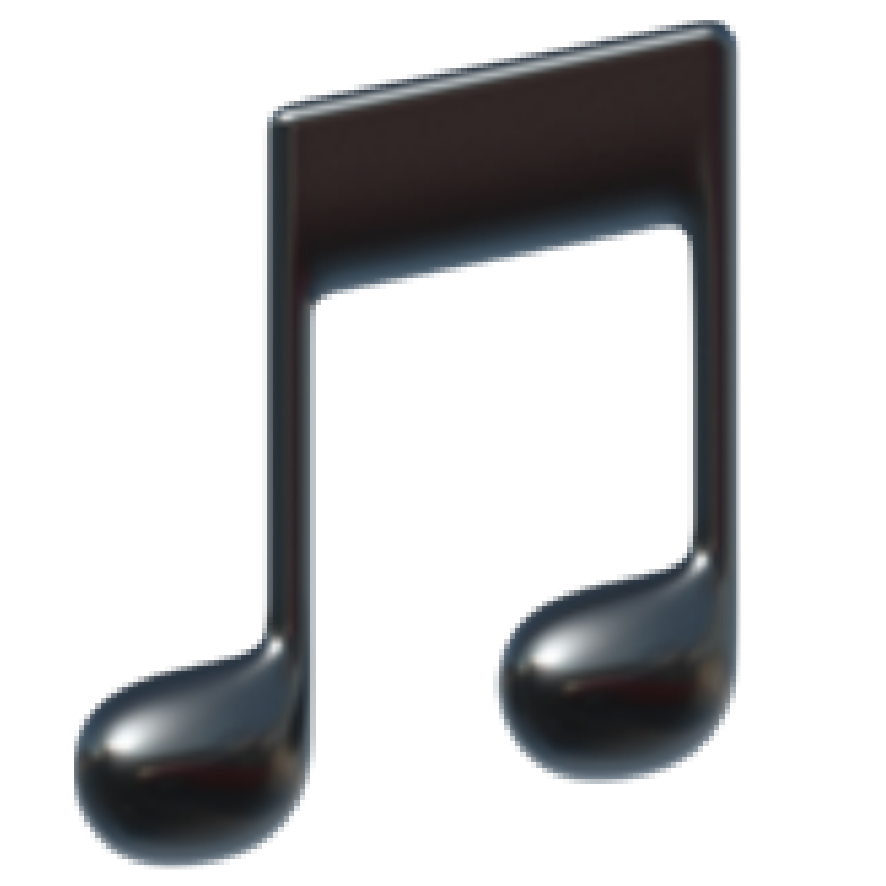}}

\begin{document}
\pagestyle{headings}
\mainmatter
\def\ECCVSubNumber{4533}  

\title{\polyphonic{}PolyphonicFormer: Unified Query Learning for Depth-aware Video Panoptic Segmentation}

\titlerunning{Abbreviated paper title}
%
\author{Haobo Yuan\inst{1\star} \and
Xiangtai Li\inst{3}\thanks{The first two authors contribute equally.} \and
Yibo Yang\inst{4\star\star} \and
Guangliang Cheng\inst{5} \and
Jing Zhang\inst{6} \and
Yunhai Tong\inst{3} \and
Lefei Zhang\inst{1,2}\thanks{Corresponding Author. Email: zhanglefei@whu.edu.cn, ibo@pku.edu.cn.} \and
Dacheng Tao\inst{4}
}

\authorrunning{H. Yuan and X. Li et al.}
\titlerunning{\polyphonic{}PolyphonicFormer}
%
\institute{Institute of Artificial Intelligence and School of Computer Science, Wuhan~University, Wuhan, 430072, P. R. China \and
Hubei Luojia Laboratory, Wuhan, 430072, P. R. China \and
Key Laboratory of Machine Perception, MOE, School of Artificial Intelligence, Peking University, Beijing, 100871,  P. R. China \and
JD Explore Academy, Beijing, P. R. China \and
SenseTime Research, Beijing, P. R. China \and
The University of Sydney, Sydney, Australia\\
\email{yuanhaobo@whu.edu.cn, lxtpku@pku.edu.cn}
}
\maketitle

\begin{abstract}
The Depth-aware Video Panoptic Segmentation (DVPS) is a new challenging vision problem that aims to predict panoptic segmentation and depth in a video simultaneously. The previous work solves this task by extending the existing panoptic segmentation method with an extra dense depth prediction and instance tracking head. However, the relationship between the depth and panoptic segmentation is not well explored -- simply combining existing methods leads to competition and needs carefully weight balancing. In this paper, we present PolyphonicFormer, a vision transformer to unify these sub-tasks under the DVPS task and lead to more robust results. Our principal insight is that the depth can be harmonized with the panoptic segmentation with our proposed new paradigm of predicting instance level depth maps with object queries. Then the relationship between the two tasks via query-based learning is explored. From the experiments, we demonstrate the benefits of our design from both depth estimation and panoptic segmentation aspects. Since each thing query also encodes the instance-wise information, it is natural to perform tracking directly with appearance learning. Our method achieves state-of-the-art results on two DVPS datasets (Semantic KITTI, Cityscapes), and ranks \textbf{1st} on the ICCV-2021 BMTT Challenge video + depth track. Code is available \href{https://github.com/HarborYuan/PolyphonicFormer}{here}.
\end{abstract}

\section{Introduction}
To enable machines to perceive the 3D world from 2D video clips, Depth-aware Video Panoptic Segmentation (DVPS)~\cite{ViPDeepLab} is proposed by extending the video panoptic segmentation~\cite{kim2020vps,STEP} with monocular depth estimation~\cite{saxena2005learning} which is a challenging and not trivial task for scene understanding. It predicts temporally consistent instance-level semantic results along with per-pixel depth prediction. Achieving accurate and robust DVPS methods in real-world scenarios can greatly promote the development of autonomous driving~\cite{geiger2012we,cordts2016cityscapes}. 

\begin{figure}[t!]
    \centering
    \includegraphics[width=0.8\textwidth]{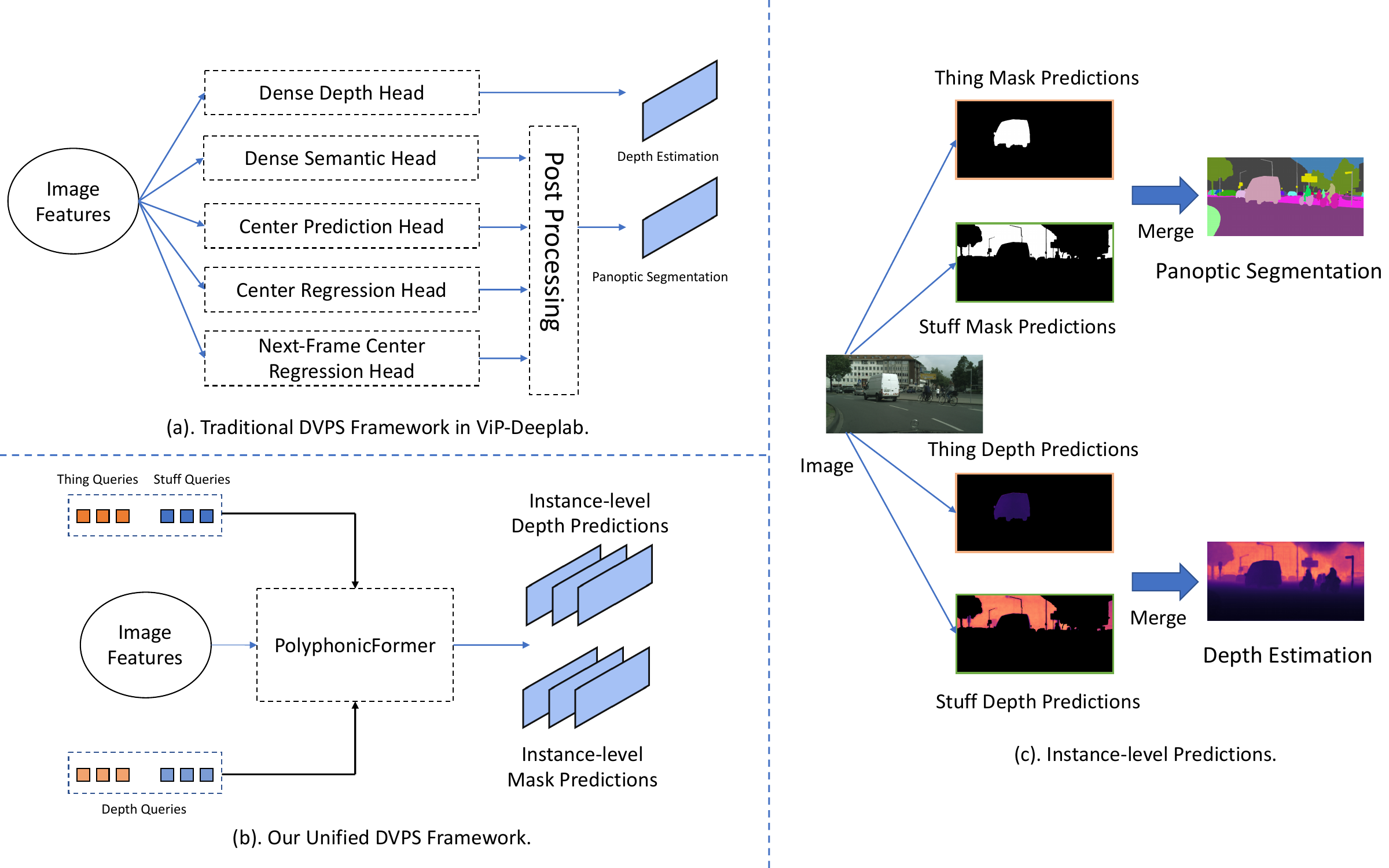}
    \caption{\small The principal idea of PolyphonicFormer. (a) Previous work's pipeline~\cite{ViPDeepLab}. (b), Illustration of our unified framework. (c). Key insights of our PolyphonicFormer: The queries are extracted from instance-level information and used to predict instance-level mask and depth results in a unified manner. Best view it on screen. }
    \label{fig:cover}
\end{figure}

To tackle this problem, previous work~\cite{ViPDeepLab} builds two DVPS datasets, including Cityscapes-DVPS and SemKITTI-DVPS by extending two existing datasets Cityscape-VPS~\cite{kim2020vps} and Semantic-KITTI~\cite{semantic_kitti}. It also offers a strong baseline ViP-Deeplab~\cite{ViPDeepLab} by adding two separate task heads based on Panoptic-Deeplab~\cite{cheng2020panoptic} including one dense depth estimation prediction head and one center prediction head for offline tracking as shown in Figure~\ref{fig:cover}(a).
However, the separate-head design will damage the performance on both the depth estimation and panoptic segmentation since both fight for suitable backbone features and conflict with each other as observed in previous work~\cite{ViPDeepLab}. Besides, the model is sensitive to the loss weights of depth estimation and panoptic segmentation, and thus requires a careful balance between the two tasks to enable satisfying performances for both. Therefore, a unified framework that stably facilitates the mutual benefit of the two tasks remains to be developed.

Recently, transformers~\cite{VIT,liu2021swin} make great progress on vision tasks. Except for the global perception property of vision transformers, the query-based set prediction design in Detection Transformer (DETR)~\cite{detr} has been proved to be effective in modeling the interactions between instance-level and global contexts. Following the DETR formulation, the query-based design has been applied with good performances on various tasks~\cite{QueryInst,zhang2021knet,cheng2021maskformer,wang2020maxDeeplab}. In our task, the instance-level semantic contexts have a strong connection with depth information, since the boundary of an instance in the image corresponds to the region where sharp changes occur in the depth map. So it is natural to utilize query-based reasoning to fulfill the mutual benefit between semantic context and geometry information.

Motivated by such analysis, in this paper, we develop a novel unified query-based representation for things, stuff, and depth, and propose \textbf{PolyphonicFormer} (Polyphonic Transformer) where different queries come from different sources (depth or panoptic) but can benefit each other. This is just like the \textbf{polyphony} used in the music field. As shown in Figure~\ref{fig:cover}(b), the PolyphonicFormer builds a unified query learning framework to enable the interaction between panoptic context and depth information and make them benefit from each other iteratively. We further link both panoptic and depth queries to facilitate mutual benefit. Our insight is that both depth prediction and panoptic prediction can be linked via the corresponding \textit{instance-wised masks and queries}, which are shown in Figure~\ref{fig:cover}(c), since one unique mask corresponds to one unique query. As a result, the joint modeling can inject semantic information from thing and stuff queries into depth prediction. Moreover, the query-based depth prediction can be seen as regularization items to force semantic consistency during training, which also improves the segmentation results. In terms of temporal modeling, our model directly links to the thing query, which avoids the complex offline post-processing that is used in ViP-Deeplab~\cite{ViPDeepLab}.

Adopting the query-based method can avoid noises and reduce the resource battle from two tasks because the query can be seen as a disentanglement from the feature. This makes our method more robust and less sensitive to the weight of loss for different tasks. Our experiment shows that our unified query learning can enhance both results of panoptic segmentation and depth estimation on a strong baseline. Besides, the experimental results show that the PolyphonicFormer has a robustness of weight choice for the losses in the depth and panoptic predictions. In general, our contributions can be summarized as follows:
\begin{itemize}
    \item We propose PolyphonicFormer, which unifies the sub-tasks of the depth-aware video panoptic segmentation. To the best of our knowledge, our work is the first unified and end-to-end model in this field. The unified design can leverage information from both the panoptic and depth features to promote the other, and thus makes PolyphonicFormer more stable and insensitive to the loss weights choices that need to be carefully tuned in the previous study.
    \item Based on the observation that the depth map of each instance has a relatively simple structure than the whole scene, we propose a novel query-based depth estimation scheme to predict the depth map of each instance and merge them into a complete depth map. The query-based depth estimation scheme takes full advantages of the instance boundaries and has better accuracy.
    \item We conduct extensive experiments on Cityscapes-DVPS, Cityscapes-VPS, and SemKITTI-DVPS to verify the effectiveness of PolyphonicFormer. Our method achieves 64.6 DSTQ on the SemKITTI-DVPS Challenge and won the ICCV-2021 BMTT Challenge video + depth track. Ablation studies show that our framework can boost the \textit{mutual benefit} between depth and panoptic predictions as well as has \textit{robustness on weight choice for losses}.
\end{itemize}

\section{Related Work}
\label{sec:relatedwork}

\noindent
\textbf{Panoptic Segmentation.}
The earlier work~\cite{kirillov2019panoptic} directly combines predictions of things and stuff from different models. To alleviate computation cost, many works~\cite{kirillov2019panopticfpn,li2019attention,chen2020banet,li2018learning,porzi2019seamless,yang2019sognet,Wu2020AutoPanopticCM} are proposed to model both stuff segmentation and thing segmentation in one model with different task heads. Detection based methods~\cite{xiong2019upsnet,kirillov2019panopticfpn,qiao2021detectors,hou2020real,li2020panopticFCN} usually represent thing within the box prediction, while several bottom-up models~\cite{cheng2020panoptic,yang2019deeperlab,gao2019ssap,axialDeeplab} perform grouping instance via the pixel-level affinities or instance centers from semantic segmentation results. Recently, several works~\cite{detr,wang2020maxDeeplab,zhang2021knet,cheng2021maskformer} use transformer-like architecture and use object queries to encode thing and stuff masks in an end-to-end manner. Our method is inspired by these works and takes a further step by jointly modeling depth estimation into panoptic segmentation to build a unified framework.

\noindent
\textbf{Joint Depth and Semantic Modeling.}
Several works~\cite{chen2019towards,ochs2019sdnet,ramirez2018geometry,saeedan2021boosting} model the depth estimation and semantic (panoptic) segmentation as a naive multi-task learning procedure and bond them with a series of consistent losses. Some other works~\cite{guizilini2019semantically,li2020semantic,wang2020sdc} utilize semantic segmentation to guide depth estimation learning. From the perspective of 3D geometry, previous works~\cite{klingner2020self,lee2021learning,casser2019depth} use semantic or instance segmentation to tackle the object motion problem left in the self-supervised monocular depth estimation~\cite{godard2019digging}. Recently, ViP-Deeplab~\cite{ViPDeepLab} extends Panoptic-Deeplab~\cite{cheng2020panoptic} by adding extra dense depth prediction heads, and builds a holistic perception framework. Different from the previous works, where depth and segmentation are tackled separately, we link both parts via object queries in a unified manner, so enjoy the mutual benefit of both tasks.

\noindent
\textbf{Segmentation and Tracking in Video.} 
Several tasks including Video Object Segmentation (VOS)~\cite{VOS_data,voigtlaender2019feelvos,stm_vos}, Video Instance Segmentation (VIS)~\cite{vis_dataset},  Multi-Object Tracking and Segmentation (MOTS)~\cite{voigtlaender2019mots}, and Video Panoptic Segmentation (VPS)~\cite{kim2020vps} are very close to DVPS. Previous works~\cite{kim2020vps,mask_pro_vis,DFF,woo2021learning_associate_vps,video_knet} mainly model the temporal consistency for these tasks. VisTR~\cite{VIS_TR} solves VIS problem using transformer~\cite{vaswani2017attention} in an end-to-end manner. However, this method is limited by the complex scene and long video clip inputs. Recently, STEP~\cite{STEP} proposes a new metric named Segmentation Tracking Quality (STQ) for better evaluating video panoptic segmentation. There are also several tracking methods~\cite{transtrack,tracking_as_points,pang2021quasi,JDE,wangUnitrack} for MOT in terms of transformer architecture and appearance learning. Our PolyphonicFormer simply applies an appearance-based method, Quasi-Dense~\cite{pang2021quasi} to extract tracking embeddings as the tracking head.

\noindent
\textbf{Vision Transformer.} 
Many researchers have adopted Transformer~\cite{vaswani2017attention} to solve computer vision problems~\cite{axialDeeplab,VIT,deit_vit,liu2021swin,xu2021vitae,zhang2022vitaev2,wang2022towards,wang2021fp}. With the help of vision transformers, DETR~\cite{detr} reformulates the object detection task to output set predictions with an end-to-end framework. The key component is the object query design, which serves as a learnable anchor to detect each instance.
This novelty has been also used in panoptic segmentation~\cite{wang2020maxDeeplab,zhang2021knet,fashionformer,cheng2021maskformer,li2021panoptic,panopticpartformer,zhou2022transvod,zhang2022eatformer} where the pure mask classification can work well on panoptic segmentation.
In this paper, inspired by DETR-like segmentation methods~\cite{detr,cheng2021maskformer}, we borrow the merits of query-based reasoning and into joint depth and panoptic prediction.

\begin{figure*}[t!]
	\centering
	\includegraphics[width=0.98\linewidth]{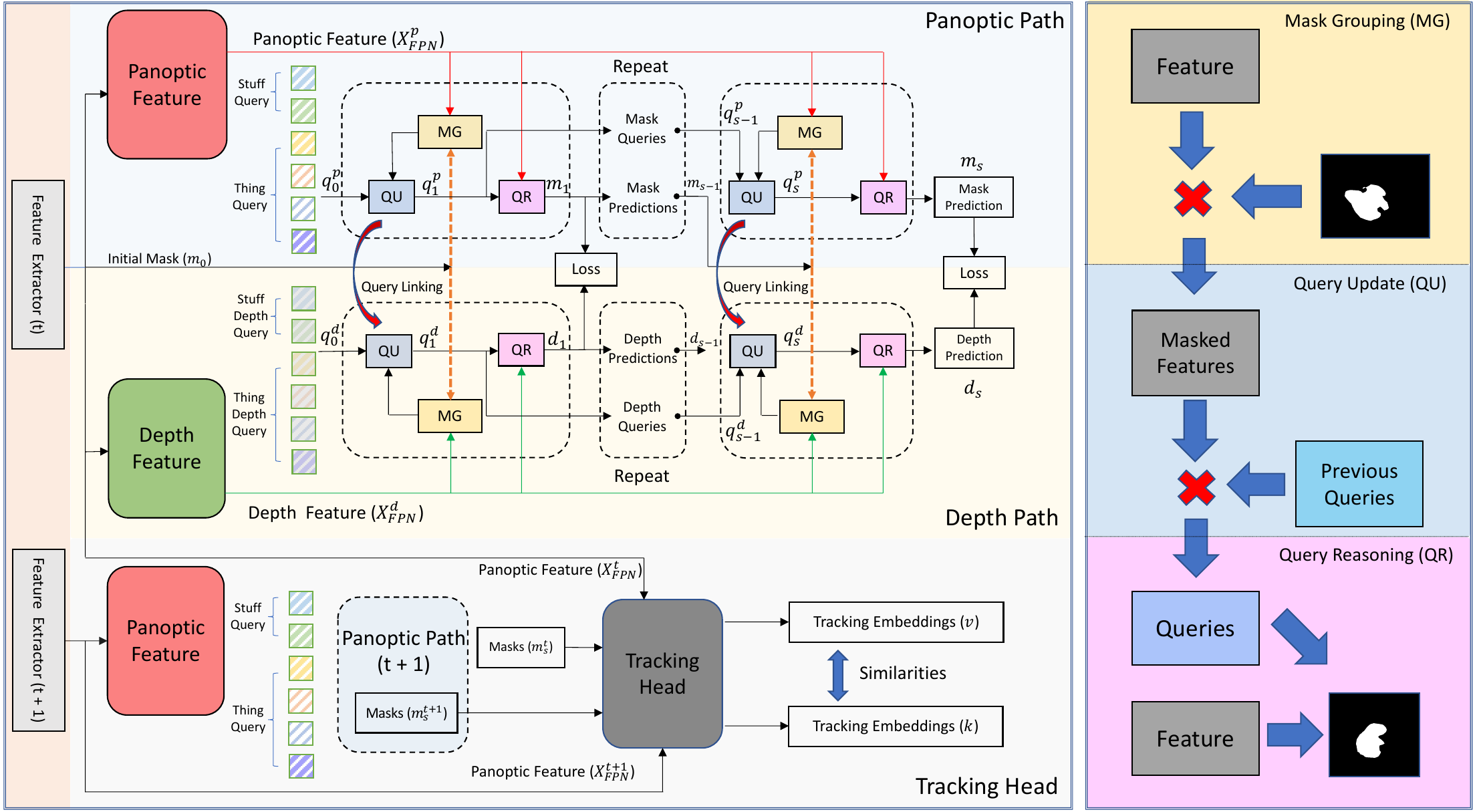}
	\caption{\small Illustration of our proposed PolyphonicFormer (left). Our method contains three parts: (1) Feature Extractor to obtain two parallel features for depth estimation and panoptic segmentation. The extracted panoptic (\textcolor[RGB]{208,1,5}{red}) and depth (\textcolor[RGB]{109,175,87}{green}) features are directly used to predict the final output. (2) Polyphonic Head containing the panoptic path and depth path to refine all the queries at the same time (middle). We use the instance masks to group (\textcolor[RGB]{211,124,58}{orange}) the panoptic and depth features for query-based learning. (3) Tracking Head to learn the feature embedding among the frames (bottom). The Mask Grouping (MG), Query Update (QU), and Query Reasoning (QR) (right). }
	\label{fig:methods}
\end{figure*}

\section{Method}
\label{sec:method}
We explain the architecture design of PolyphonicFormer in detail. As shown in Figure~\ref{fig:methods}, our PolyphonicFormer is composed of three parts: (1) A feature extractor to extract features for each frame. (2) a unified query learning sub-network, which takes three different types of queries and backbone features as inputs and outputs both panoptic segmentation and depth estimation predictions. It contains two paths: depth path and panoptic path, which are interacted with each other iteratively. (3) A lightweight tracking head with several convolution layers to learn the instance-wise tracking embeddings for each thing query.

\noindent
\subsection{Feature Extractor}
We use a shared backbone network (Residual Network~\cite{resnet} or Swin Transformer~\cite{liu2021swin}) along with Feature Pyramid Network as the feature extractor. We adopt the semantic FPN~\cite{kirillov2019panopticfpn} design to simply fuse the multiscale information since both the depth and panoptic predictions need the high-resolution feature to find the fine structures as well as high-level semantic information. In particular, leveraging a shared backbone, the neck generates different feature maps for panoptic and depth predictions in parallel. These feature maps output from FPN are denoted as $\boldsymbol{X}^{p}_{\mathtt{FPN}}$ and  $\boldsymbol{X}^{d}_{\mathtt{FPN}}$, respectively. They are directly used to generate the final and intermediate predictions of depth estimation and panoptic segmentation with the queries in each stage, as described in Section~\ref{sec:uni_query_learning}.

\noindent
\subsection{Unified Query Learning} \label{sec:uni_query_learning}
The unified query learning sub-network is composed of two parallel paths for panoptic and depth predictions, respectively. Each path takes the panoptic or depth feature as input and uses \textit{mask grouping} and \textit{query update} to iteratively update the queries. The updated queries are adopted to refine the corresponding mask and depth predictions by the \textit{query reasoning} module.

\noindent
\textbf{Mask Grouping.} At each stage $s$, we first use the mask results generated at the former stage $s-1$ to group the $\boldsymbol{X}^{p}_{\mathtt{FPN}}$ and $\boldsymbol{X}^{d}_{\mathtt{FPN}}$. The key motivation has two folds: The first is that using grouping can avoid noise and select the most salient part for query learning. The second is that adopting grouping or pooling can speed up the learning process of DETR-like models~\cite{peize2020sparse,zhang2021knet,video_knet}. Thus, we formulate the process as follows:
\begin{equation}
    \boldsymbol{X}_{s}^{(n, c)} = \sum_u^W\sum_v^H \boldsymbol{m}_{s-1}^n(u, v) \cdot \boldsymbol{X}^{c}_{\mathtt{FPN}}(u, v),
\end{equation}
where $\boldsymbol{m}_{s-1}$ is the mask predictions from former stage. The $\boldsymbol{X}^{p}_{\mathtt{FPN}}$ is with a $C \times H \times W$ shape, $\boldsymbol{m}$ is with a $N \times H \times W$ shape, if we use $C$ and $N$ to represent number of channels and instances respectively. In PolyphonicFormer, we set $C = 256$ and $N = 100$, respectively. For each mask, we generate a corresponding feature to represent the instance-level information. The instance-level features ($N \times C$) are used to update the queries better focusing on each instance. This process is done in both panoptic path and depth path, with the same former stage mask $\boldsymbol{m}_{s-1}$, as shown in Figure~\ref{fig:methods} with \textcolor[RGB]{211,124,58}{orange} line. With the help of previous stage masks, both the depth and panoptic predictions can be refined with instance-level features. 

For the first stage, since the former stage is not available, we use an initial stage (denoted as stage 0), to generate the initial depth and panotpic results. Specifically, we use different convolution module to generate dense predictions on instance segmentation, semantic segmentation, and depth estimation, which are all guided with the ground truth. The initial panoptic segmentation results are delivered to the first stage for mask grouping. The convolution kernels are also transferred to the first stage as the initial queries. Mask Grouping operation is conducted on both 
panoptic path and depth path separately.

\noindent
\textbf{Query Update.} 
The masked instance-level features are then delivered to update panoptic and depth queries. We describe the panoptic path first. Inspired by K-Net~\cite{zhang2021knet}, we adopt an adaptive query update process. The query update process avoids the heavy pixel-wised affinity calculation that is used in DETR-like models~\cite{cheng2021maskformer,detr}. The query update process forces the model to learn the instance-aware feature to speed up the training procedure. Specifically, we use a gated fusion design to capture the information from queries and features. We first calculate the gate features $F_G$ as follows:
\begin{equation}
		F_G^{p} = \phi(\boldsymbol{X}_{s}^{p}) \cdot \phi(q_{s-1}^{p}),
		\label{eq:qu1}
\end{equation}
where $\phi$ is linear transformations, and all of the $\phi$s in this section are independent. The $F_G$ has the information from both the queries and the instance-level features. With $F_G$, we can calculate gates for features and queries with the following formulation:
\begin{equation}
	\begin{aligned}
		{G}_{q}^{p} = \sigma(\psi(F_G^{p})), {G}^{p}_{\boldsymbol{X}} = \sigma(\psi(F_G^{p})),
		\label{eq:qu2}
	\end{aligned}
\end{equation}
where $\psi$s are independent fully connected layers and $\sigma$ are sigmoid functions. With the gates, for the panoptic queries, the updated queries can be calculated as follows:
\begin{equation}
	\begin{aligned}
        \boldsymbol{q}_{s}^{p} = {G}_{\boldsymbol{X}}^{p} \cdot \psi(\boldsymbol{X}_{s}^{p}) + {G}_{\boldsymbol{q}}^{p} \cdot \psi(\boldsymbol{q}_{s-1}^{p}),
        \label{eq:qu3}
	\end{aligned}
\end{equation}
where $\psi$s are also independent fully connected layers. The updated queries in this stage take advantages of both the queries from the former stage and the grouped features with the former stage masks.

For the depth path, we adopt a similar process as Equation~\ref{eq:qu1} and Equation~\ref{eq:qu2} to calculate depth features $F_G^{d}$ and gates ${G}_{q}^{d}$, ${G}_{X}^{d}$. The features and queries from both the panoptic path and depth path will contribute to the current stage depth query:
\begin{equation}
	\begin{aligned}
        \boldsymbol{q}_{s}^{d} = {G}_{\boldsymbol{X}}^{d} \cdot \psi(\boldsymbol{X}_{s}^{d}) + {G}_{\boldsymbol{q}}^{d} \cdot \psi(\boldsymbol{q}_{s-1}^{d}) + \boldsymbol{q}_{s}^{p}.
	\end{aligned}
\end{equation}
We call this process Query Linking. The query linking process will make the depth query get the instance-level information from both the panoptic path and the depth path, and further bolster the mutual benefit between depth and panoptic predictions.

\noindent
\textbf{Query Reasoning.} 
With the updated queries, we perform QR to predict the depth and mask results. We simply adopt Multi-Head Self Attention~\cite{vaswani2017attention} (MSA), Feed-Forward Neural Network~\cite{vaswani2017attention} (FFN), followed by FC-LN-ReLU layers on queries, which has been proved to be effective in~\cite{zhang2021knet}. The depth and mask predictions are then calculated by:
\begin{equation}
	\begin{aligned}
        \boldsymbol{{m}}_s = FFN\left(MSA\left({q}_{s}^{m}\right)\right)\cdot\boldsymbol{X}^{m}_{\mathtt{FPN}}, \\
        \boldsymbol{{d}}_s = FFN\left(MSA\left({q}_{s}^{d}\right)\right)\cdot\boldsymbol{X}^{d}_{\mathtt{FPN}}.
	\end{aligned}
\end{equation}
The $\boldsymbol{{m}}_s$ and $\boldsymbol{{d}}_s$ are instance-level predictions of stage $s$. Note that both query updating and reasoning perform individually (Panoptic Path and Depth Path in Figure~\ref{fig:methods}). The effect of each component can be found in the experiment part.

In our framework, the queries are refined in each stage to be applied to the features from FPN for panoptic and depth paths to get the predictions. The predictions are conducted with the bipartite match for calculating loss during training and merged into dense predictions when inference. All predictions contribute to the final loss, but only the final stage predictions are used when inference. Please refer to Section~\ref{sec:inference} for details about merging the instance-level depth prediction $\boldsymbol{{d}}_s$ into the final dense depth map.

\noindent
\subsection{Tracking with Thing Query} For the tracking part, we adopt previous work design~\cite{vis_dataset,pang2021quasi} and add a tracking head, where we directly learn appearance embeddings among the different frames. During training, we first match predicted thing masks from the panoptic path to ground truth thing mask according to mask IoU. Then, we use the bound box generated by the ground truth thing mask to conduct ROIAlign~\cite{maskrcnn} for extracting the tracking embeddings. We denote tracking embeddings for each instance from the source frame as $\boldsymbol{v}$, and positive and negative tracking embeddings as $\boldsymbol{k^{+}}$ and $\boldsymbol{k^{-}}$. The tracking embedding loss can be calculated as:
\begin{align}
    \mathcal{L}_\text{track} =
    \text{log}[1 + \sum_{\boldsymbol{k}^{+}} \sum_{\boldsymbol{k}^{-}} \text{exp}(\boldsymbol{v} \cdot \boldsymbol{k}^{-} - \boldsymbol{v} \cdot \boldsymbol{k}^{+})].
    \label{eq:quasi:apperance}
\end{align}
During the inference, we use the thing masks, which are generated from the thing queries to obtain final tracking embeddings for tracking, as shown in the bottom part of Fig.~\ref{fig:methods}. The bi-directional softmax is calculated to associate the instances between two frames (annotated as $\textbf{n}$ and $\textbf{m}$):
\begin{equation}
    \textbf{f}(i, j) = [\frac{ \text{exp}(\textbf{n}_i \cdot \textbf{m}_j)}{\sum_{k=0}^{M-1} \text{exp}(\textbf{n}_i \cdot \textbf{m}_k )} +
    \frac{\text{exp}(\textbf{n}_i \cdot \textbf{m}_j)}{\sum_{k=0}^{N-1} \text{exp}(\textbf{n}_k \cdot \textbf{m}_j )}] / 2.
\end{equation}

\noindent
\subsection{Training} To train the proposed PolyphonicFormer, we need to assign ground truth according to cost since all the outputs are encoded via queries. We also assign ground truth for the initial-stage instance segmentation. In particular, we mainly follow the design of MaskFormer and Max-Deeplab~\cite{cheng2021maskformer,wang2020maxDeeplab} to use bipartite matching via a cost considering both segmentation mask and classification results of them. For depth prediction, we use the same bipartite matching results to assign depth ground truth for each query.

After the bipartite matching, we apply a loss jointly considering depth estimation, mask prediction, and classification for each thing or stuff. For depth loss, following ViP-Deeplab~\cite{ViPDeepLab}, we use a joint loss including scale-invariant depth loss~\cite{eigen2014depth}, absolute relative loss, and square relative loss. We did not find an advantage in our framework with the monodepth2~\cite{godard2019digging} activation strategy, so we use a sigmoid activation function and multiply the max distance for simplicity.
For panoptic segmentation, we apply focal loss~\cite{focal_loss} on classification and adopt dice loss, focal loss, and cross-entropy loss on mask predictions. The loss for each stage can be formulated as follows:
\begin{equation}
    \mathcal{L}_{i} = \lambda_{depth} \cdot \mathcal{L}_{depth} + \lambda_{mask} \cdot \mathcal{L}_{mask} + \lambda_{cls} \cdot \mathcal{L}_{cls}.
\end{equation}
Note that the losses are applied to each stage, i.e.,
\begin{equation}
     \mathcal{L}_{final} = \sum_{i}^N (\lambda_{i} \cdot \mathcal{L}_i) + \lambda_{track} \cdot \mathcal{L}_{track},
\end{equation}
where $N$ is the total stages applied to the framework. We adopt $N = 3$ in PolyphonicFormer, and we set $\lambda_{i} = 1$ for all stages. The $\lambda_{track}$ is set to 0.25.

\noindent
\subsection{Inference} \label{sec:inference}
We directly get the instance-level panoptic and depth predictions from the corresponding queries. For final panoptic segmentation results, we adopt the method used in Panoptic-FPN~\cite{kirillov2019panopticfpn} to merge panoptic masks. Since the depth queries and panoptic queries have a one-to-one correspondence, the panoptic segmentation results are used to merge the instance-level depth predictions, i.e.:
\begin{equation}
    \boldsymbol{D}(u, v) = \boldsymbol{d}(\boldsymbol{I}(u,v), u, v), \boldsymbol{D}\in\boldsymbol{R}^{H \times W},
\end{equation}
where $\boldsymbol{d}$ is the instance-level depth predictions with a $N \times H \times W$ shape from the last stage, $\boldsymbol{I}$ is the instance ID for each pixel, which is from the panoptic segmentation results. The dense depth prediction $\boldsymbol{D}$ ($H \times W$) is then calculated. 

For tracking, after getting the embeddings of each instance, we calculate the similarities between the embeddings stored before with the bidirectional softmax similarity. If we detect a new instance embedding that does not appear before, we will store the embedding for future matching. We note that this process only needs \textit{single image input} for each sample during the inference process.

\section{Experiment}
\label{sec:experiment}
\noindent
\textbf{Experimental Setup.}  We implement our models with Pytorch~\cite{paszke2019pytorch}, and follow the same training settings from Panoptic Deeplab~\cite{cheng2020panoptic} and ViP-Deeplab~\cite{ViPDeepLab} where we first pretrain our model on both Mapillary~\cite{neuhold2017mapillary} and Cityscapes~\cite{cordts2016cityscapes} datasets for Panoptic Segmentation, and then we fine-tune the model with our depth query on Cityscapes-DVPS and SemKITTI-DVPS. During pretraining, we randomly resize the origin images with the scale from 1.0 to 2.0, and then we perform random crops with the $2048 \times 1024$ for Cityscapes-DVPS and $1280 \times 384$ for SemKITTI-DVPS. The batch size is set to 16, and we adopt the synchronized batch normalization during the training. For video training settings, we randomly sample one nearly frame to learn the tracking embeddings. All the models use the \textit{single scale inference}. We adopt Swin-B and Resnet-50 backbones, and we add RFP~\cite{qiao2021detectors} following ViP-Deeplab~\cite{ViPDeepLab} but for Swin-B backbone only. We carry out ablation studies on Cityscapes-DVPS datasets for evaluating panoptic segmentation results and depth results.

\begin{table*}[!t]
\renewcommand{\arraystretch}{0.80}
\small
\centering
\resizebox{\textwidth}{!}{
    \begin{tabular}{l|c|c|c|c|c|c}
    \toprule[0.2em]
    $\text{DVPQ}_\lambda^k$ on Cityscapes-DVPS & k = 1 & k = 2 & k = 3 & k = 4 & Average & FLOPs\\
    \toprule[0.2em]
    PolyphonicFormer $\lambda$ = 0.50 & 70.6 $\vert$ 63.0 $\vert$ 76.0 & 62.9 $\vert$ 49.2 $\vert$ 72.9 & 59.3 $\vert$ 42.3 $\vert$ 71.7 & 56.5 $\vert$ 36.9 $\vert$ 70.8 & 62.3 $\vert$ 47.9 $\vert$ 72.9 & - \\
    PolyphonicFormer $\lambda$ = 0.25 & 67.8 $\vert$ 61.0 $\vert$ 72.8 & 60.4 $\vert$ 47.6 $\vert$ 69.8 & 56.9 $\vert$ 40.8 $\vert$ 68.6 & 54.3 $\vert$ 35.8 $\vert$ 67.8 & 59.9 $\vert$ 46.3 $\vert$ 69.8 & - \\
    PolyphonicFormer $\lambda$ = 0.10 & 50.2 $\vert$ 43.4 $\vert$ 55.2 & 44.4 $\vert$ 33.4 $\vert$ 52.4 & 41.5 $\vert$ 28.6 $\vert$ 51.0 & 39.5 $\vert$ 24.7 $\vert$ 50.4 & 43.9 $\vert$ 32.5 $\vert$ 52.3 & - \\
    Average: PolyphonicFormer & 62.9 $\vert$ 55.8 $\vert$ 68.0 & 55.9 $\vert$ 43.4 $\vert$ 65.0 & 52.6 $\vert$ 37.2 $\vert$ 63.8 & 50.1 $\vert$ 32.5 $\vert$ 63.0 &  \textbf{55.4} $\vert$ 42.2 $\vert$ 65.0 & 1,675G \\ 
    \midrule
    Average: ViP-Deeplab~\cite{ViPDeepLab} & 61.9 $\vert$ 55.9 $\vert$ 66.3 & 55.6 $\vert$ 44.3 $\vert$ 63.8 & 52.4 $\vert$ 38.4 $\vert$ 62.6 & 50.4 $\vert$ 34.6 $\vert$ 61.9 & 55.1 $\vert$ 43.3 $\vert$ 63.6 & 9,451G \\
    
    \bottomrule[0.1em]
    \toprule[0.2em]
    $\text{DVPQ}_\lambda^k$ on SemKITTI-DVPS & k = 1 & k = 5 & k = 10 & k = 20 & Average & FLOPs\\
    \toprule[0.2em]
    PolyphonicFormer $\lambda$ = 0.50 & 58.5 $\vert$ 55.1 $\vert$ 61.0 & 52.0 $\vert$ 42.3 $\vert$ 59.1 & 50.6 $\vert$ 39.9 $\vert$ 58.5 & 49.9 $\vert$ 38.6 $\vert$ 58.0 & 52.8 $\vert$ 44.0 $\vert$ 59.2 & - \\
    PolyphonicFormer $\lambda$ = 0.25 & 56.3 $\vert$ 54.0 $\vert$ 57.9 & 49.7 $\vert$ 41.1 $\vert$ 56.0 & 48.4 $\vert$ 38.7 $\vert$ 55.5 & 47.7 $\vert$ 37.6 $\vert$ 55.0 & 50.5 $\vert$ 42.9 $\vert$ 56.1 & - \\
    PolyphonicFormer $\lambda$ = 0.10 & 41.8 $\vert$ 41.1 $\vert$ 42.4 & 35.1 $\vert$ 28.2 $\vert$ 40.1 & 33.7 $\vert$ 26.0 $\vert$ 39.3 & 33.0 $\vert$ 25.1 $\vert$ 38.7 & 35.9 $\vert$ 30.1 $\vert$ 40.1 & - \\
    Average: PolyphonicFormer & 52.2 $\vert$ 50.1 $\vert$ 53.8 & 45.6 $\vert$ 37.2 $\vert$ 51.7 & 44.2 $\vert$ 34.9 $\vert$ 51.1 & 43.4 $\vert$ 33.8 $\vert$ 50.6 & \textbf{46.4} $\vert$ 39.0 $\vert$ 51.8 & 402G \\
    \midrule
    Average: ViP-Deeplab~\cite{ViPDeepLab} & 48.9 $\vert$ 42.0 $\vert$ 53.9 & 45.8 $\vert$ 36.9 $\vert$ 52.3 & 44.4 $\vert$ 34.6 $\vert$ 51.6  & 43.4 $\vert$ 33.0 $\vert$ 51.1 & 45.6 $\vert$ 36.6 $\vert$ 52.2& ~2,267G \\
    \bottomrule[0.1em]
    \end{tabular}
}
    \caption{\small Experiment results on Cityscapes-DVPS and SemKITTI-DVPS. Each cell shows $\text{DVPQ}_{\lambda}^{k} ~\vert~ \text{DVPQ}_{\lambda}^{k}\text{-Thing} ~\vert~ \text{DVPQ}_{\lambda}^{k}\text{-Stuff}$ where $\lambda$ is the threshold of relative depth error, and $k$ is the number of frames. Smaller $\lambda$ and larger $k$ correspond to a higher accuracy requirement. PolyphonicFormer adopts Swin-B backbone in this table.
    }
    \label{tab:dvps_results}
\end{table*}

\noindent
\textbf{Evaluation Metrics.}
We use DVPQ metric following ViP-Deeplab~\cite{ViPDeepLab}. Let $P$ and $Q$ be the prediction and ground truth respectively, k be the window size, and $\lambda$ be the depth threshold. We use $P^c$, $P^{id}$, and $P^d$ to represent semantic segmentation, the instance segmentation, and depth estimation results respectively; $Q^c$, $Q^{id}$, and $Q^d$ are alike but for ground truth. The $DVPQ^k_{\lambda}$ metric can be formulated as: $\text{PQ}\Big( \Big[\concat_{i=t}^{t+k-1}\big(\hat{P_i^c}, P_i^{id}\big), \concat_{i=t}^{t+k-1}\big(Q^c_i, Q_i^{id}\big)\Big]_{t=1}^{T-k+1} \Big).$
where $\hat{P_i^c} := P_i^c$ for pixels that have an absolute relative depth error inside the error threshold ($\lambda$) and $\hat{P_i^c} := void$ otherwise. $\concat_{i=t}^{t+k-1}(P_i^c, P_i^{id})$ denotes the horizontal concatenation of the pair $(P_i^c, P_i^{id})$ from $t$ to $t+k-1$. 

However, the $DVPQ^k_{\lambda}$ metric does not measure the depth quality and tracking quality explicitly; instead, they are observed by the $DVPQ^k_{\lambda}$ dropping when $k$ and $\lambda$ are getting larger. Therefore, we also report $DSTQ$, which is an extension of $STQ$~\cite{STEP}. 
The definition of $DSTQ$ is: $DSTQ = (AQ \times SQ \times DQ)^{\frac{1}{3}},$ where the $AQ$ and $SQ$ are association quality (tracking quality) and segmentation quality (mIoU) respectively~\cite{STEP}. $DQ$ is depth prediction quality.
In the ablation study, we report widely used PQ~\cite{kirillov2019panoptic} and absolute relative error (abs rel) for panoptic segmentation and depth map quality assessment.


\subsection{Experimental Results}
\noindent
\textbf{Cityscapes-(D)VPS.}
The Cityscapes-DVPS dataset~\cite{ViPDeepLab} is an extension of the Cityscapes-VPS~\cite{kim2020vps} dataset, which contains extra depth annotations.
Different from VPSNet~\cite{kim2020vps}, we only use the keyframes for training and inference, which means our setting needs fewer inputs and execution rounds to get the results compared to VPSNet~\cite{kim2020vps}, and thus might be more challenging. The results on the Cityscapes-VPS dataset are shown in Table~\ref{tab:vps_results}. In the Cityscapes-VPS dataset, with the very same ResNet-50 backbone, the proposed PolyphonicFormer outperforms the existing methods, VPSNet~\cite{kim2020vps} and SiamTrack~\cite{woo2021learning_associate_vps} drastically, although we successfully get rid of the optical flow input. Note that we did not fine-tune our methods on Cityscapes-VPS after training on Cityscapes-DVPS since depth estimation in our framework will not damage the performance of panoptic segmentation performance in our framework. As shown in Table~\ref{tab:dvps_results}, in the Cityscapes-DVPS dataset, our methods achieve 55.4 average DVPQ with a Swin-B~\cite{liu2021swin} backbone. The DVPQ results demonstrate that the proposed PolyphonicFormer has a good performance on the DVPS task and its sub-tasks.

\begin{table}[!t]
\renewcommand{\arraystretch}{0.9}
\small
    \centering
    \begin{minipage}{\textwidth}
    \centering
	\begin{minipage}{\dimexpr.57 \linewidth}
	\resizebox{0.80\textwidth}{!}{
        \begin{tabular}{l|ccccc}
        \toprule[0.2em]
        Method &  k = 1 & k = 2 & k = 3 & k = 4 & VPQ \\
        \toprule[0.2em]
    
        VPSNet~\cite{kim2020vps}  & 
        65.0 & 57.6 & 54.4 & 52.8 & 57.5 \\
        SiamTrack~\cite{woo2021learning_associate_vps} & 
        64.6 & 57.6 & 54.2 & 52.7 & 57.3 \\
        ViP-Deeplab~\cite{ViPDeepLab} & 
        69.2 & 62.3 & 59.2 & 57.0 & 61.9 \\
        \midrule
        Ours (ResNet50) & 
        65.4 & 58.6 & 55.4 & 53.3 & 58.2 \\
        Ours (Swin-b) & 
        70.8 & 63.1 & 59.5 & 56.8 & 62.3 \\
        \bottomrule[0.1em]
        \end{tabular}
    }
    \end{minipage}
    \medskip
    \begin{minipage}{\dimexpr.33 \linewidth}
    \resizebox{0.80\textwidth}{!}{
        \begin{tabular}{l c c}
    		\toprule[0.15em]
    		Method & $DSTQ$  \\
    		\toprule[0.15em]
            rl-lab & 54.8 \\
            yang26 & 55.6 \\
            Vip-Deeplab & 63.3 \\ 
            \midrule
            PolyphonicFormer & 63.6 \\
            PolyphonicFormer* & 64.6 \\
    	\bottomrule[0.1em]
    	\end{tabular}
    }
	\end{minipage}
    \end{minipage}
    \caption{\small (Left) Experiment results on Cityscapes-VPS validation set. $k=\{0, 5, 10, 15\}$ in~\cite{kim2020vps} correspond to $k=\{1, 2, 3, 4\}$ in this paper as we use different notations. All the methods in this table is \textit{without} test-time augmentation. (Right) Results on ICCV-2021 SemKITTI-DVPS challenge. * indicates post-challenge submission.}
    \label{tab:vps_results}
    \label{tab:results_dvps_kitti}
\end{table}

\noindent
\textbf{SemKITTI-DVPS.}
The SemKITTI-DVPS dataset has 19,130 / 4,071 / 4,342 images for training, evaluation, and testing respectively. SemKITTI-DVPS is a challenging dataset since the annotation of both the depth and panoptic segmentation are sparse. As the sparse nature of annotations of SemKITTI-DVPS, following ViP-Deeplab~\cite{ViPDeepLab}, we consider the unannotated pixels as unknown when performing the evaluation. The Table~\ref{tab:dvps_results} reports our DVPQ results on SemKITTI-DVPS validation set. The PolyphonicFomer achieves a 46.4 average DVPQ. On the test set (ICCV-2021 SemKITTI-DVPS challenge), our method won the championship with 63.6 DSTQ, as shown in Table~\ref{tab:results_dvps_kitti}. Beyond that, we re-train the proposed PolyphonicFormer with the additional validation set and achieve 64.6 DSTQ, which is the state-of-the-art result on the test set.

\noindent
\textbf{Comparison on DVPQ.} As in Table~\ref{tab:dvps_results}, we compare our proposed PolyphonicFormer (with Swin-B backbone) with previous work ViP-Deeplab~\cite{ViPDeepLab}, and PolyphonicFormer outperforms ViP-Deeplab on both datasets. On Cityscapes-DVPS dataset, the proposed PolyphonicFormer / ViP-Deeplab has $\sim$1675 GFLOPs / $\sim$9451 GFLOPs, respectively. And on SemKITTI-DVPS dataset, PolyphonicFormer / ViP-Deeplab has $\sim$402 GFLOPs / $\sim$2267 GFLOPs, respectively. Note that we do not do the experiments on the much larger WR-41 backbone~\cite{chen2020naive}, which is the backbone used in ViP-Deeplab~\cite{ViPDeepLab}, due to the limitation of computational resources. We also do not apply \textit{test-time augmentation} and \textit{semi-supervised learning} (both proved to be effective in~\cite{ViPDeepLab}) for simplicity.


\noindent
\subsection{Ablation Studies and Analysis}
\noindent
\begin{table*}[!t]
    \footnotesize
	\centering
    \scalebox{0.55}{\subfloat[]{
        \small
        \label{tab:eff_query}
	    \begin{tabularx}{9.3cm}{c|c|c|c|c|c} 
		    \toprule[0.15em]
    		 Method & Depth & Panoptic & Ins & PQ $\uparrow$  & abs rel $\downarrow$ \\
    		\midrule[0.15em]
    		\textbf{ViP-Deeplab~\cite{ViPDeepLab}}  & \checkmark & \checkmark & -  & 60.6 &0.112 \\ 
    		\midrule[0.15em]
    		\textbf{Depth} & \checkmark & -  & -&  N/A & 0.084\\ 
            \textbf{Panoptic} & - & \checkmark & -  & 63.7 & N/A\\
            \hline
            \textbf{Hybrid (ours)} & \checkmark & \checkmark & -  & 65.1 & 0.089\\
            \textbf{PolyphonicFormer (ours)} & \checkmark & \checkmark & \checkmark & 65.2 & 0.080\\
        	\bottomrule[0.1em]
	    \end{tabularx}
    }} \hfill
    \scalebox{0.7}{\subfloat[]{
        \label{tab:depth_loss}
		\begin{tabularx}{3.0cm}{c|c|c} 
			\toprule[0.15em]
			$L_{depth}$ & PQ $\uparrow$ & abs rel $\downarrow$ \\
			\midrule[0.15em]
			    0.1 & 65.4 & 0.101 \\ 
                1.0 & 65.3 & 0.089\\
                5.0 & 65.2 & 0.080 \\
                10 & 65.4 & 0.079 \\
			\bottomrule[0.1em]
		\end{tabularx}
    }} \hfill
    \scalebox{0.7}{\subfloat[]{
        \label{tab:interaction}
		\begin{tabularx}{4.9cm}{l|c|c} 
			\toprule[0.15em]
			Method & PQ $\uparrow$ & abs rel $\downarrow$ \\
			\midrule[0.15em]
			    PolyphonicFormer & 65.2 & 0.080 \\ 
			    w/o Query Linking & 64.8 & 0.088 \\ 
			    w/o Dense Init & 63.2 & 0.094 \\
			    w/o Both & 63.0 & 0.104 \\
			\bottomrule[0.1em]
		\end{tabularx}
    }} \hfill
    \vspace{1mm}
    \scalebox{0.65}{\subfloat[]{
        \label{tab:ql_design}
	    \begin{tabularx}{3.3cm}{c|c|c} 
		\toprule[0.15em]
    		 Method  & PQ $\uparrow$ & abs rel $\downarrow$ \\
    		\toprule[0.15em]
    	       P & 65.2  &0.080 \\
    	       B &  63.8 &0.087 \\
    	       A  & 65.3  &0.081 \\
        	\bottomrule[0.1em]
	    \end{tabularx}
    }} \hfill
    \scalebox{0.65}{\subfloat[]{
        \label{tab:tracking_head}
	    \begin{tabularx}{7.5cm}{c|c|c} 
		\toprule[0.15em]
    		 Method & DSTQ $\uparrow$ & AQ $\uparrow$ \\
    		\toprule[0.15em]
    	       PolyphonicFormer + DeepSort~\cite{wojke2017simple} & 51.8 & 25.9 \\
    	       PolyphonicFormer + Unitrack~\cite{wangUnitrack} &  49.3 &  22.5 \\
    	       PolyphonicFormer + QuasiDense~\cite{pang2021quasi} & 63.6 &  46.2 \\
        	\bottomrule[0.1em]
	    \end{tabularx}
    }} \hfill
    \scalebox{0.65}{\subfloat[]{
        \label{tab:iter_rounds}
	    \begin{tabularx}{3.1cm}{c|c|c} 
		\toprule[0.15em]
    		 Stages  & PQ $\uparrow$ & abs rel $\downarrow$ \\
    		\toprule[0.15em]
    	       1 & 64.1  &0.081 \\
    	       2 &  64.6 &0.081 \\
    	       3  & 65.2 &0.080 \\
        	\bottomrule[0.1em]
	    \end{tabularx}
    }} \hfill
	\label{tab:ablation}
	\caption{\small Ablation studies on PolyphonicFormer. \textbf{N/A} indicates \textbf{``not applicable"}, and \textbf{w/o} indicates \textbf{``without"}. ``$\uparrow$" indicates higher is better, and ``$\downarrow$" indicates lower is better. Please refer to each paragraph for the description of the abbreviations and results analysis. All experiments use ResNet-50 backbone if no further description.}
\end{table*}

\noindent
\textbf{(a).Effect of Unified Framework.}
We evaluate the effect of the unified framework, and the results can be found in Table~\ref{tab:eff_query}. In Table~\ref{tab:eff_query}, \textbf{Depth} refers to the dense prediction baseline of depth estimation. \textbf{Panoptic} refers to the panoptic segmentation baseline via query learning, i.e., panoptic path only design in Figure~\ref{fig:methods}. \textbf{Hybrid} means that PolyphonicFormer removes the instance-level depth prediction scheme (\textbf{Ins} in the table) but only contains the initial dense prediction head. As shown in the table, although the single-task baselines are strong for multi-task methods, our method still improves beyond two single-task baselines on both depth and panoptic segmentation via query learning. The performance enhancement on depth estimation from \textbf{Hybrid} to \textbf{PolyphonicFormer} indicates the effectiveness of instance-level depth estimation scheme. We also notice that the performance on panoptic segmentation increased from \textbf{Panoptic} to \textbf{Hybrid}, and we think it may be because depth supervision at least does not affect and may have some benefits on segmentation in our framework.

\noindent
\textbf{(b).Robustness of PolyphonicFormer.}
We modify the depth loss weight to evaluate the robustness of setting different depth loss weights. We set the depth loss weights from 0.1, 1, to 5 and 10. The results are shown in Table~\ref{tab:depth_loss}. We found that with different depth loss weights, the PQ results do not change drastically. The phenomenon is different from ViP-Deeplab~\cite{ViPDeepLab} where the depth estimation and panoptic segmentation are with two separate heads and sensitive to the loss choices. We speculate that it is the unified framework that makes two different tasks ``polyphonic" instead of battling with each other.

\noindent
\textbf{(c).Ablation Study on Panoptic Depth Interaction.}
We perform ablation study on panoptic depth interaction. As shown in Table~\ref{tab:interaction}, we do experiments on the original settings along with dismissing the ``Query Linking" and ``Dense Init" (Dense Depth Initialization). ``Dense Depth Initialization" is a strategy to initialize the depth queries. We empirically found that instead of initializing the depth queries randomly, initializing it with the queries at the initial stage of depth path is with a better performance. We also found that the ``Query Linking" module is a simple but effective module to boost the mutual benefit.

\noindent
\textbf{(d).Effect of Query Linking Module Design.}
We have tried different forms of query linking module design. In Table~\ref{tab:ql_design}, ``\textbf{P}" refers to the current PolyphonicFormer design, which simply uses an addition from the panoptic queries to the depth queries. We try to replace the current design to ``\textbf{B}", a bidirectional query linking module that links queries not only from panoptic to depth but also from depth to panoptic. We found that the information from depth queries may not be that helpful to panoptic queries. We also try to apply a more advanced query linking module, for instance, a self-attention module to fuse panoptic queries and depth queries (``\textbf{A}" in the table). The results may indicate that the query linking itself is more important than the linking module design choice. So, we would like to keep it simple and use the current query linking design, an \textbf{ADD} module, to propagate the semantic context to the geometry context.

\noindent
\textbf{(e).Ablation Study on Tracking Head Design.}
The DVPQ~\cite{ViPDeepLab} does not have an explicit tracking quality metric, so we apply Association Quality (AQ) in DSTQ metric~\cite{STEP} for tracking head evaluation. To alleviate the deviation of tracking quality caused by the mask prediction's inaccuracy, we choose to use Swin-B backbone~\cite{liu2021swin} here. As demonstrated in Table~\ref{tab:tracking_head}, we explore three different tracking methods including DeepSort~\cite{wojke2017simple}, Unitrack~\cite{wangUnitrack} and QuansiDense~\cite{pang2021quasi}. We found the QuansiDense tracker works best and is chosen as the final tracking head. The QuansiDense tracker is an online tracker and only uses appearance embeddings. For the fair comparison with ViP-Deeplab~\cite{ViPDeepLab}, we do not explore the geometry information for tracking in the current framework.

\noindent
\textbf{(f).Ablation Study on Iterative Rounds.}
The PolyphonicFormer performs in an iterative manner. We carry out the ablation study on the choice of iterative stages. We perform studies from 1 to 3 stages to study the effectiveness of iterative design. The results can be found in Table~\ref{tab:iter_rounds}. With more iterative rounds, as we expected, the instance-level information from the features could be more used to refine the query to predict more accurate mask and depth results. We have also tried to use more iterative rounds and do not found salient performance enhancement, so we choose to set the number of stages to 3 following~\cite{zhang2021knet}.

\noindent
\subsection{Visualization and Limitation Analysis}


\begin{figure*}[!t]
\centering
\begin{subfigure}{.245\linewidth}
  \centering
  \includegraphics[trim={0 246 0 246},clip,width=\linewidth]{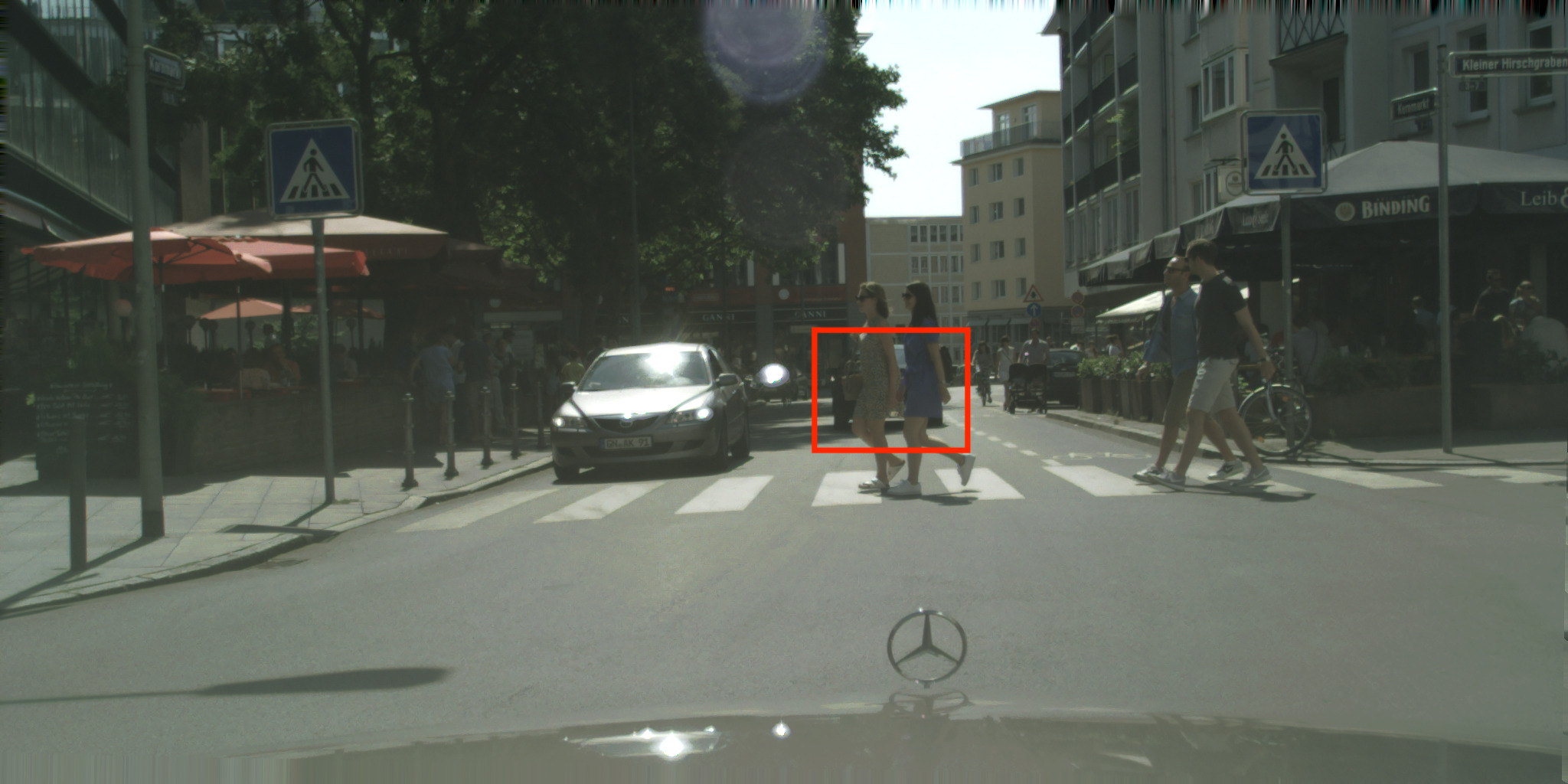}
\end{subfigure}\hfill
\begin{subfigure}{.245\linewidth}
  \centering
  \includegraphics[trim={0 246 0 246},clip,width=\linewidth]{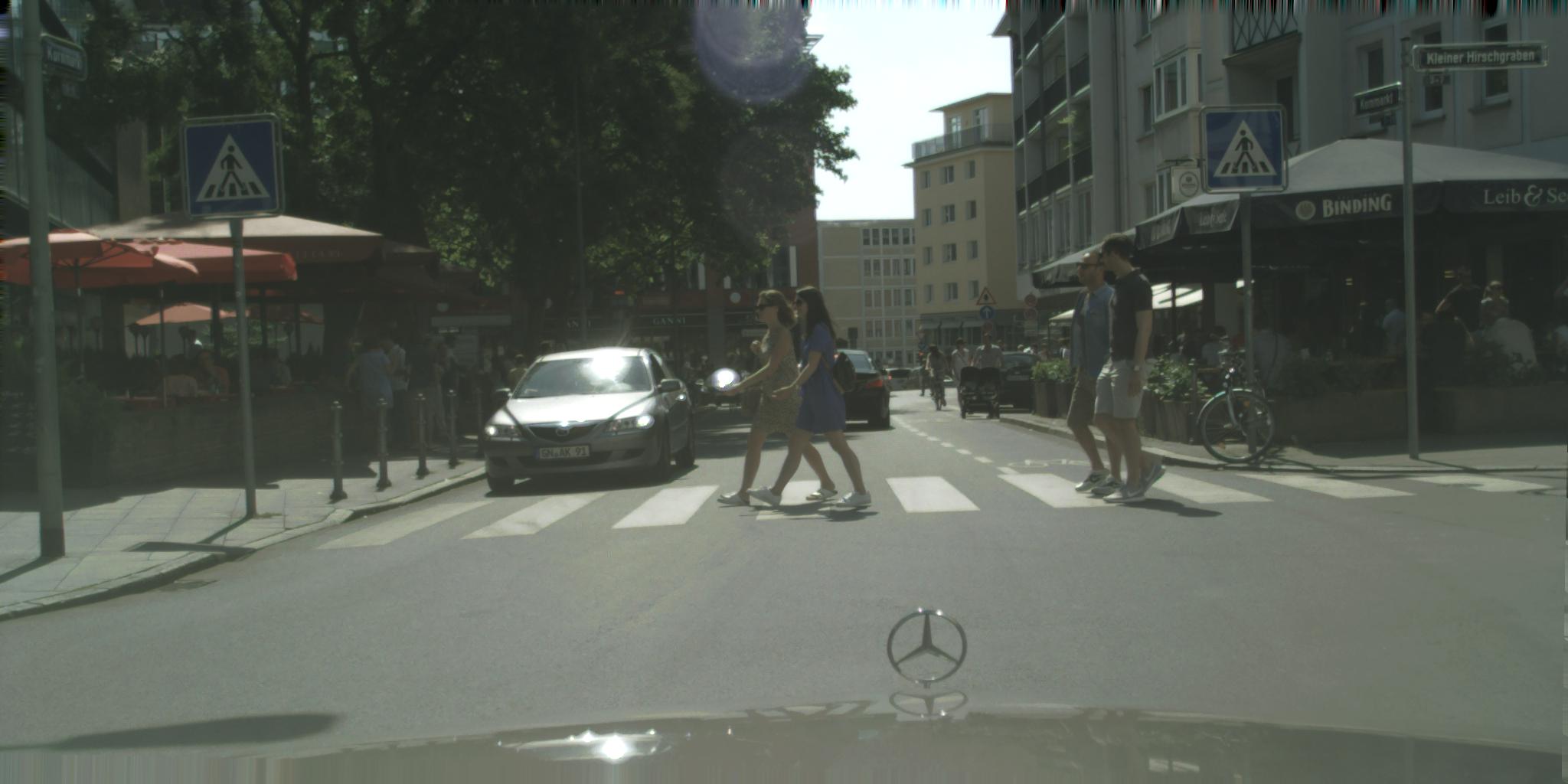}
\end{subfigure}\hfill
\begin{subfigure}{.245\linewidth}
  \centering
  \includegraphics[trim={150 0 150 125},clip,width=\linewidth]{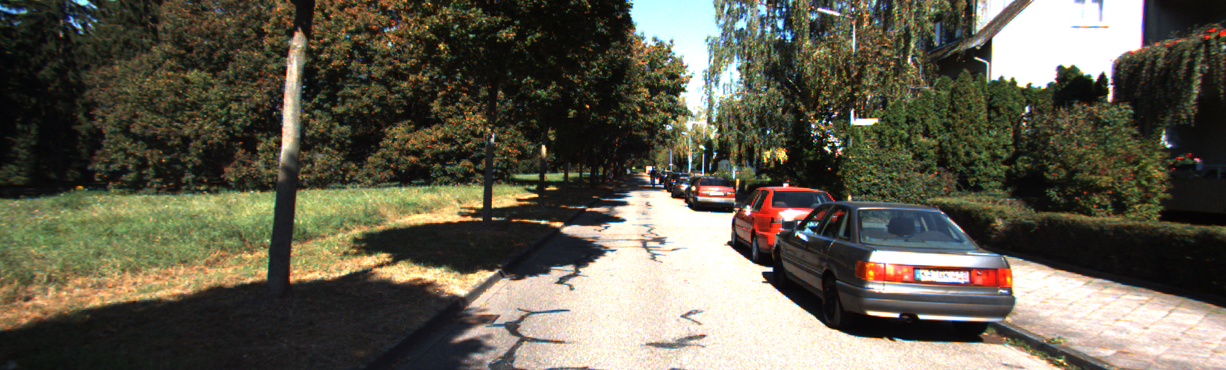}
\end{subfigure}\hfill
\begin{subfigure}{.245\linewidth}
  \centering
  \includegraphics[trim={150 0 150 125},clip,width=\linewidth]{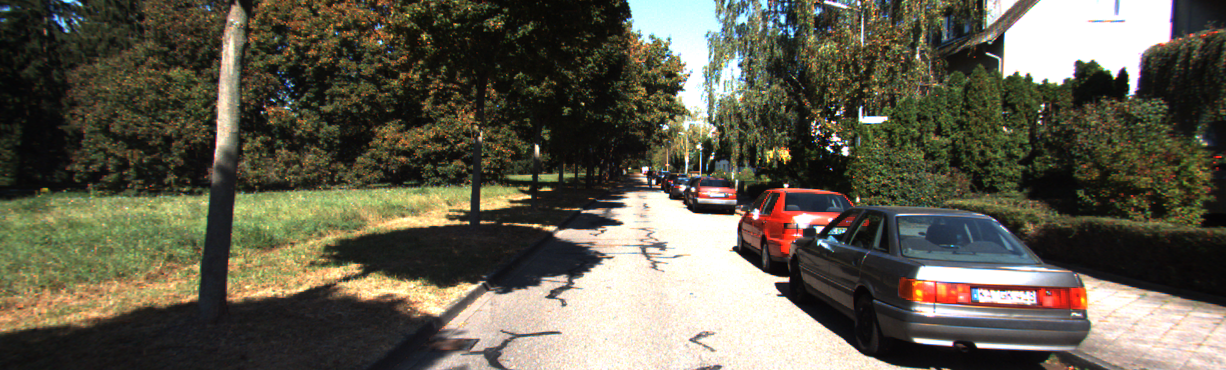}
\end{subfigure}\\

\begin{subfigure}{.245\linewidth}
  \centering
  \includegraphics[trim={0 246 0 246},clip,width=\linewidth]{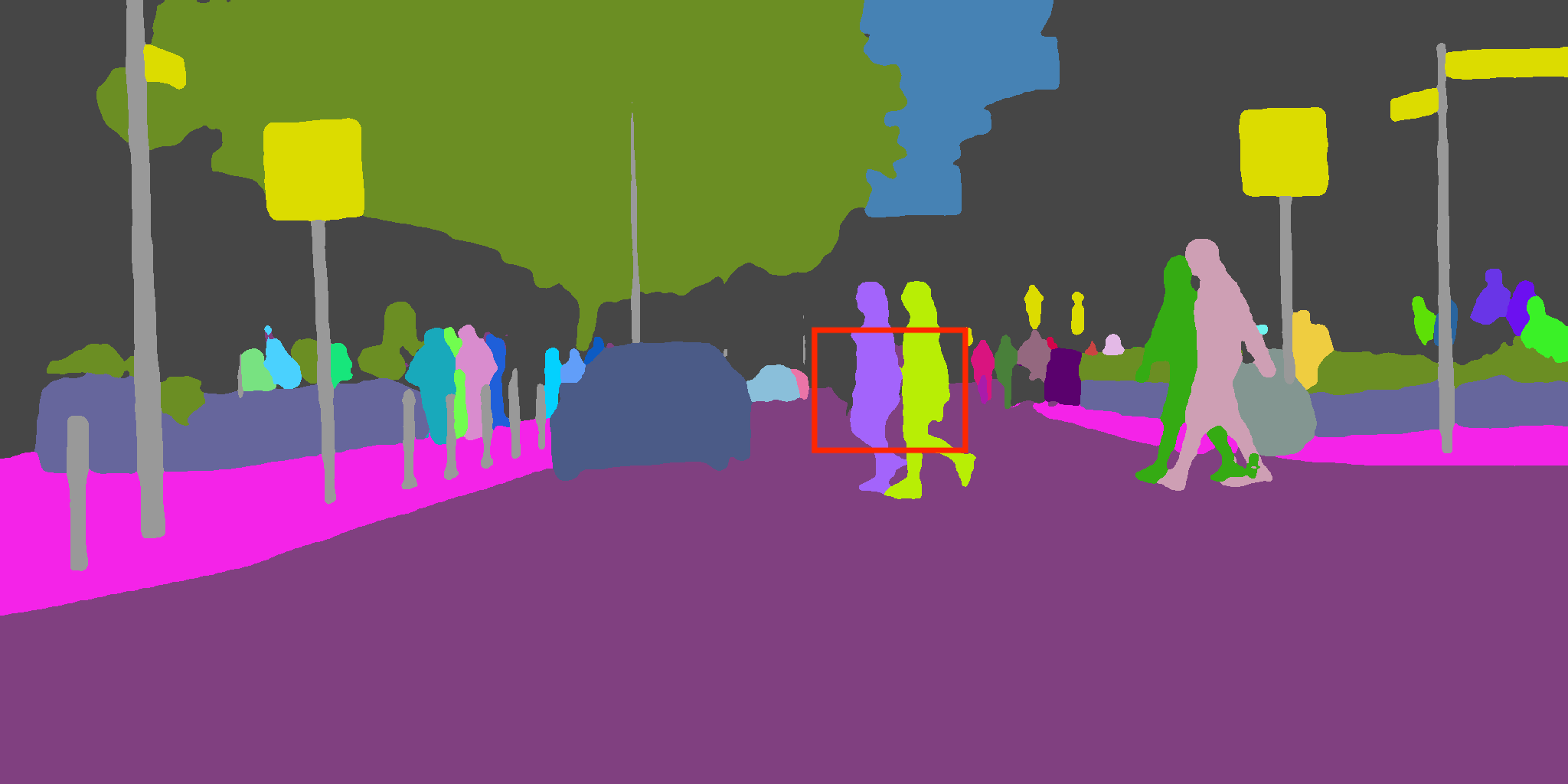}
\end{subfigure}\hfill
\begin{subfigure}{.245\linewidth}
  \centering
  \includegraphics[trim={0 246 0 246},clip,width=\linewidth]{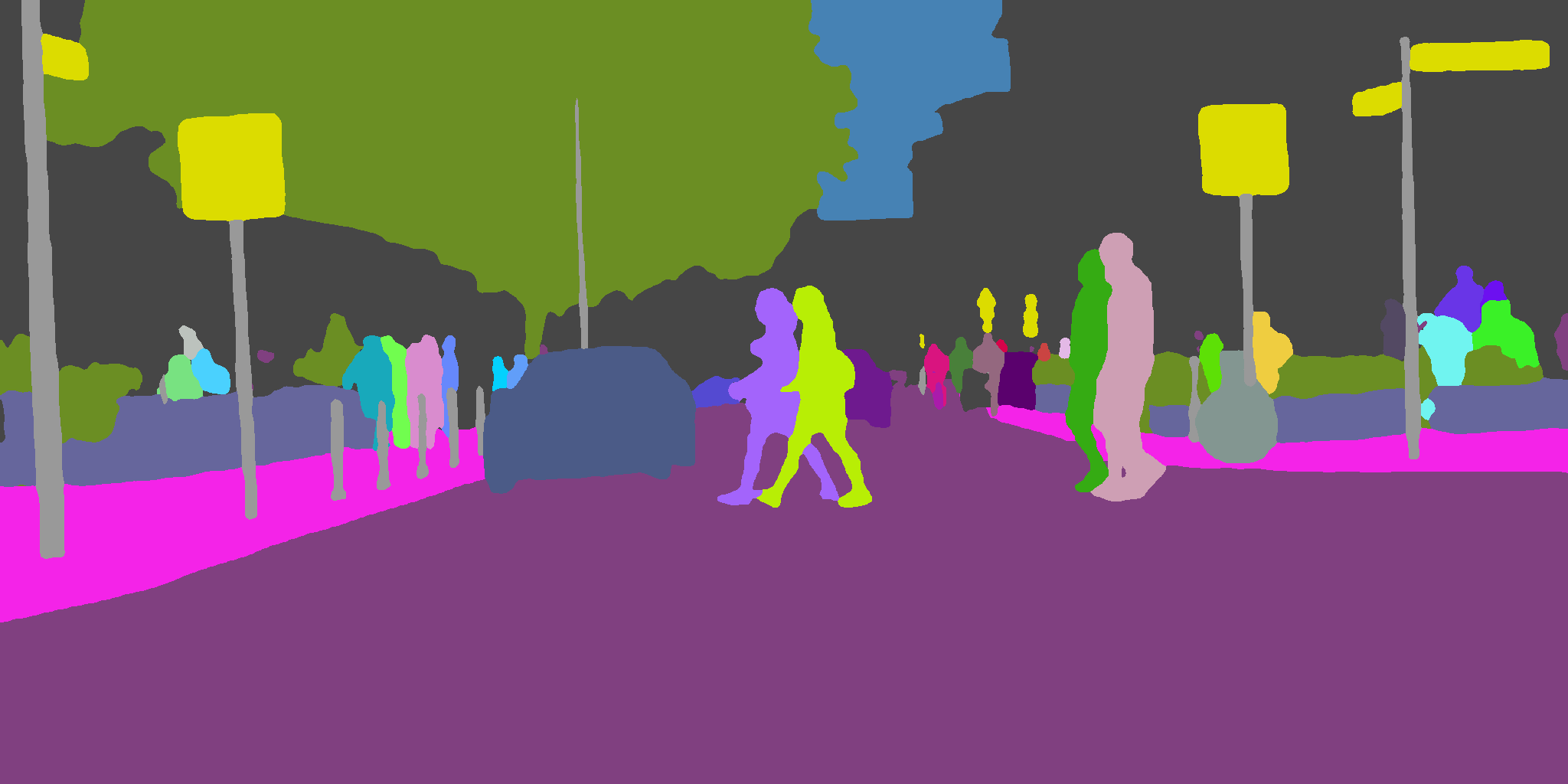}
\end{subfigure}\hfill
\begin{subfigure}{.245\linewidth}
  \centering
  \includegraphics[trim={150 0 150 125},clip,width=\linewidth]{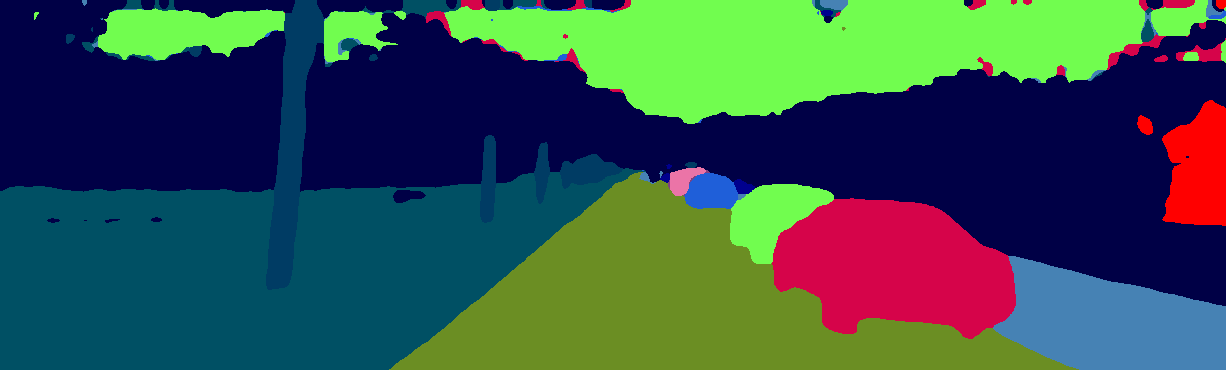}
\end{subfigure}\hfill
\begin{subfigure}{.245\linewidth}
  \centering
  \includegraphics[trim={150 0 150 125},clip,width=\linewidth]{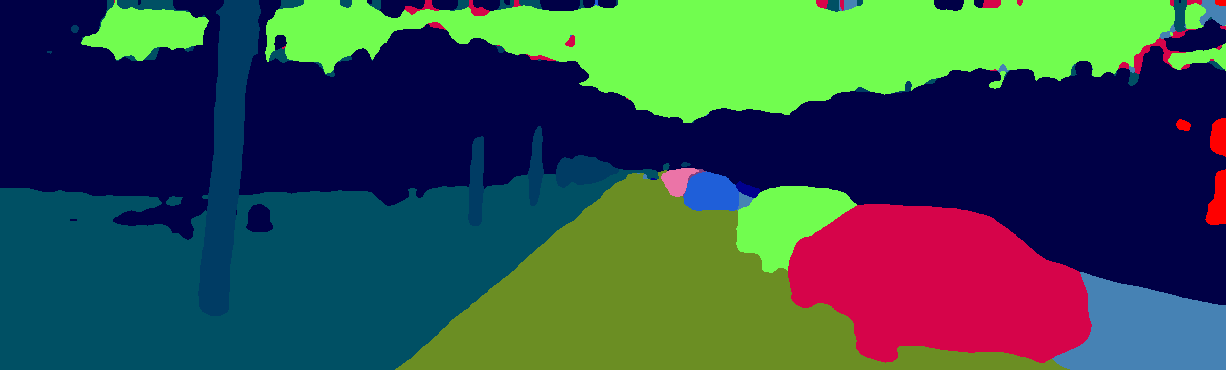}
\end{subfigure}\\

\begin{subfigure}{.245\linewidth}
  \centering
  \includegraphics[trim={0 246 0 246},clip,width=\linewidth]{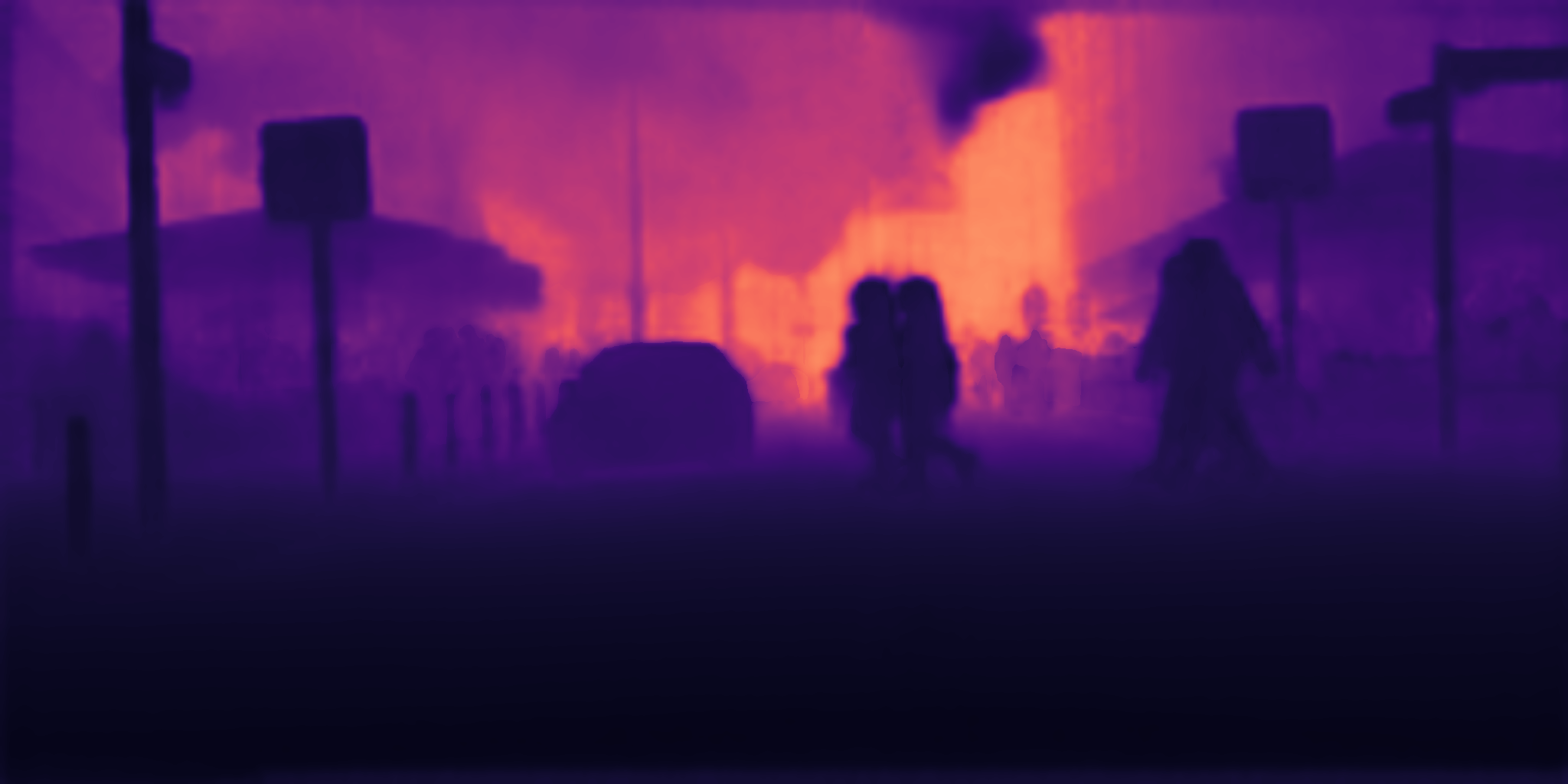}
\end{subfigure}\hfill
\begin{subfigure}{.245\linewidth}
  \centering
  \includegraphics[trim={0 246 0 246},clip,width=\linewidth]{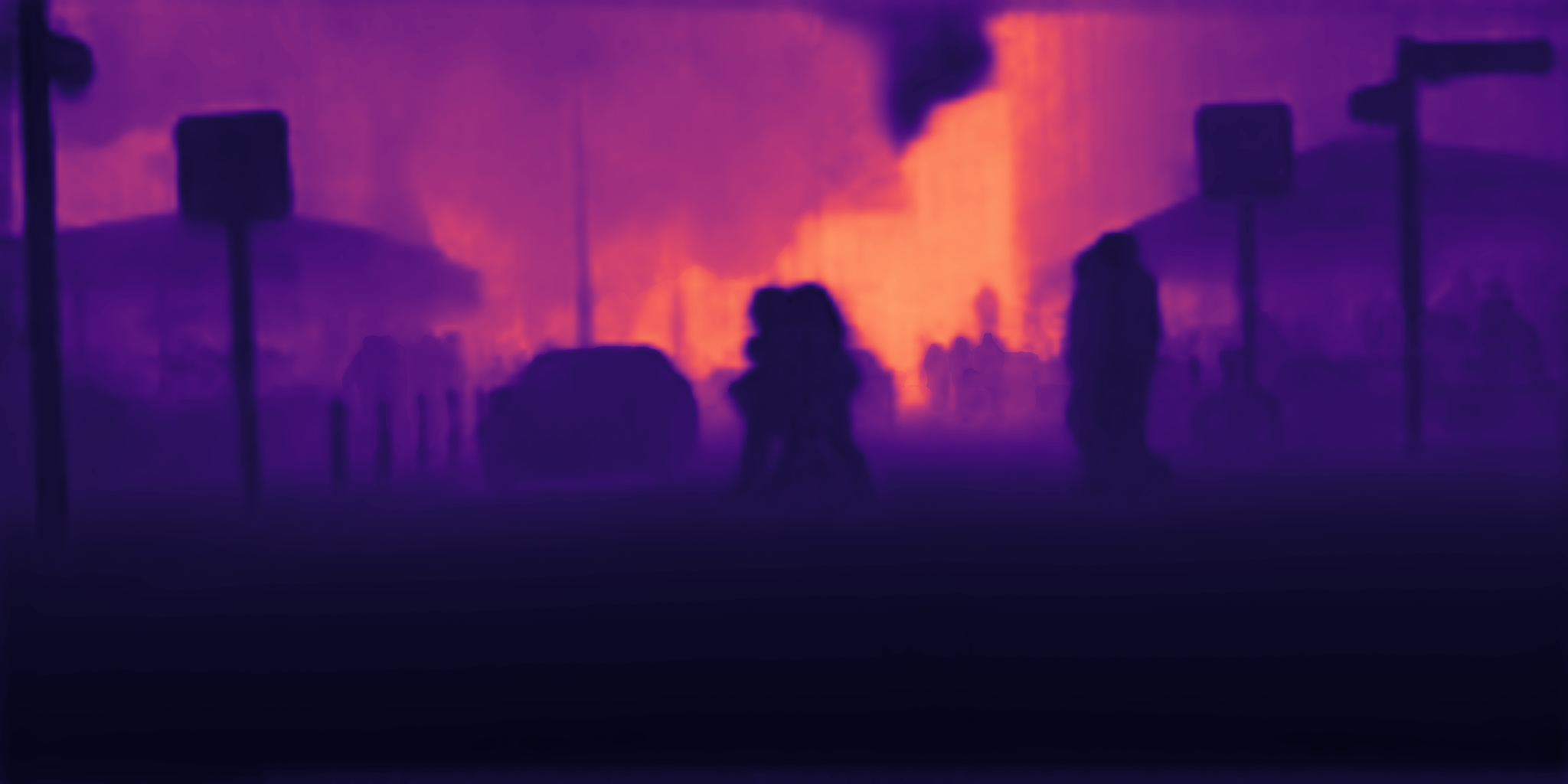}
\end{subfigure}\hfill
\begin{subfigure}{.245\linewidth}
  \centering
  \includegraphics[trim={150 0 150 120},clip,width=\linewidth]{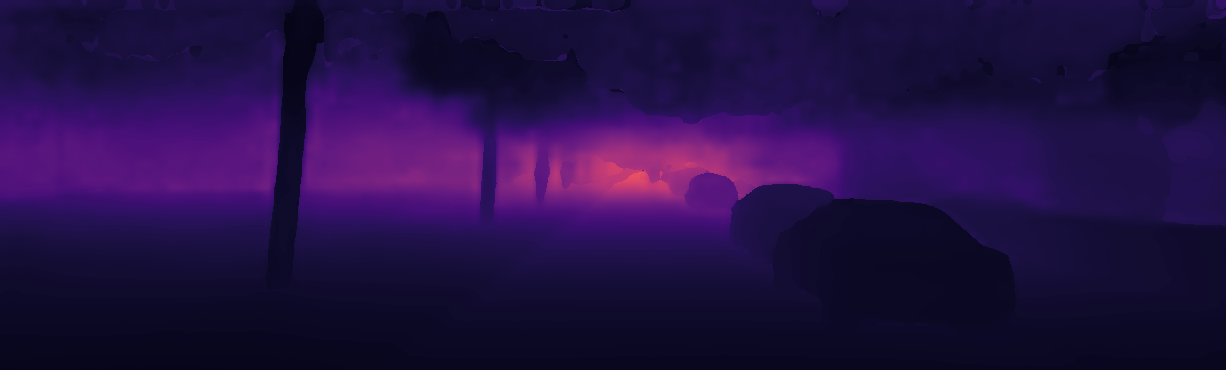}
\end{subfigure}\hfill
\begin{subfigure}{.245\linewidth}
  \centering
  \includegraphics[trim={150 0 150 125},clip,width=\linewidth]{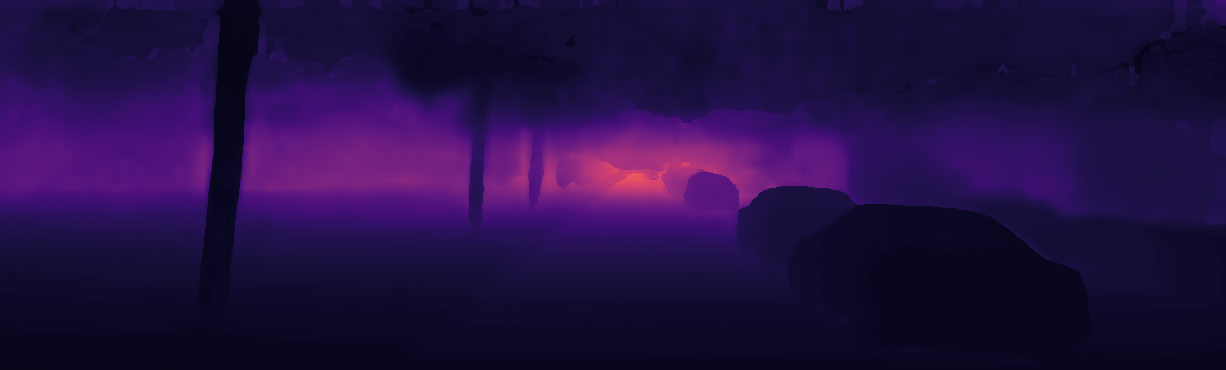}
\end{subfigure}\\

\caption{\small Visualizations of PolyphonicFormer on Cityscapes-DVPS and SemKITTI-DVPS. From top to down: input image, temporally consistent panoptic segmentation prediction, and monocular depth prediction. The \textcolor[RGB]{220,0,0}{red} box is a failure case example.}
\label{fig:visualization_main}
\end{figure*}

\begin{figure}[!t]
    \centering
    \includegraphics[width=0.65\linewidth]{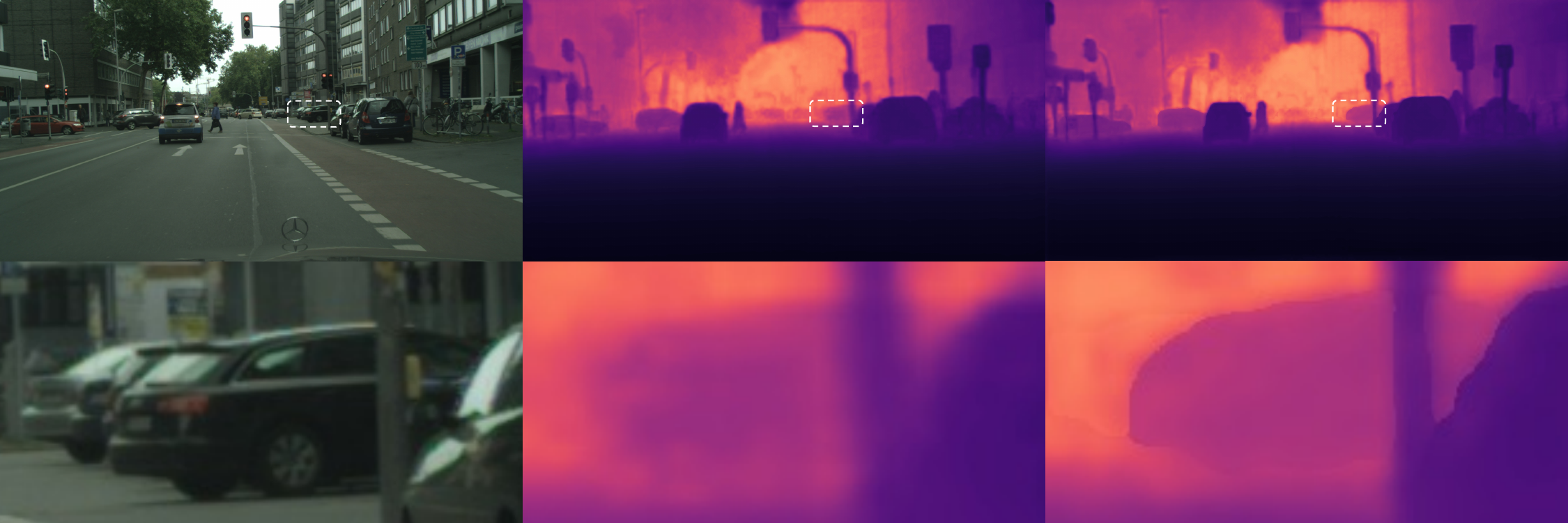}
    \caption{\small The depth estimation results of the traditional dense depth estimation (middle) and PolyphonicFormer (right). For the small objects, the proposed unified framework could successfully distinguish the boundary for enhancing depth results.}
    \label{fig:vis_ablation}
\end{figure}

\noindent
\textbf{Visualization and Analysis.} 
We do visualization analysis on the two DVPS datasets. As shown in Figure~\ref{fig:visualization_main}, the PolyphonicFormer can predict frame-consistent panoptic segmentation along with depth in a unified manner. In the panoptic segmentation map, things with the same color are a single identity over the different frames. Figure~\ref{fig:vis_ablation} shows that the depth estimation task could benefit from the polyphonic design. We take the patch cropped from the depth map (bottom) as an example. The dense prediction neural network has an uncertainty on the boundary and tend to predict vague results. Instead, our PolyphonicFormer can clearly distinguish the boundary of the car and thus predicts a depth map with a clearer boundary. Then we measure the depth-aware capability of the PolyphonicFormer and compare it with the traditional dense depth estimation. Specifically, we verify the depth prediction accuracy of each instance. We consider each instance as \textbf{depth awareness} when less than a specific proportion ($10\%$) of pixels with absolute relative errors in depth estimation exceeds a certain threshold ($25\%$). The thresholds here follow the design of DQ. Simply combination of depth estimation and panoptic segmentation (similar structure as in~\cite{ViPDeepLab}) leads to 5,851 depth-aware instances out of 7,998 (73.2\%) instances in total, while the PolyphonicFormer achieves 6,119 out of 7,998 (76.5\%) results on the Cityscapes-DVPS validation set. The depth-aware capability of PolyphonicFormer is consistent with the visualization results and demonstrates the effectiveness of the unified framework for the DVPS task.

\noindent
\textbf{Limitation and Future Work.} 
As shown in Figure~\ref{fig:visualization_main}, the tracking head may fail when handling extreme occlusion cases (e.g. the car in the \textcolor[RGB]{220,0,0}{red} box). This is because PolyphonicFormer mainly performs tracking with appearance embeddings like the previous work. When applying PolyphonicFormer directly to safety-critical tasks like autonomous driving, the failure cases may cause accidents. The reliability needs further consideration and is potentially future work.

\section{Conclusion}
\label{sec:conclusion}
In this paper, we explore the relationship between semantics and geometry and build PolyphonicFormer for joint modeling panoptic segmentation and depth estimation. Our key insight is to unify all scene queries (thing, stuff) and depth queries into one framework. Based on our experiments, the PolyphonicFormer enhances the robustness towards panoptic segmentation as well as improves the performance of depth estimation. The proposed PolyphonicFormer achieves state-of-the-art results on both Cityscapes-DVPS and SemKITTI-DVPS datasets, and outperforms other methods on the SemKITTI-DVPS challenge. We hope PolyphonicFormer could serve as a good baseline in the DVPS task.

\noindent
\textbf{Acknowledgments.} This work was supported by the National Natural Science Foundation of China under Grants 62122060 and 62076188, and the Special Fund of Hubei Luojia Laboratory under Grant 220100014. The numerical calculations in this work had been supported by the supercomputing system in the Supercomputing Center of Wuhan University. This research is also supported by the National Key Research and Development Program of China under Grant No.2020YFB2103402.

\appendix
\begin{center}{
    \bf \Large Appendix
}
\end{center}
\vspace{-0.5em}

\setcounter{table}{0} 
\setcounter{figure}{0}
\setcounter{equation}{0}
\renewcommand{\thetable}{A\arabic{table}}
\renewcommand\thefigure{A\arabic{figure}} 
\renewcommand\theequation{A\arabic{equation}}

\begin{figure}[]
    \centering
    \includegraphics[width=1.\linewidth]{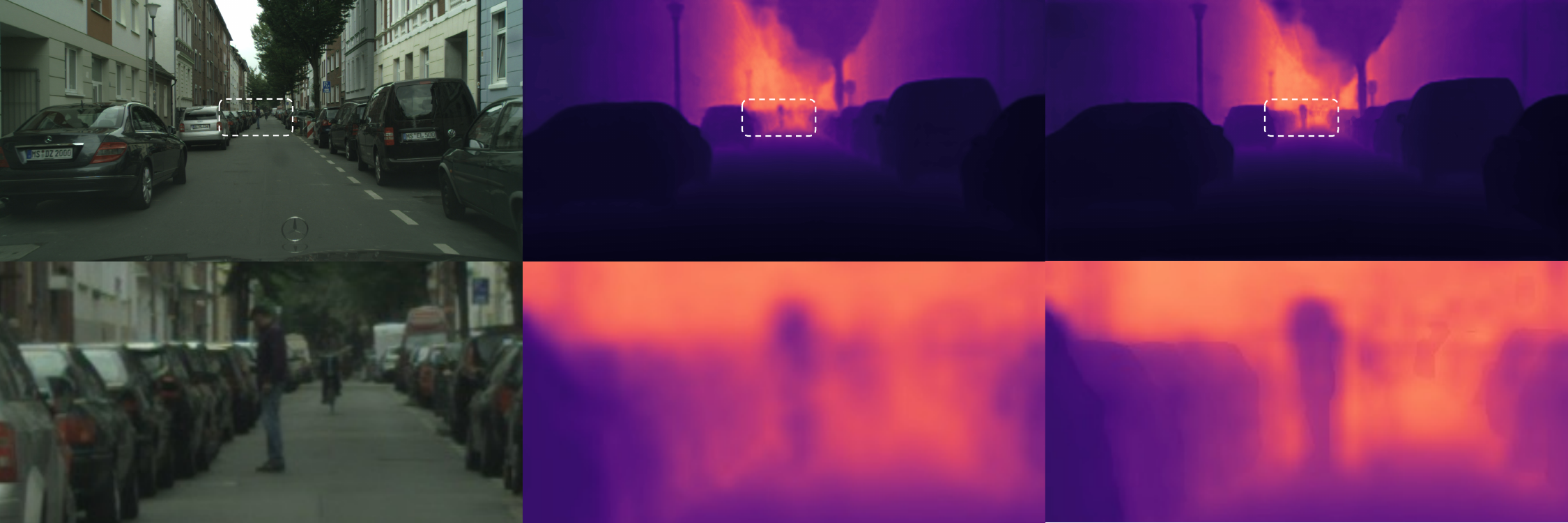}\\
    \vspace{0.25em}
    \includegraphics[width=1.\linewidth]{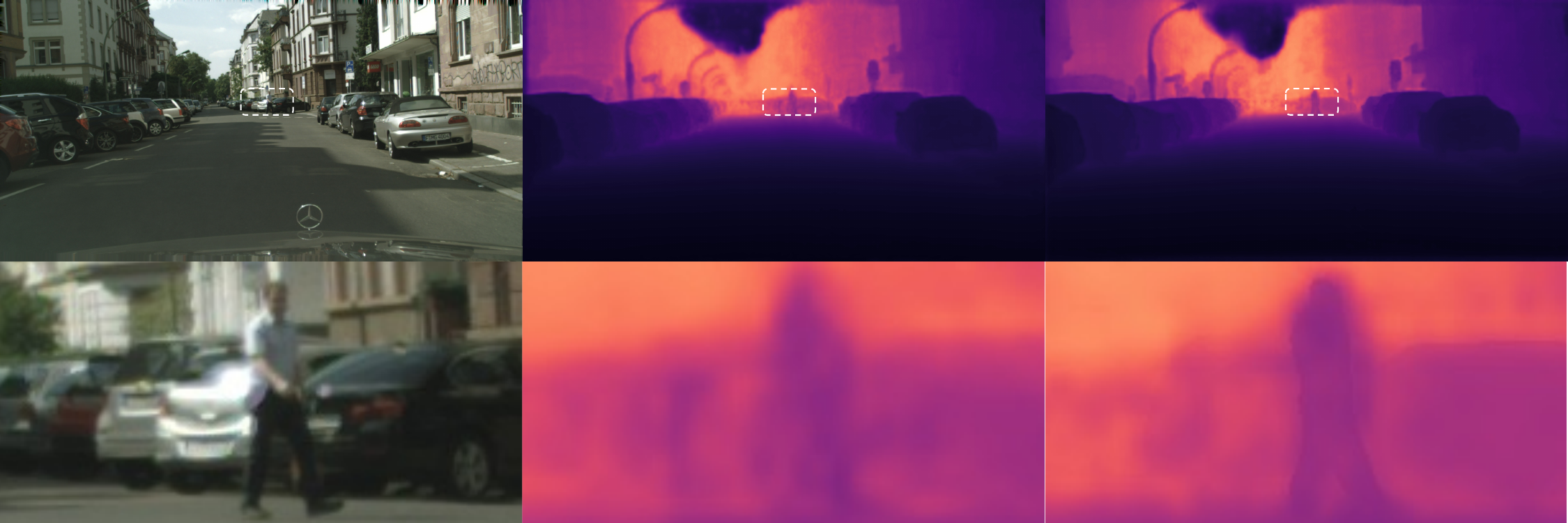}\\
    \caption{Comparison of depth predictions of traditional dense depth predictions and depth predictions of PolyphonicFormer. From left to right: input images, traditional dense depth predictions, and depth predictions of PolyphonicFormer.}
    \label{fig:vis_ablation_appendix}
    \vspace{-2em}
\end{figure}
\section{Implementation Details}
We report the implementation details in this section. To train the PolyphonicFormer, we first pre-train the backbone on ImageNet-1K~\cite{russakovsky2015imagenet} and the pre-train the panoptic path with Mapillary~\cite{neuhold2017mapillary} and Cityscapes~\cite{cordts2016cityscapes} datasets, following the ViP-Deeplab~\cite{ViPDeepLab}. When performing the pre-training on Mapillary, we resize the original images to a random scale from $2048 \times 1024$ to $4096 \times 2048$ and randomly crop a $1024 \times 1024$ sample. We do the Mapillary pre-training for 300 epochs. For Cityscapes pre-training, we also perform random resize from $2048 \times 1024$ to $4096 \times 2048$, but crop to $2048 \times 1024$. After pre-training, we train the image baseline of PolyphonicFormer on Cityscapes-DVPS. Training on Cityscapes-DVPS requires 192 epochs and takes the same data augmentation strategy as the Cityscapes dataset. The depth ground truth needs to be divided by the resize scale factor because resizing an image means zooming the image for depth perception. The ablation studies are performed on the image baseline. With the image baseline, we fine-tune the PolyphonicFormer with a tracking head on Cityscapes-DVPS and SemKITTI-DVPS respectively for 48 epochs. For each sample, we randomly choose a reference frame from time $t - 2$, $t - 1$, $t + 1$, and $t + 2$ for source frame at time $t$. We do not perform random scale resizing on SemKITTI-DVPS, and only pad the images to $1280 \times 384$ instead, which is the minimum size that can be divided by 32 to cover all of the KITTI images. All of the datasets we used are \textit{without} extra data with pseudo labels~\cite{chen2020naive} for self-supervised or semi-supervised training. During inference, for simplicity, we use single scale inference with the original image size from the datasets, and we do not use the test time online depth refinement~\cite{casser2019depth}. \textbf{In general}, except for that we do not use the \textit{test-time augmentation} and \textit{semi-supervised learning}, we adopt similar settings with ViP-Deeplab~\cite{ViPDeepLab}.

\begin{figure*}[!t]
\centering
\centering
  \includegraphics[width=\linewidth]{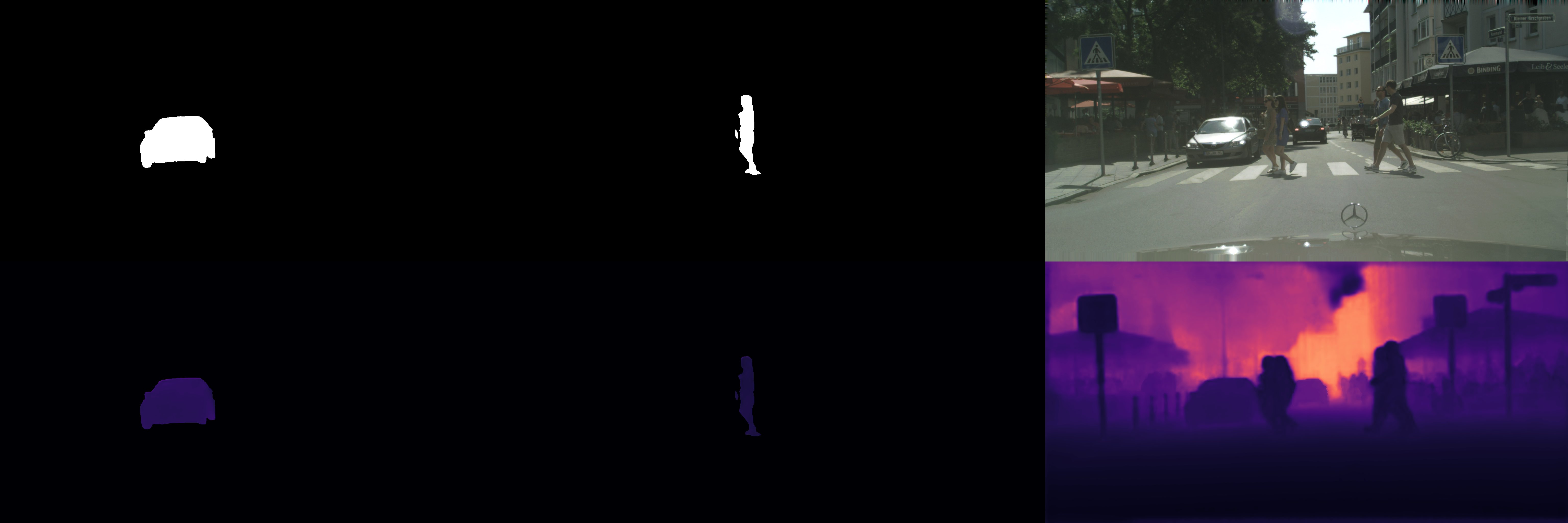}
  \caption{We present several videos \href{https://github.com/HarborYuan/PolyphonicFormer}{here}. The left and middle are mask and depth results of tracked instance examples output by PolyphonicFormer. The depth results are merged into the final prediction (bottom right). }
  \label{fig:video_demo}
  \vspace{-0.5em}
\end{figure*}

For the ICCV-2021 SemKITTI-DVPS Challenge submission, we take advantage of the validation set for training. Before and after adding the extra validation samples, PolyphonicFormer can achieve 63.6 and 64.6 DSTQ respectively in the SemKITTI-DVPS test set.

\section{More Visualization analyses}
We show more visualization analysis results in Figure~\ref{fig:vis_ablation_appendix}. The depth predictions of PolyphonicFormer are merged from depth predictions for each thing or stuff mask. As shown in Figure~\ref{fig:vis_ablation_appendix}, the final depth predictions of PolyphonicFormer successfully distinguish the boundary between the instances or instances and corresponding background and thus are more accurate than the dense prediction results. We note that the results are presented with a high resolution, so we recommend the readers zoom in to check the details about the depth results of other instances.

We also illustrate the unified query learning with a video, as shown in Figure~\ref{fig:video_demo}.  The PolyphonicFormer generates temporal-consistent instance-level mask and depth predictions and merges them into the final results.

\begin{table*}[t!]
\renewcommand{\arraystretch}{0.80}
\small
\centering
\resizebox{\textwidth}{!}{
    \begin{tabular}{l|c|c|c|c|c|c}
    \toprule[0.2em]
    $\text{DVPQ}_\lambda^k$ on Cityscapes-DVPS & k = 1 & k = 2 & k = 3 & k = 4 & Average & FLOPs\\
    \toprule[0.2em]
    PolyphonicFormer $\lambda$ = 0.50 & 64.3 $\vert$ 56.0 $\vert$ 70.3 & 57.1 $\vert$ 43.1 $\vert$ 67.2 & 54.0 $\vert$ 37.0 $\vert$ 66.3 & 52.3 $\vert$ 34.3 $\vert$ 65.3 & 56.9 $\vert$ 42.6 $\vert$ 67.3 & - \\
    PolyphonicFormer $\lambda$ = 0.25 & 59.7 $\vert$ 53.3 $\vert$ 64.4 & 53.0 $\vert$ 41.3 $\vert$ 61.5 & 49.9 $\vert$ 35.3 $\vert$ 60.5 & 48.6 $\vert$ 33.0 $\vert$ 60.0 & 52.8 $\vert$ 40.7 $\vert$ 61.6 & - \\
    PolyphonicFormer $\lambda$ = 0.10 & 39.3 $\vert$ 31.8 $\vert$ 44.7 & 34.3 $\vert$ 23.3 $\vert$ 42.3 & 32.7 $\vert$ 20.3 $\vert$ 41.7 & 31.5 $\vert$ 18.6 $\vert$ 40.8 & 34.5 $\vert$ 23.5 $\vert$ 42.4 & - \\
    Average: PolyphonicFormer & 54.4 $\vert$ 47.0 $\vert$ 59.8 & 48.1 $\vert$ 35.9 $\vert$ 57.0 & 45.5 $\vert$ 30.9 $\vert$ 56.2 & 44.1 $\vert$ 28.6 $\vert$ 55.4 &  \textbf{48.1} $\vert$ 35.6 $\vert$ 57.1 & 411G \\
    
    \bottomrule[0.1em]
    \toprule[0.2em]
    $\text{DVPQ}_\lambda^k$ on SemKITTI-DVPS & k = 1 & k = 5 & k = 10 & k = 20 & Average & FLOPs\\
    \toprule[0.2em]
    PolyphonicFormer $\lambda$ = 0.50 & 50.5 $\vert$ 44.0 $\vert$ 55.3 & 45.7 $\vert$ 34.8 $\vert$ 53.7 & 44.4 $\vert$ 32.4 $\vert$ 53.1 &   43.7 $\vert$ 31.4 $\vert$ 52.7 & 46.1 $\vert$ 35.7 $\vert$ 53.7 & - \\
    PolyphonicFormer $\lambda$ = 0.25 & 47.9 $\vert$ 42.2 $\vert$ 52.1 & 43.2 $\vert$ 33.3 $\vert$ 50.4 & 42.0 $\vert$ 31.1 $\vert$ 49.9 & 41.3 $\vert$ 30.3 $\vert$ 49.4 & 43.6 $\vert$ 34.2 $\vert$ 50.5 & - \\
    PolyphonicFormer $\lambda$ = 0.10 & 35.9 $\vert$ 33.6 $\vert$ 37.6 & 31.2 $\vert$ 25.2 $\vert$ 35.5 & 29.6 $\vert$ 22.9 $\vert$ 34.5 & 28.5 $\vert$ 21.5 $\vert$ 33.6 & 31.3 $\vert$ 25.8 $\vert$ 35.3 & - \\
    Average: PolyphonicFormer & 44.8 $\vert$ 39.9 $\vert$ 48.3 & 40.0 $\vert$ 31.1 $\vert$ 46.5 & 38.7 $\vert$ 28.8 $\vert$ 45.8 & 37.8 $\vert$ 27.7 $\vert$ 45.2 & \textbf{40.3} $\vert$ 31.9 $\vert$ 46.5 & 99G \\
    \bottomrule[0.1em]
    \end{tabular}
    }
    \caption{\small Experimental results on Cityscapes-DVPS and SemKITTI-DVPS datasets with Resnet-50 backbone. Each cell shows $\text{DVPQ}_{\lambda}^{k} ~\vert~ \text{DVPQ}_{\lambda}^{k}\text{-Thing} ~\vert~ \text{DVPQ}_{\lambda}^{k}\text{-Stuff}$ where $\lambda$ is the threshold of relative depth error, and $k$ is the number of frames. Smaller $\lambda$ and larger $k$ correspond to a higher accuracy requirement.  We also estimate the computational cost (FLOPs) of ViP-Deeplab with Resnet-50 backbone and get 1,096G and 280G on Cityscapes-DVPS and SemKITTI-DVPS respectively.
    }
    \label{tab:dvps_results_r50}
    \vspace{-0.5em}
\end{table*}

\definecolor{orange}{rgb}{0.910,0.631,0.580}
\definecolor{blue}{rgb}{0.580,0.733,0.910}
\begin{table*}[t!]
\normalsize
\begin{center}
\resizebox{\textwidth}{!}{
\begin{tabular}{|l||c|c|c|c|c|c|c|}
\hline
method & \cellcolor{orange}abs rel & \cellcolor{orange}sq rel & \cellcolor{orange}RMSE & \cellcolor{orange}RMSE log & \cellcolor{blue}$\sigma < 1.25$ & \cellcolor{blue}$\sigma < 1.25^2$ & \cellcolor{blue}$\sigma < 1.25^3$ \\
\hline
DPT-Hybrid~\cite{ranftl2021vision}& 0.0697 & 0.4515 & 4.115 & 0.1106 & 0.9434 & 0.9914 & 0.9976\\
\hline
PolyphonicFormer & \textbf{0.0647} & \textbf{0.3454} & \textbf{3.800} &  \textbf{0.1013} & \textbf{0.9524} & \textbf{0.9950} & \textbf{0.9985}\\
\hline
\end{tabular}
}
\caption{\small Comparison results of PolyphonicFormer and the representative depth estimation method. The metrics with \textcolor{orange}{orange} background means "\textbf{lower} is better". The metrics with \textcolor{blue}{blue} background means "\textbf{higher} is better". ViP-Deeplab~\cite{ViPDeepLab} has a \textbf{0.0721} abs rel.}
\label{tab:comparision_with_dpt}
\end{center}
\vspace{-2em}
\end{table*}

\section{More Experiments}
We report more results of PolyphonicFormer in this section. The results of PolyphonicFormer with Swin-B backbone are already provided, and we report the DVPQ results with Resnet-50 backbone in this section. As shown in Table~\ref{tab:dvps_results_r50}, with a Resnet-50 backbone, the PolyphonicFormer achieves \textbf{48.1}, and \textbf{40.3} in DVPQ on the Cityscapes-DVPS and SemKITTI-DVPS datasets, respectively. 

We compare PolyphonicFormer with recently proposed DPT~\cite{ranftl2021vision}, which is one of the state-of-the-art supervised monocular depth estimation methods on KITTI (eigen split)~\cite{eigen2015predicting}. As the KITTI dataset lacks the panoptic segmentation annotation and the SemanticKITTI dataset has a very different split strategy compared with eigen split, we cannot directly get the results on the KITTI eigen split. We adopt the pre-trained model of DPT-Hybrid on MIX6~\cite{ranftl2021vision} (meta-datasets containing 10 datasets) and KITTI eigen split, and fine-tune on Cityscapes-DVPS with the same schedule of PolyphonicFormer. As in Table~\ref{tab:comparision_with_dpt}, our proposed PolyphonicFormer outperforms the DPT-Hybrid~\cite{ranftl2021vision} and ViP-Deeplab~\cite{ViPDeepLab}.

\section{More Visualization Results}
\begin{figure*}[!t]
\centering

\begin{subfigure}{.332\linewidth}
  \centering
  \includegraphics[trim={0 150 0 150},clip,width=\linewidth]{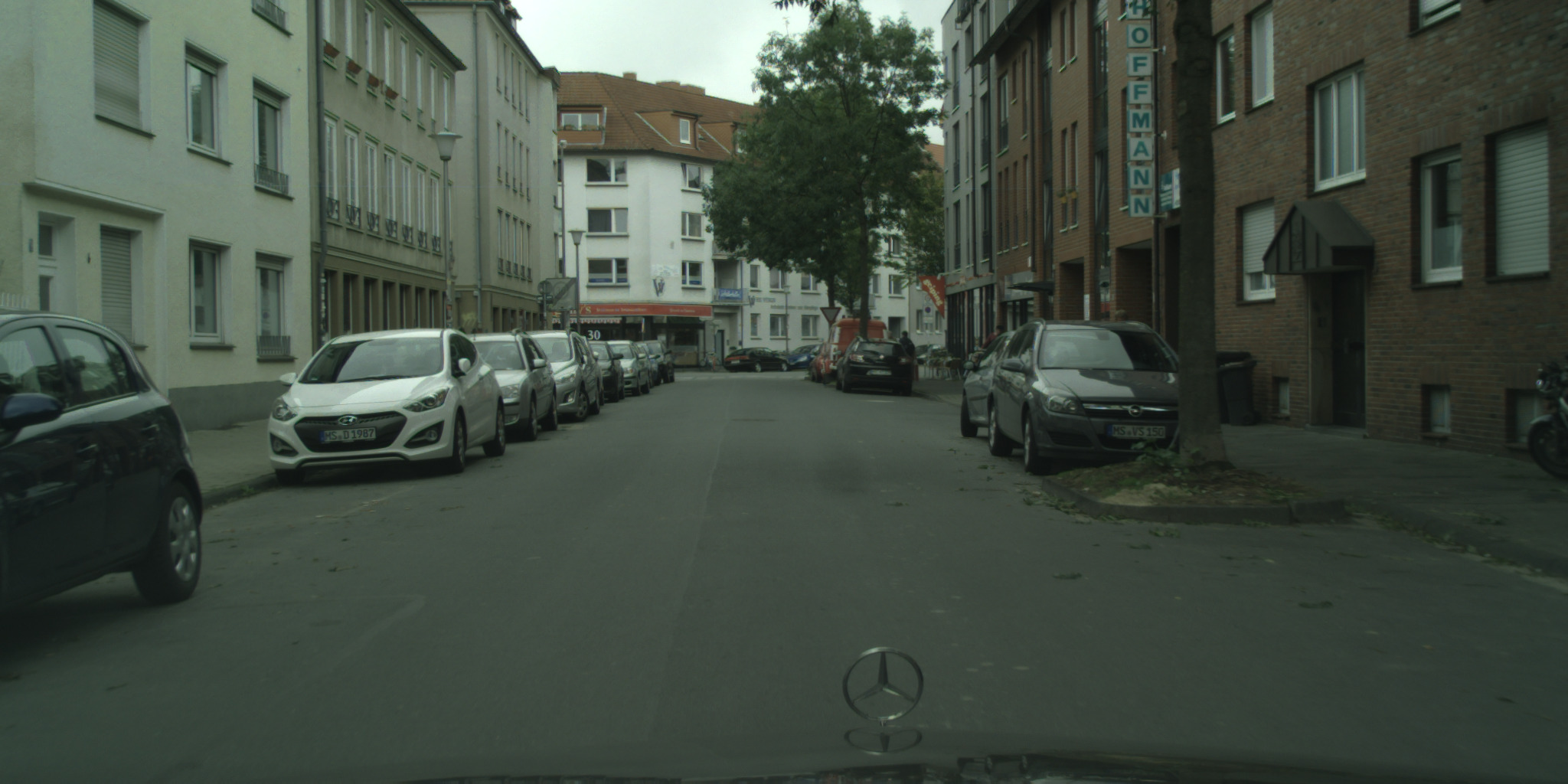}
\end{subfigure}\hfill
\begin{subfigure}{.332\linewidth}
  \centering
  \includegraphics[trim={0 150 0 150},clip,width=\linewidth]{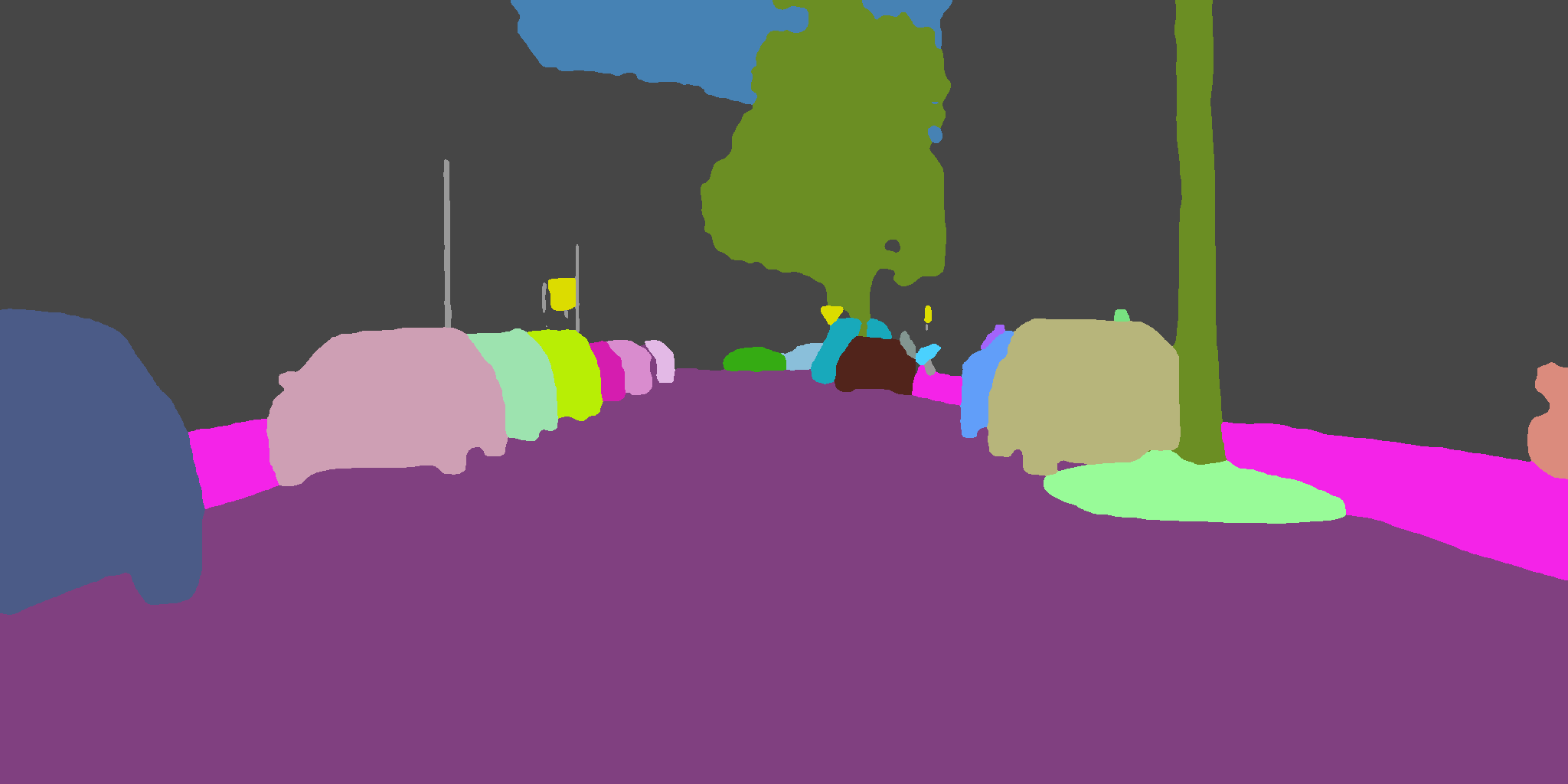}
\end{subfigure}\hfill
\begin{subfigure}{.332\linewidth}
  \centering
  \includegraphics[trim={0 150 0 150},clip,width=\linewidth]{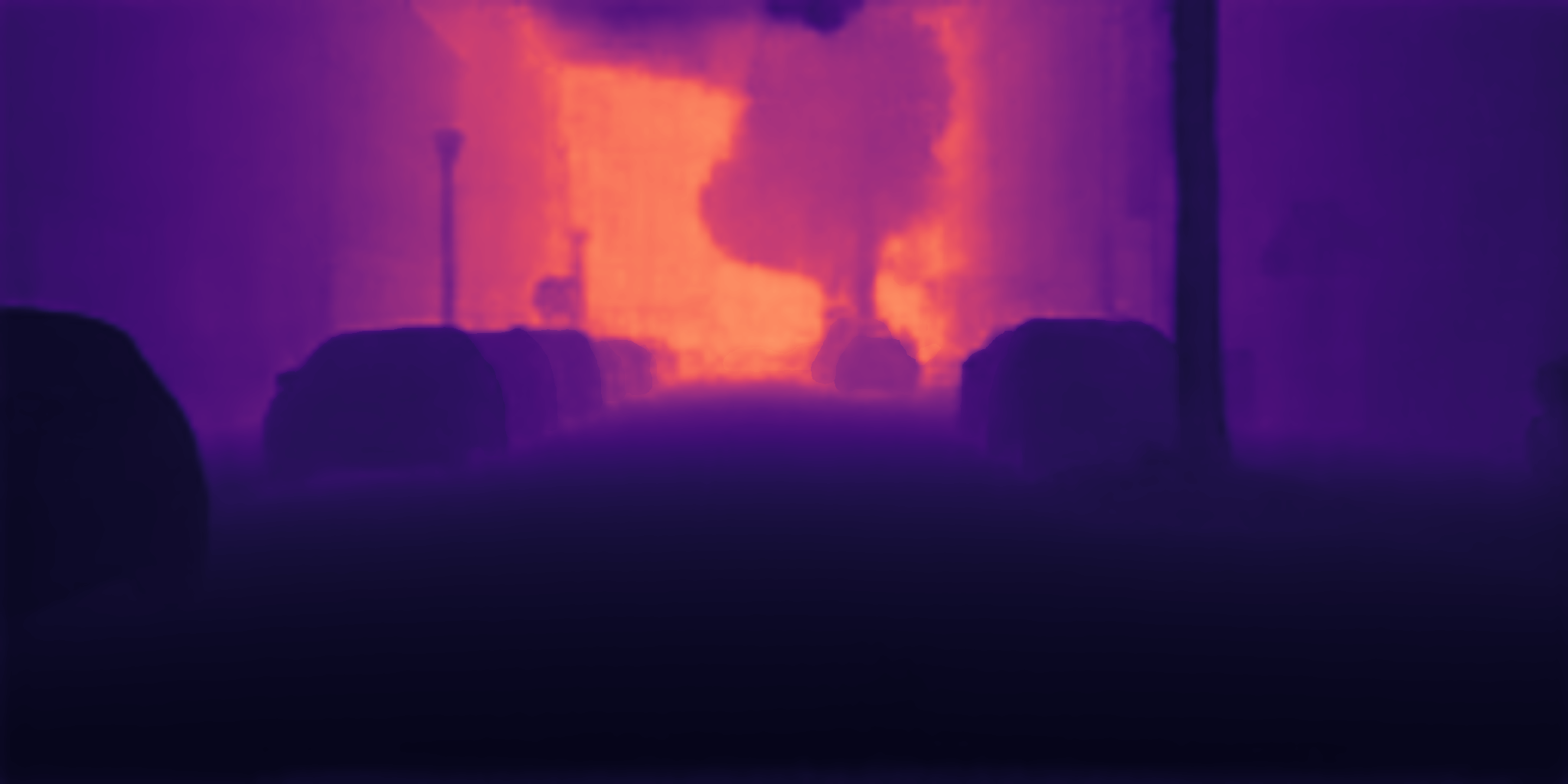}
\end{subfigure}\\

\begin{subfigure}{.332\linewidth}
  \centering
  \includegraphics[trim={0 150 0 150},clip,width=\linewidth]{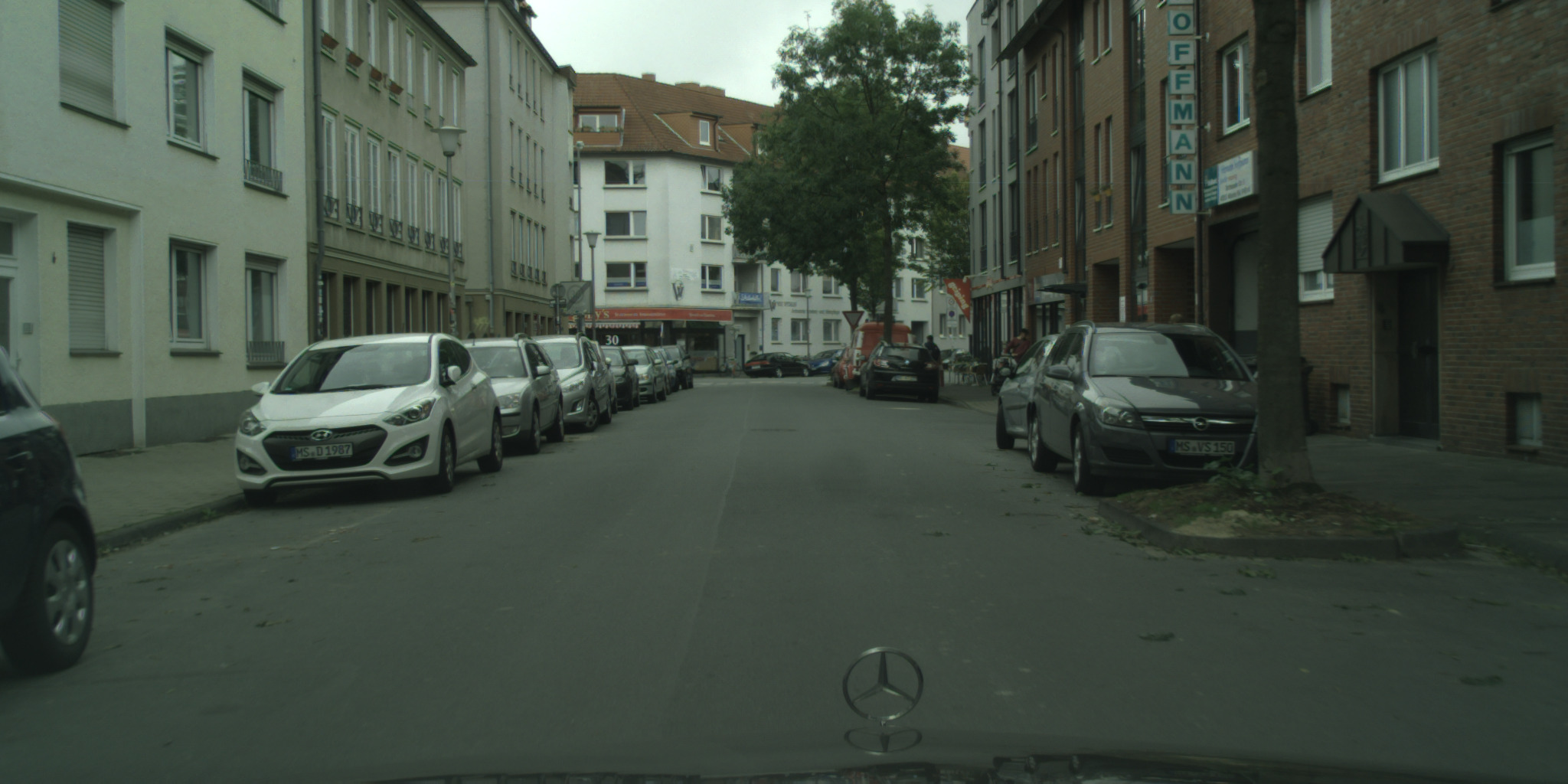}
\end{subfigure}\hfill
\begin{subfigure}{.332\linewidth}
  \centering
  \includegraphics[trim={0 150 0 150},clip,width=\linewidth]{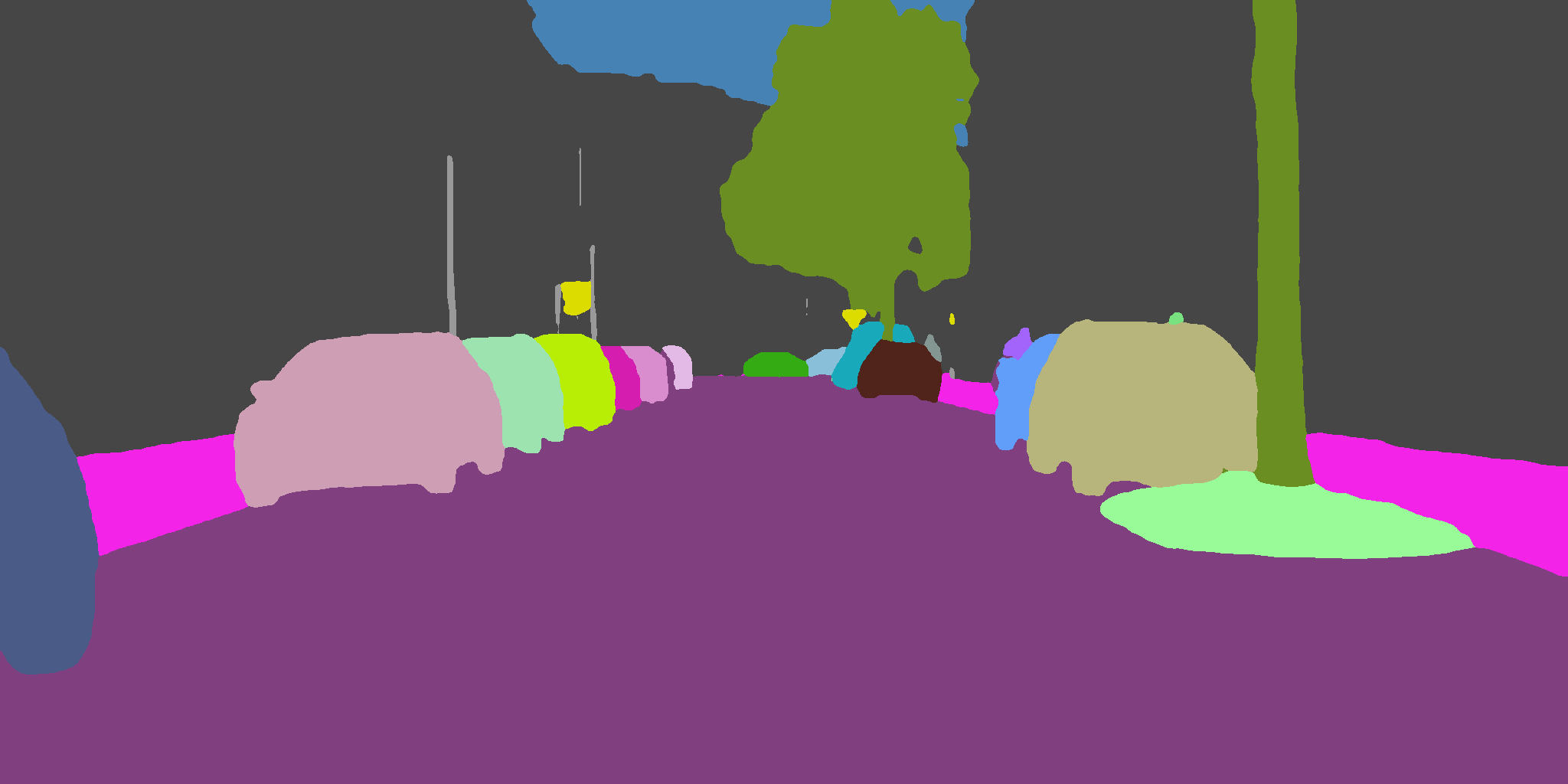}
\end{subfigure}\hfill
\begin{subfigure}{.332\linewidth}
  \centering
  \includegraphics[trim={0 150 0 150},clip,width=\linewidth]{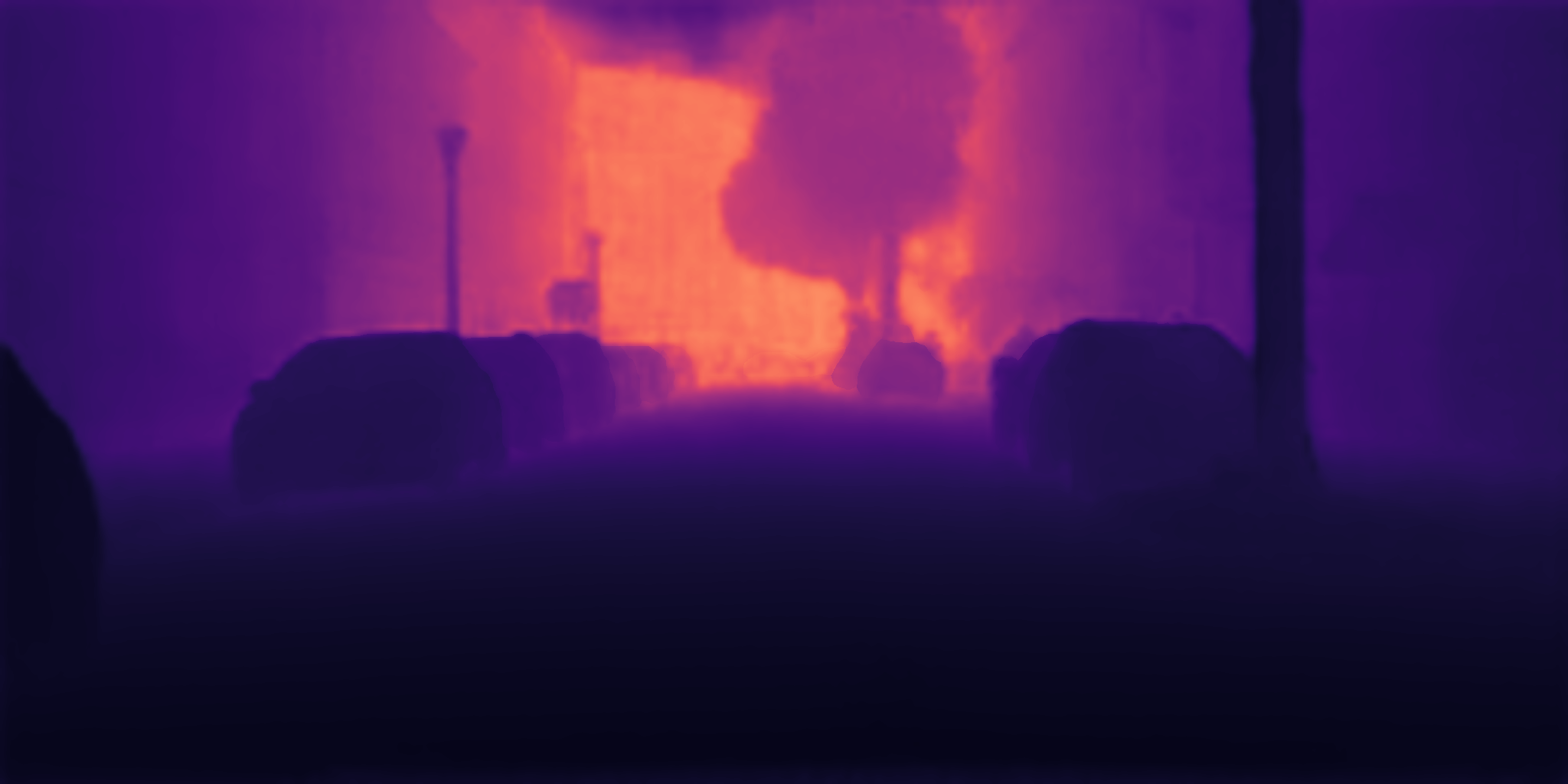}
\end{subfigure}\\

\begin{subfigure}{.332\linewidth}
  \centering
  \includegraphics[trim={0 150 0 150},clip,width=\linewidth]{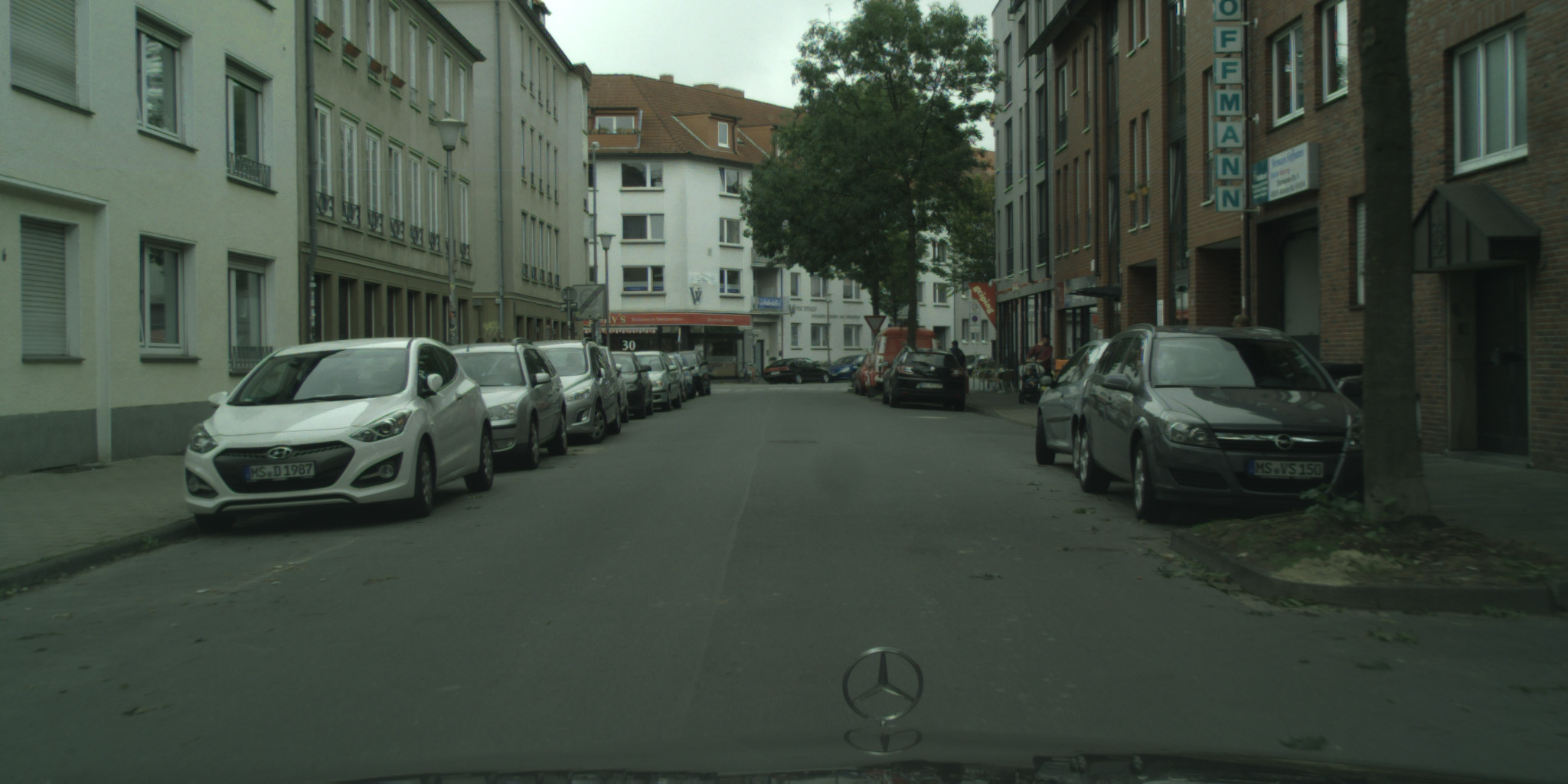}
\end{subfigure}\hfill
\begin{subfigure}{.332\linewidth}
  \centering
  \includegraphics[trim={0 150 0 150},clip,width=\linewidth]{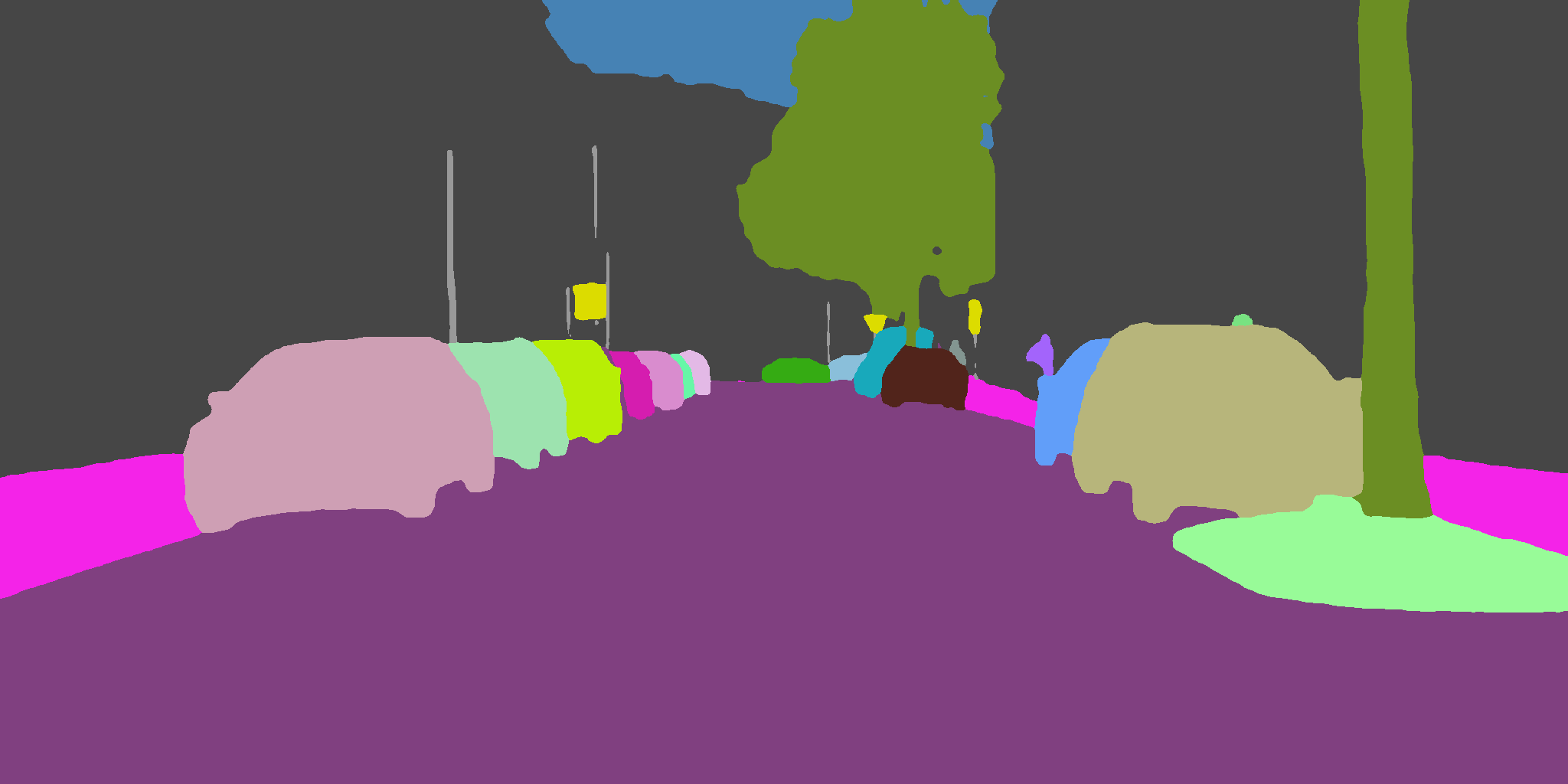}
\end{subfigure}\hfill
\begin{subfigure}{.332\linewidth}
  \centering
  \includegraphics[trim={0 150 0 150},clip,width=\linewidth]{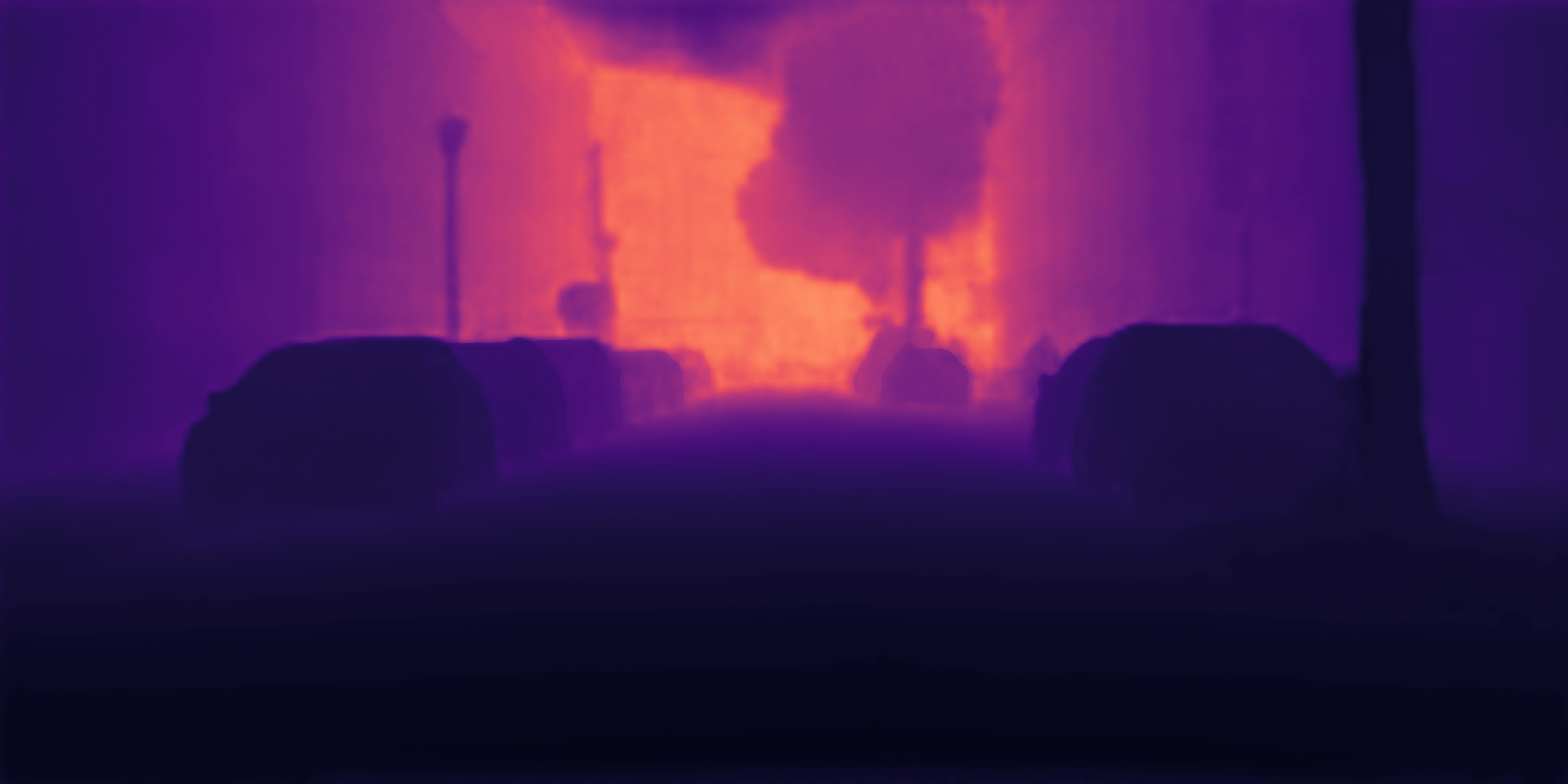}
\end{subfigure}\\

\begin{subfigure}{.332\linewidth}
  \centering
  \includegraphics[trim={0 150 0 150},clip,width=\linewidth]{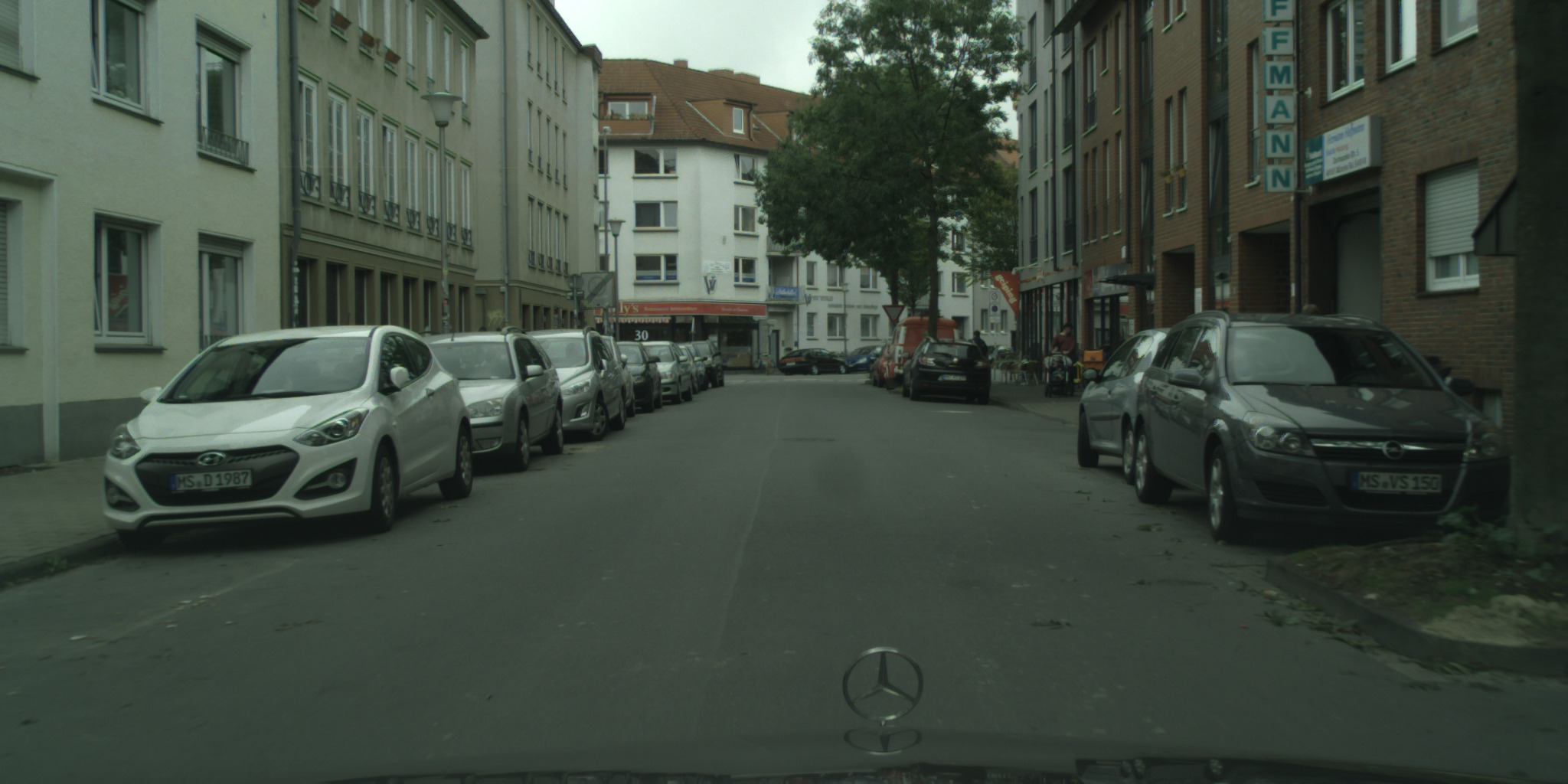}
\end{subfigure}\hfill
\begin{subfigure}{.332\linewidth}
  \centering
  \includegraphics[trim={0 150 0 150},clip,width=\linewidth]{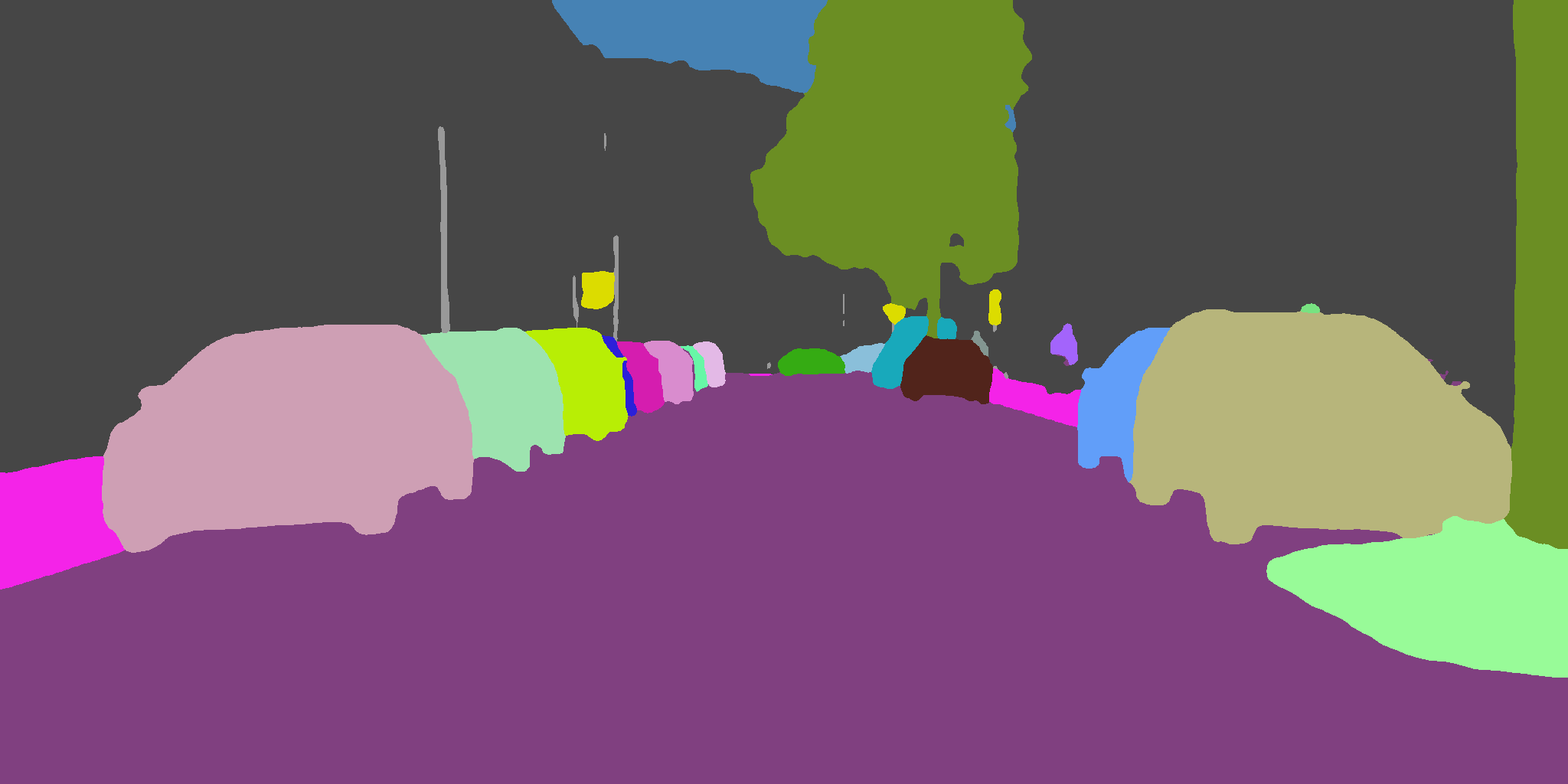}
\end{subfigure}\hfill
\begin{subfigure}{.332\linewidth}
  \centering
  \includegraphics[trim={0 150 0 150},clip,width=\linewidth]{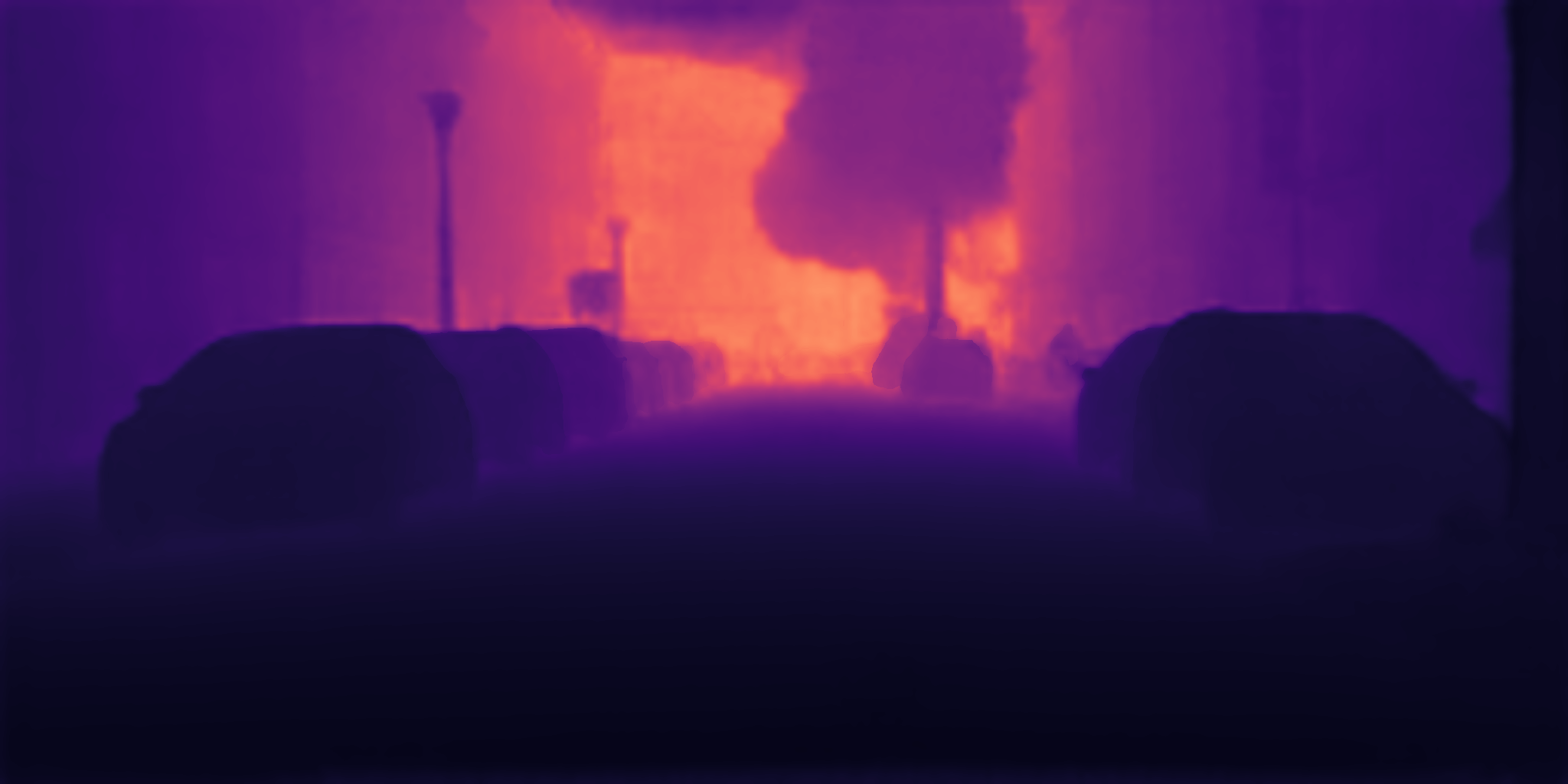}
\end{subfigure}\\

\begin{subfigure}{.332\linewidth}
  \centering
  \includegraphics[trim={0 150 0 150},clip,width=\linewidth]{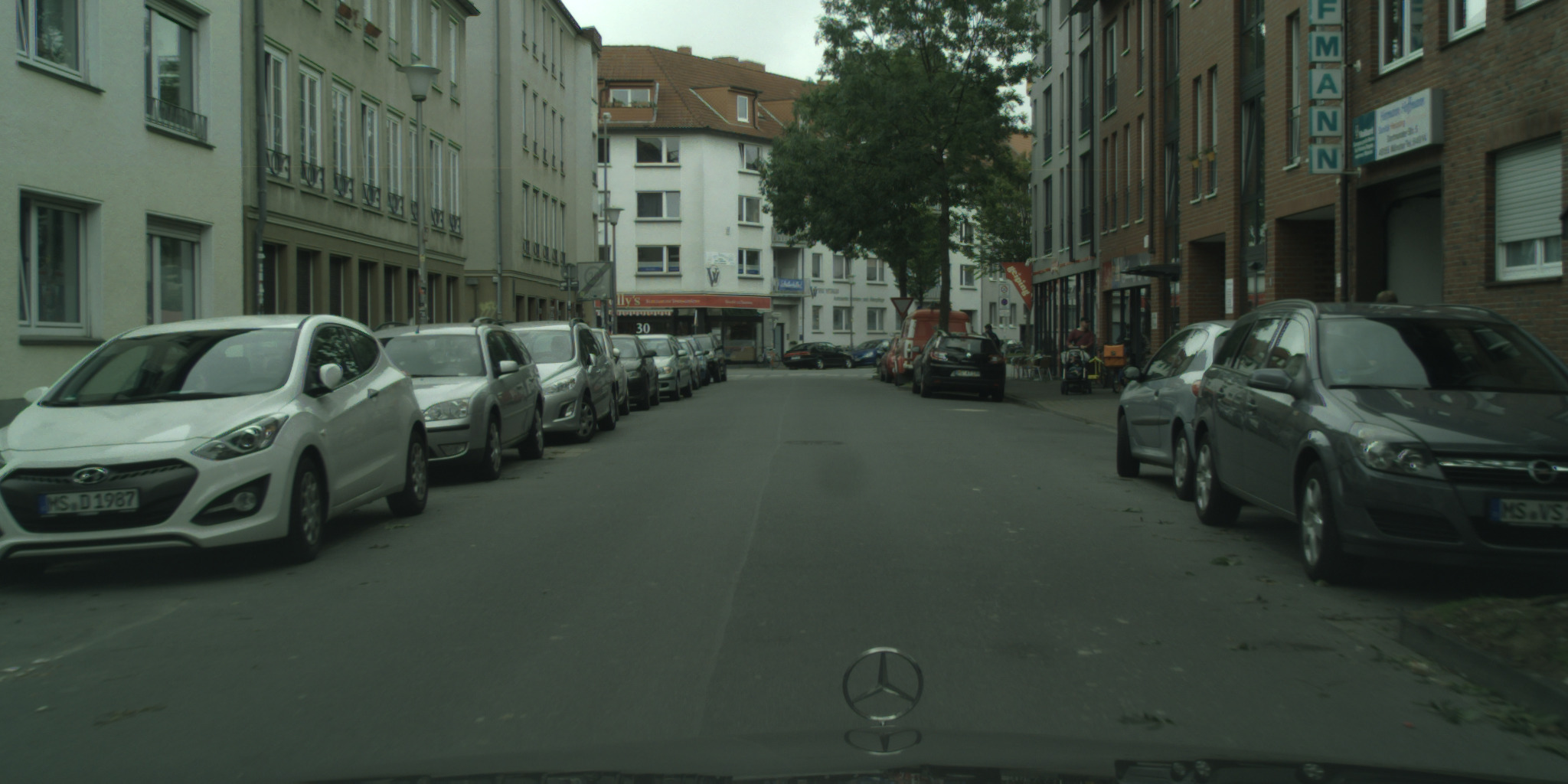}
\end{subfigure}\hfill
\begin{subfigure}{.332\linewidth}
  \centering
  \includegraphics[trim={0 150 0 150},clip,width=\linewidth]{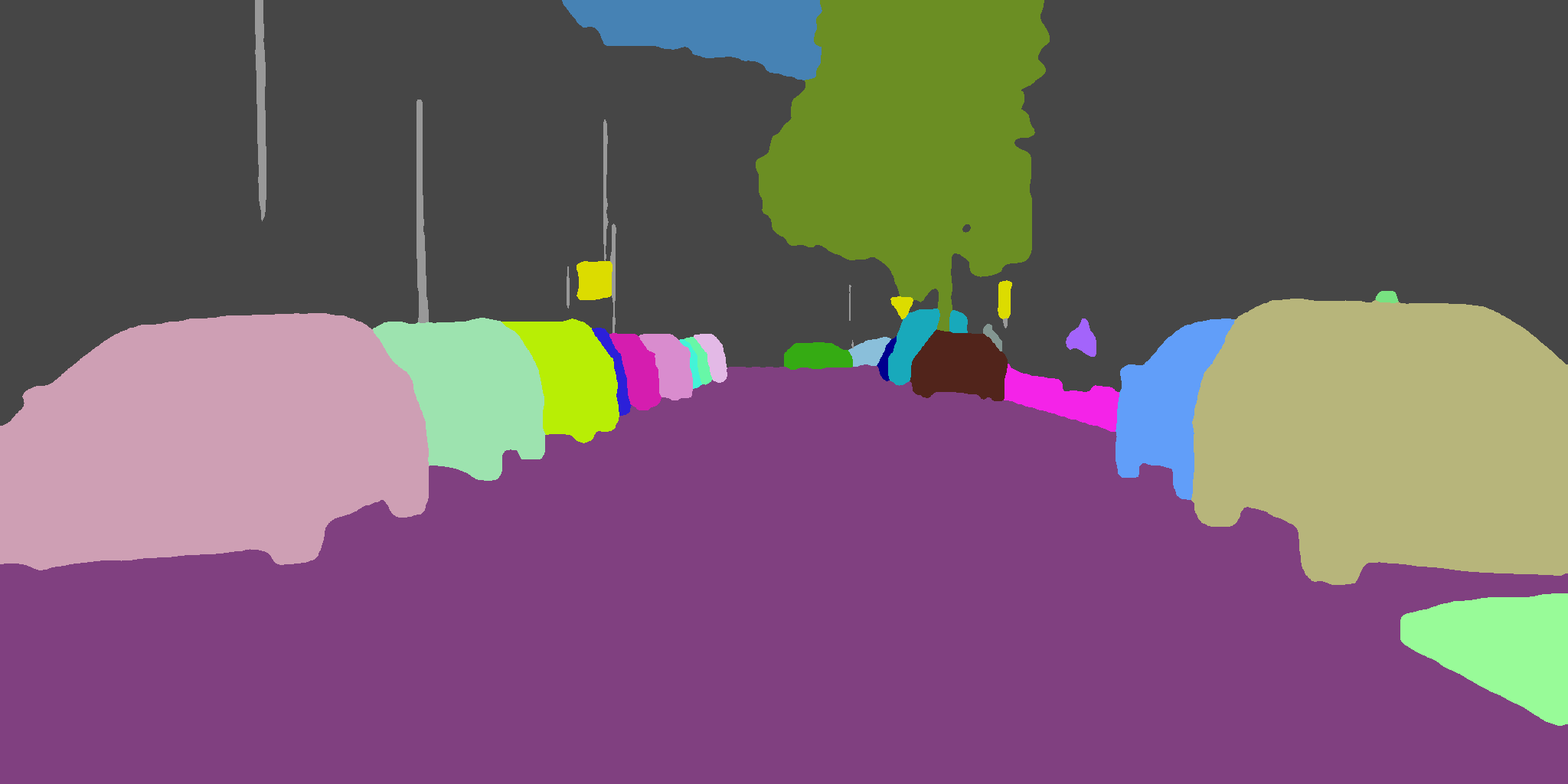}
\end{subfigure}\hfill
\begin{subfigure}{.332\linewidth}
  \centering
  \includegraphics[trim={0 150 0 150},clip,width=\linewidth]{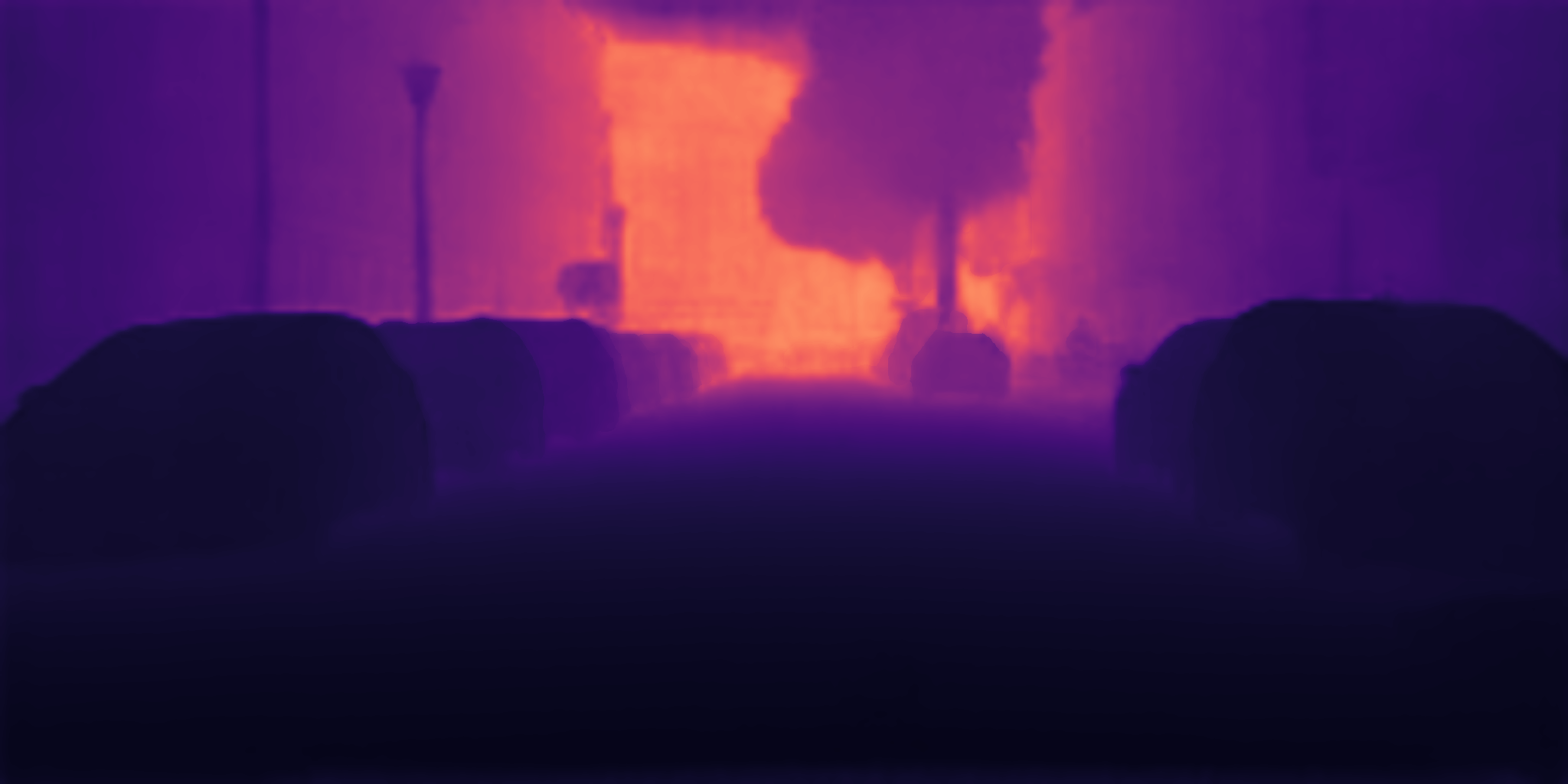}
\end{subfigure}\\

\begin{subfigure}{.332\linewidth}
  \centering
  \includegraphics[trim={0 150 0 150},clip,width=\linewidth]{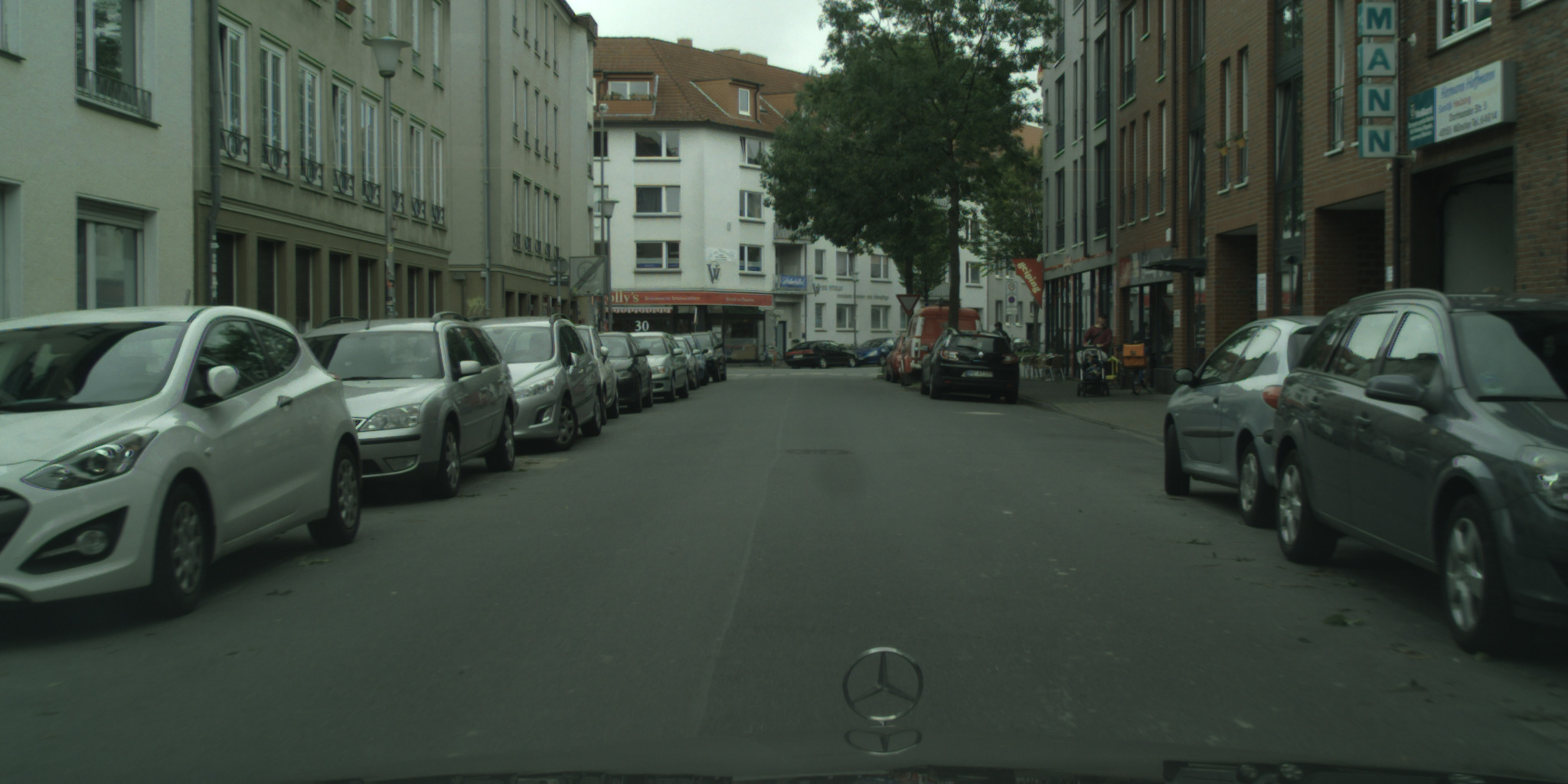}
\end{subfigure}\hfill
\begin{subfigure}{.332\linewidth}
  \centering
  \includegraphics[trim={0 150 0 150},clip,width=\linewidth]{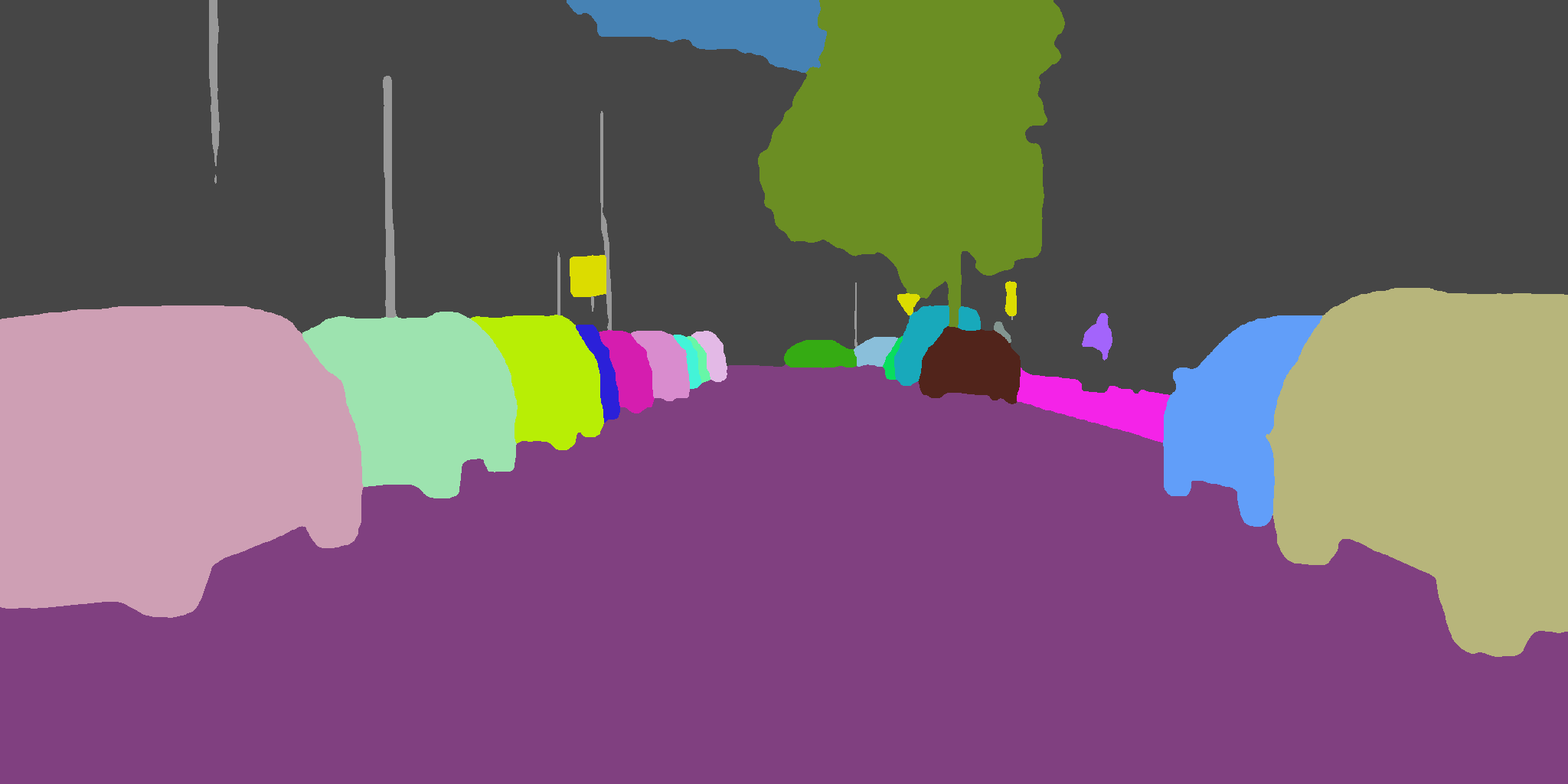}
\end{subfigure}\hfill
\begin{subfigure}{.332\linewidth}
  \centering
  \includegraphics[trim={0 150 0 150},clip,width=\linewidth]{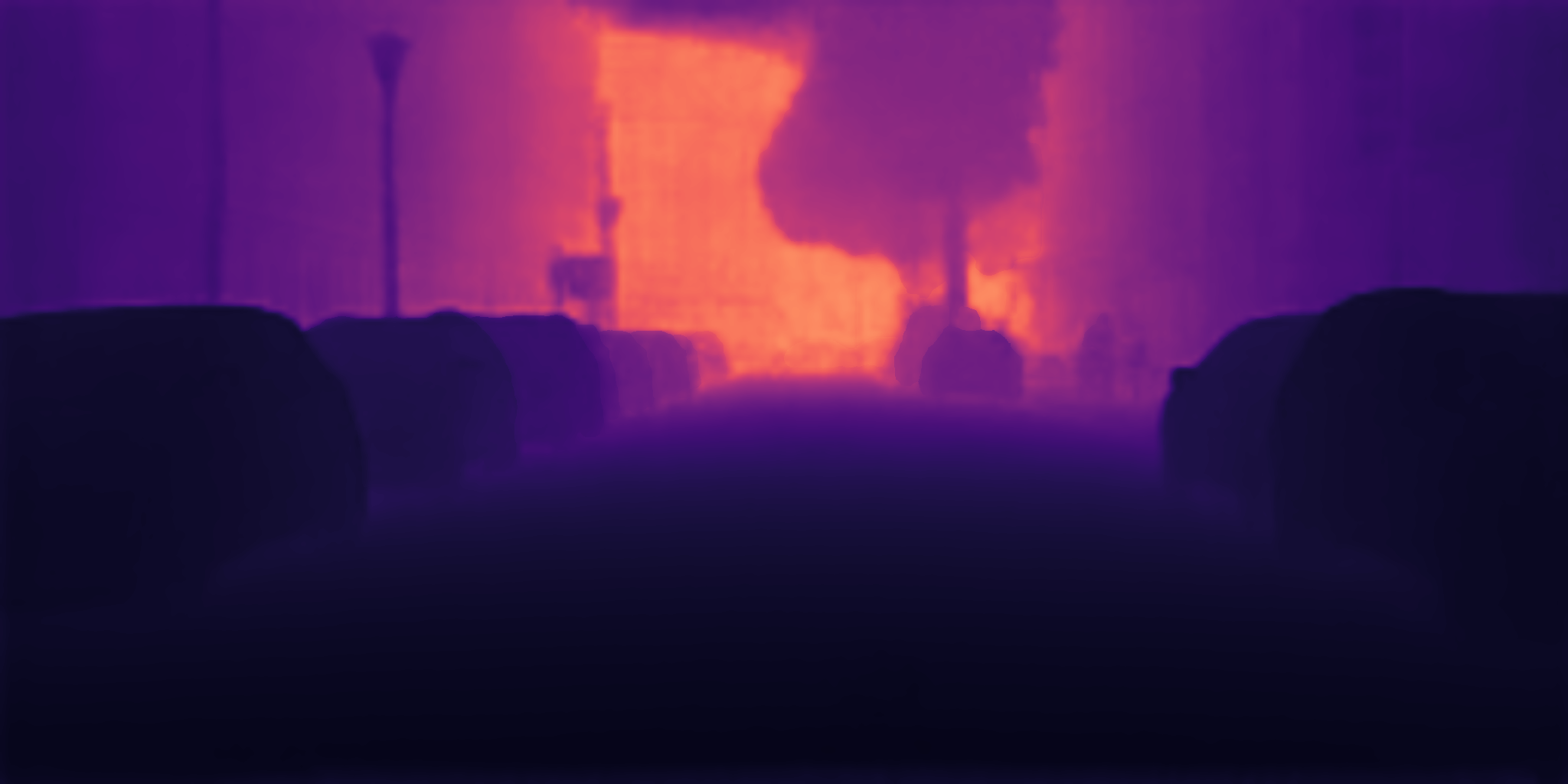}
\end{subfigure}\\

\begin{subfigure}{.332\linewidth}
  \centering
  \includegraphics[trim={0 150 0 150},clip,width=\linewidth]{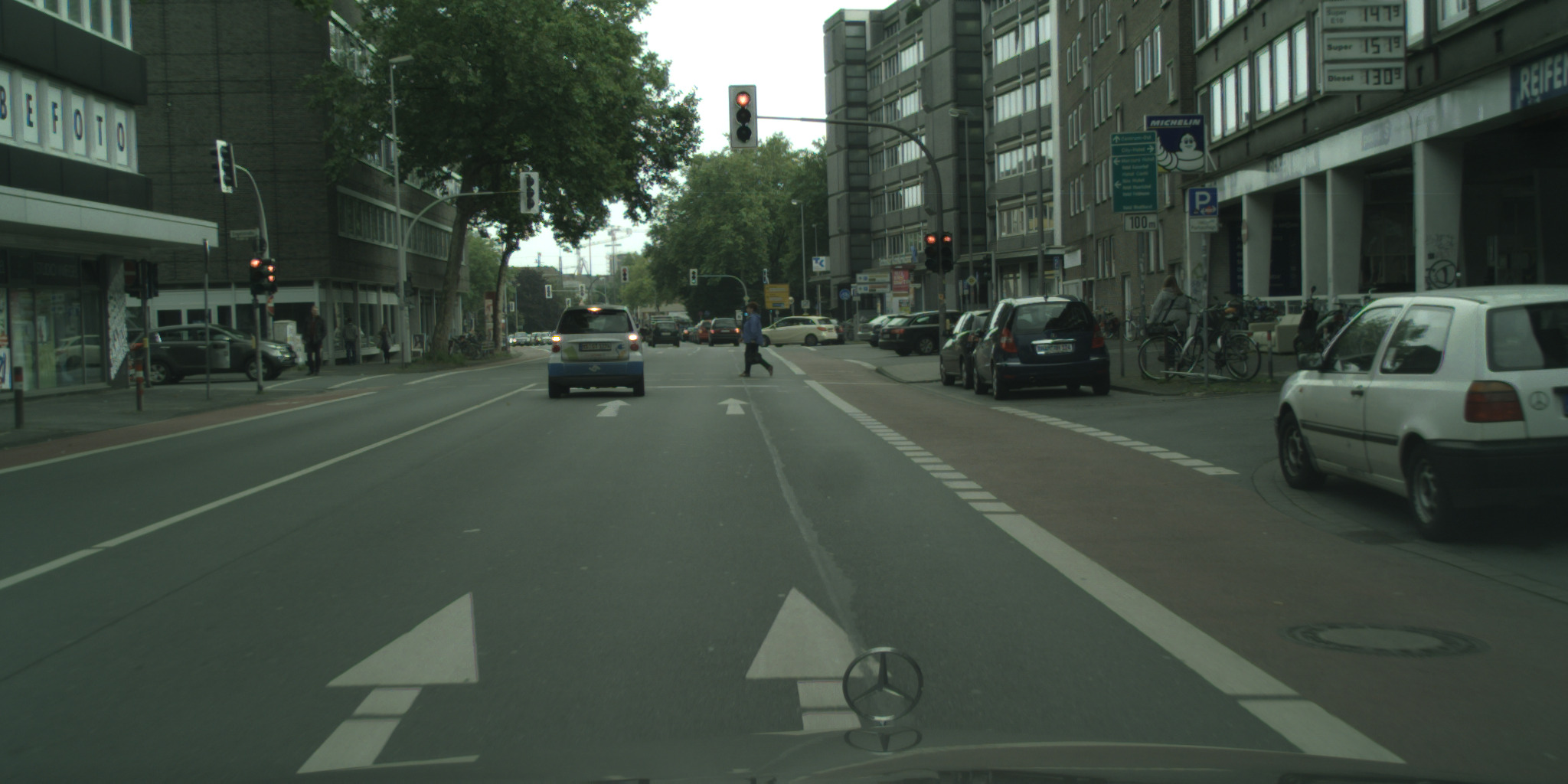}
\end{subfigure}\hfill
\begin{subfigure}{.332\linewidth}
  \centering
  \includegraphics[trim={0 150 0 150},clip,width=\linewidth]{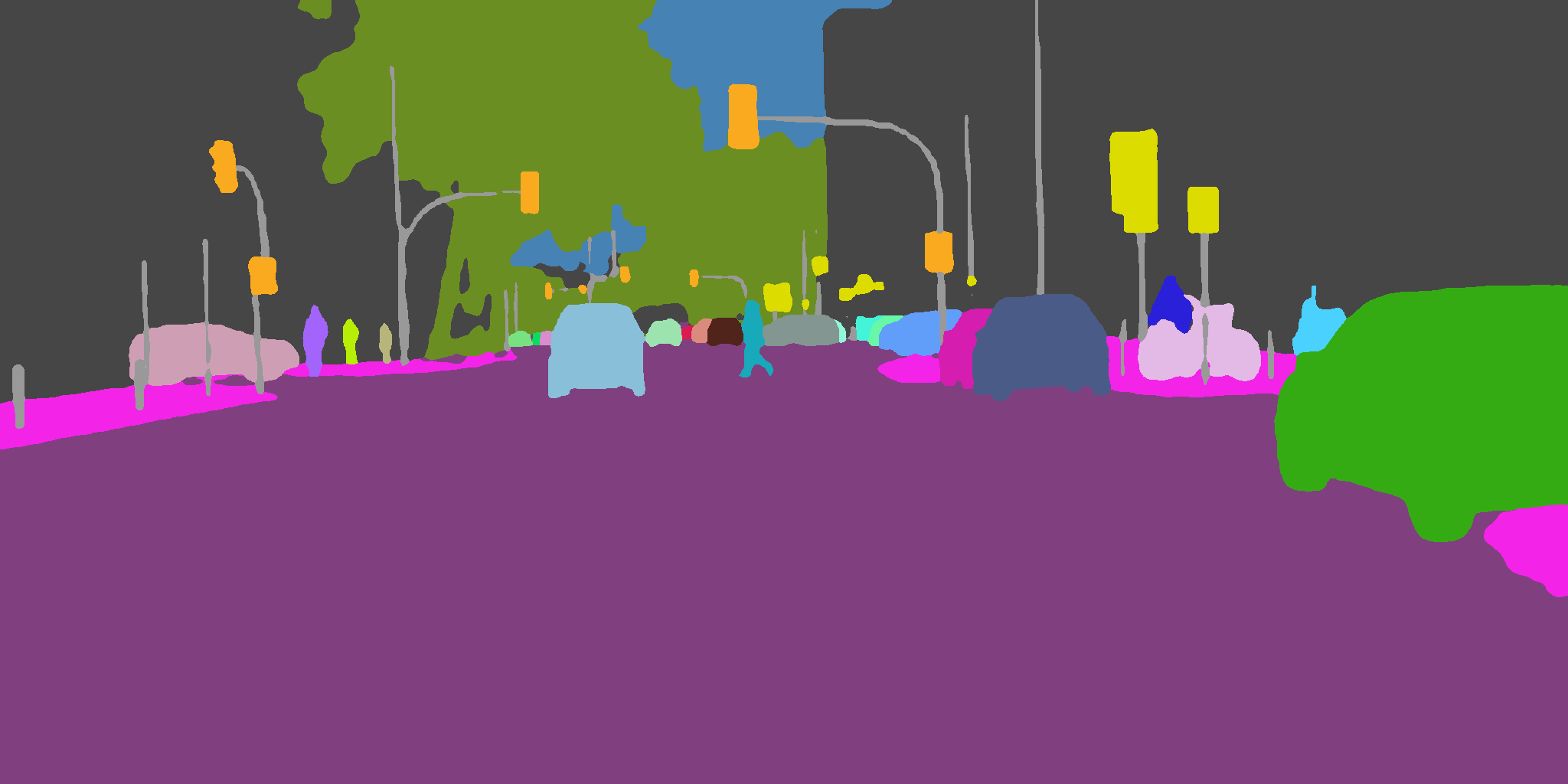}
\end{subfigure}\hfill
\begin{subfigure}{.332\linewidth}
  \centering
  \includegraphics[trim={0 150 0 150},clip,width=\linewidth]{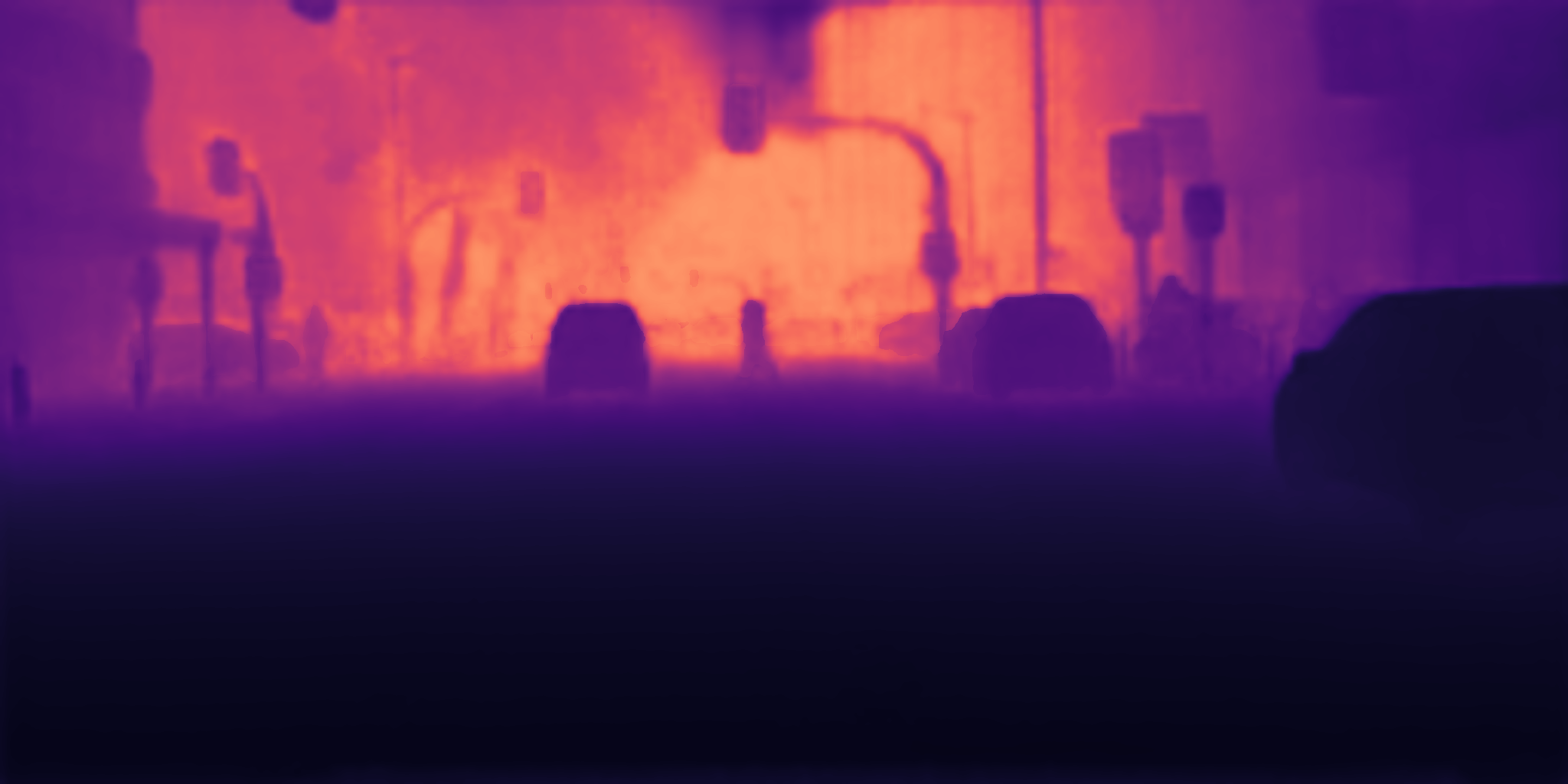}
\end{subfigure}\\

\begin{subfigure}{.332\linewidth}
  \centering
  \includegraphics[trim={0 150 0 150},clip,width=\linewidth]{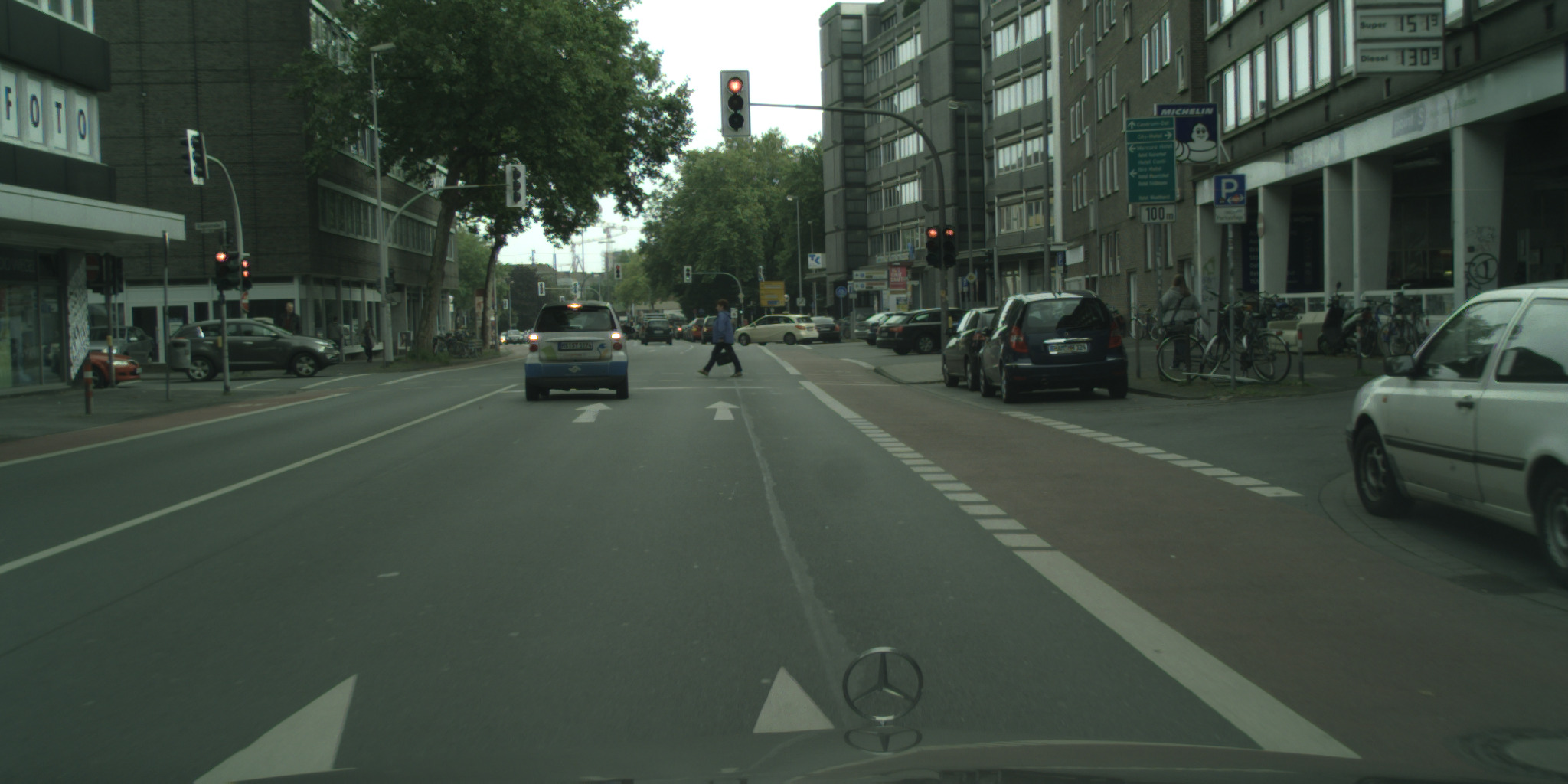}
\end{subfigure}\hfill
\begin{subfigure}{.332\linewidth}
  \centering
  \includegraphics[trim={0 150 0 150},clip,width=\linewidth]{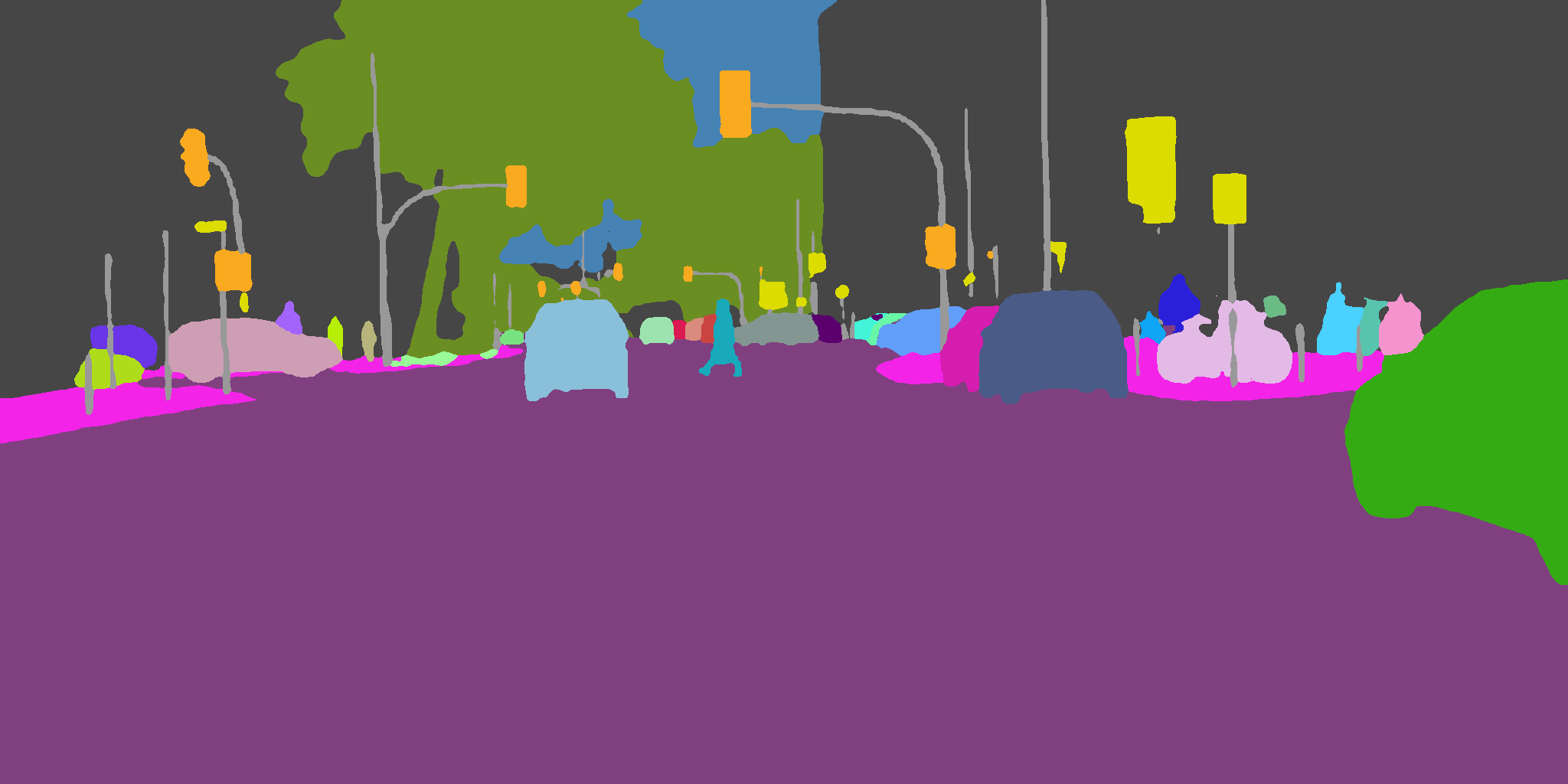}
\end{subfigure}\hfill
\begin{subfigure}{.332\linewidth}
  \centering
  \includegraphics[trim={0 150 0 150},clip,width=\linewidth]{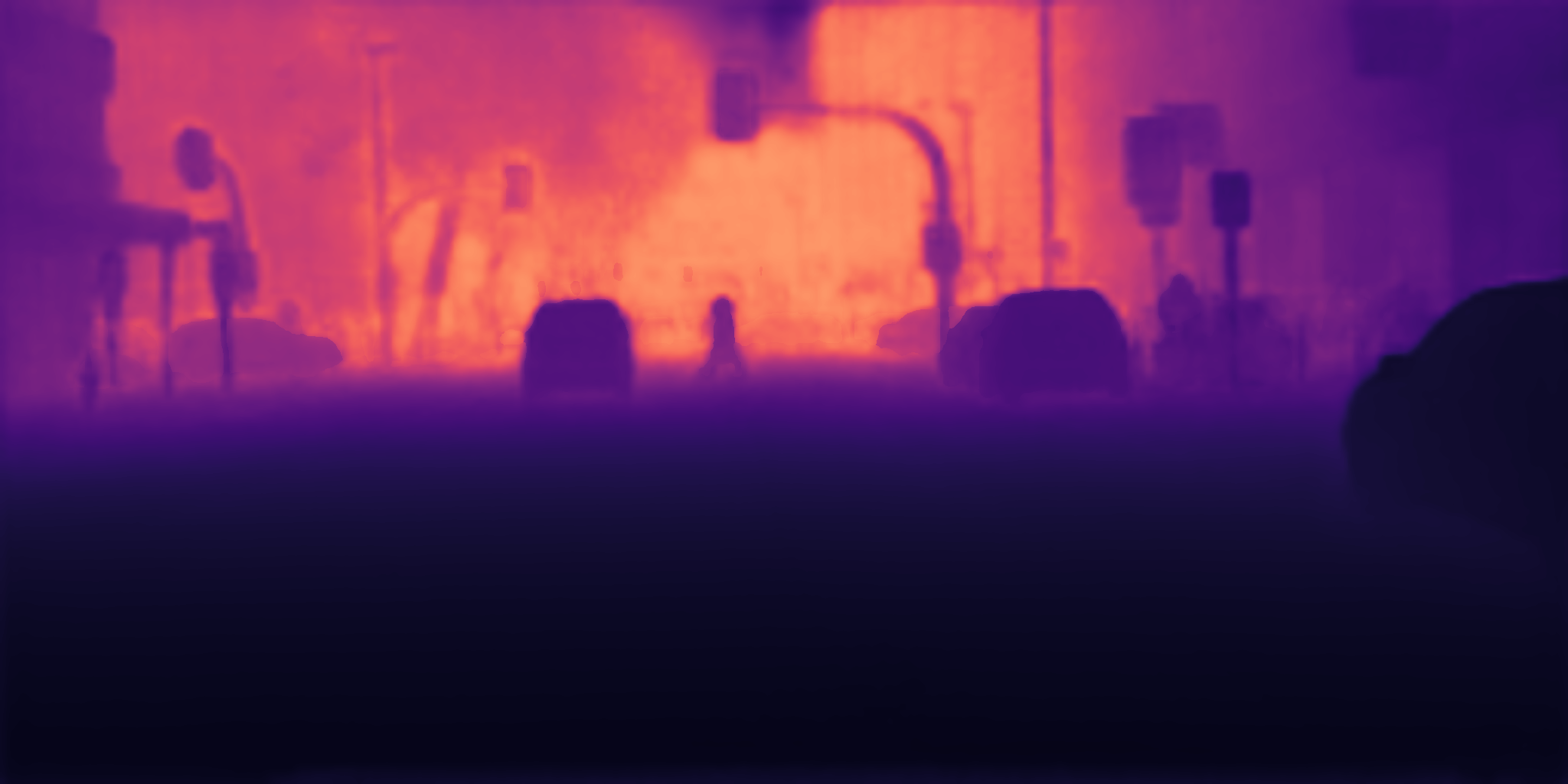}
\end{subfigure}\\

\begin{subfigure}{.332\linewidth}
  \centering
  \includegraphics[trim={0 150 0 150},clip,width=\linewidth]{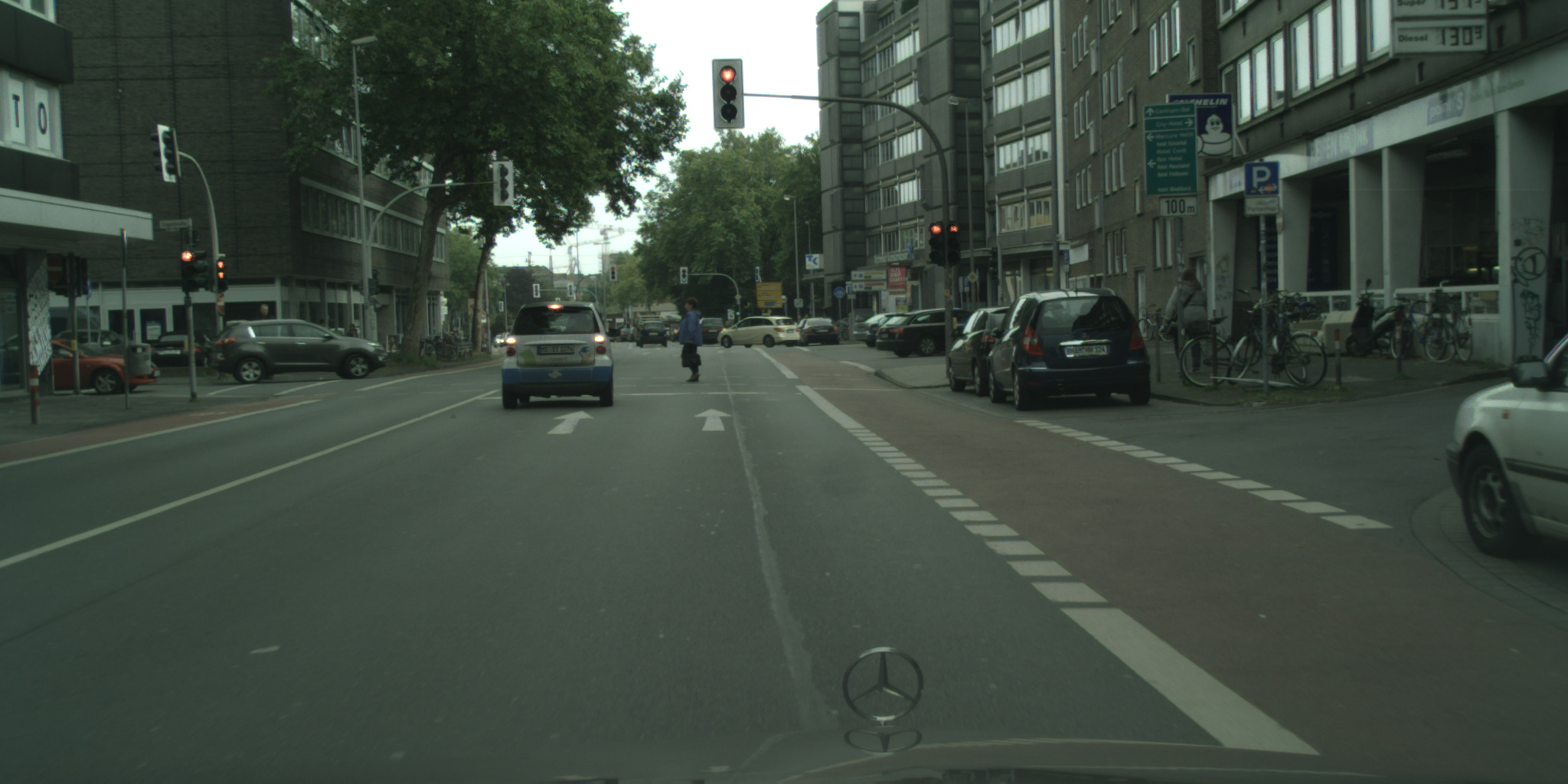}
\end{subfigure}\hfill
\begin{subfigure}{.332\linewidth}
  \centering
  \includegraphics[trim={0 150 0 150},clip,width=\linewidth]{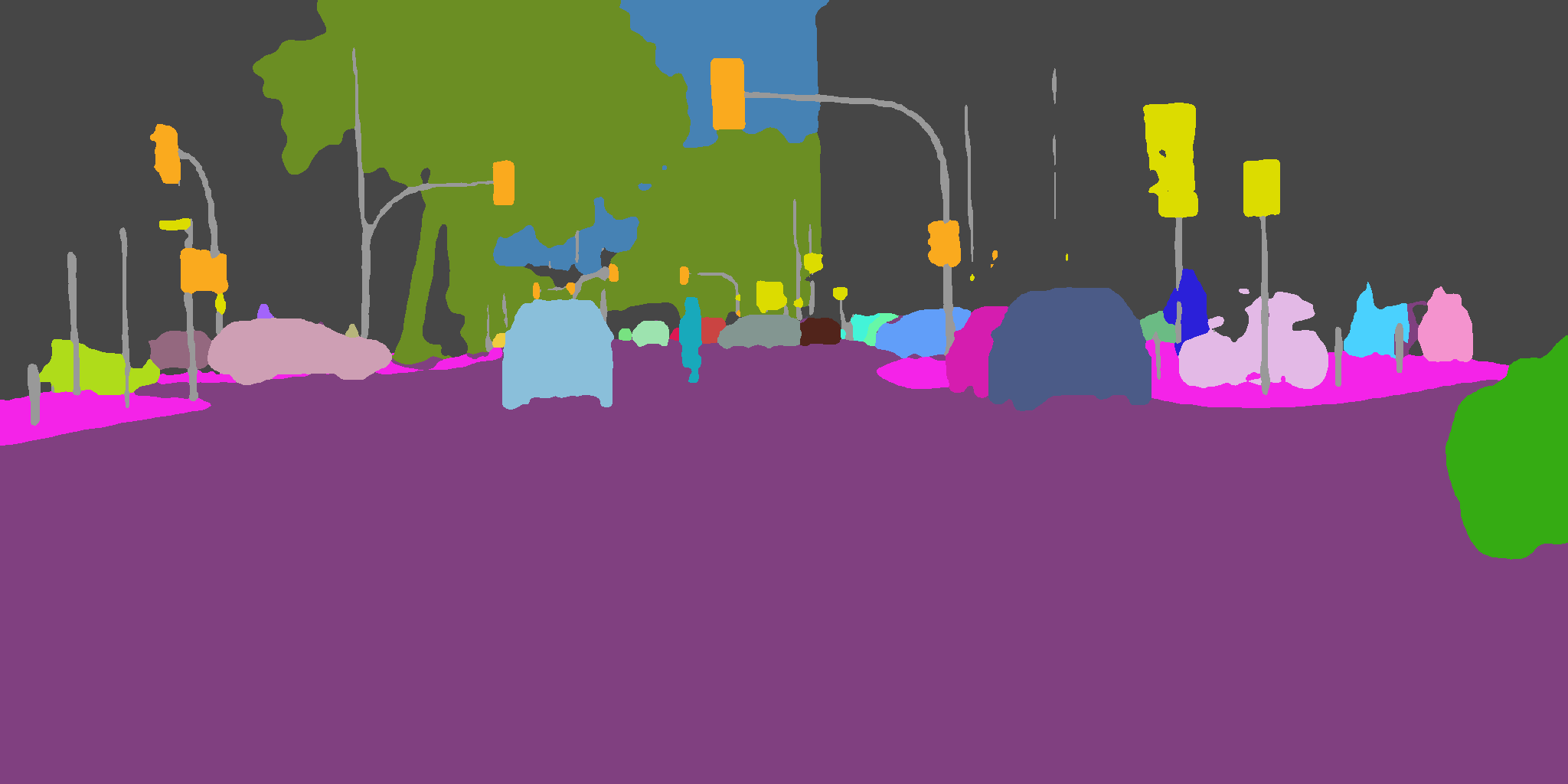}
\end{subfigure}\hfill
\begin{subfigure}{.332\linewidth}
  \centering
  \includegraphics[trim={0 150 0 150},clip,width=\linewidth]{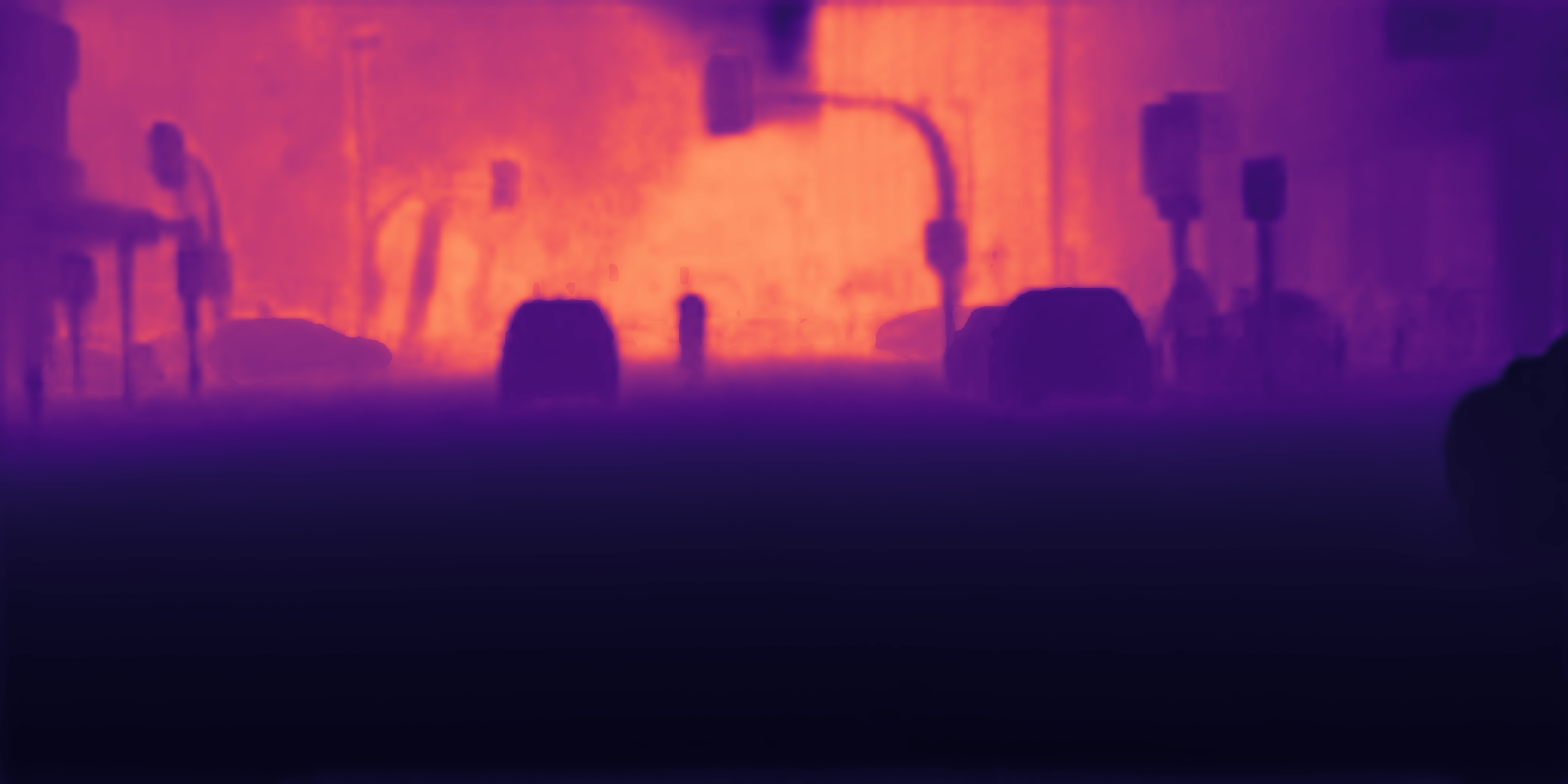}
\end{subfigure}\\

\begin{subfigure}{.332\linewidth}
  \centering
  \includegraphics[trim={0 150 0 150},clip,width=\linewidth]{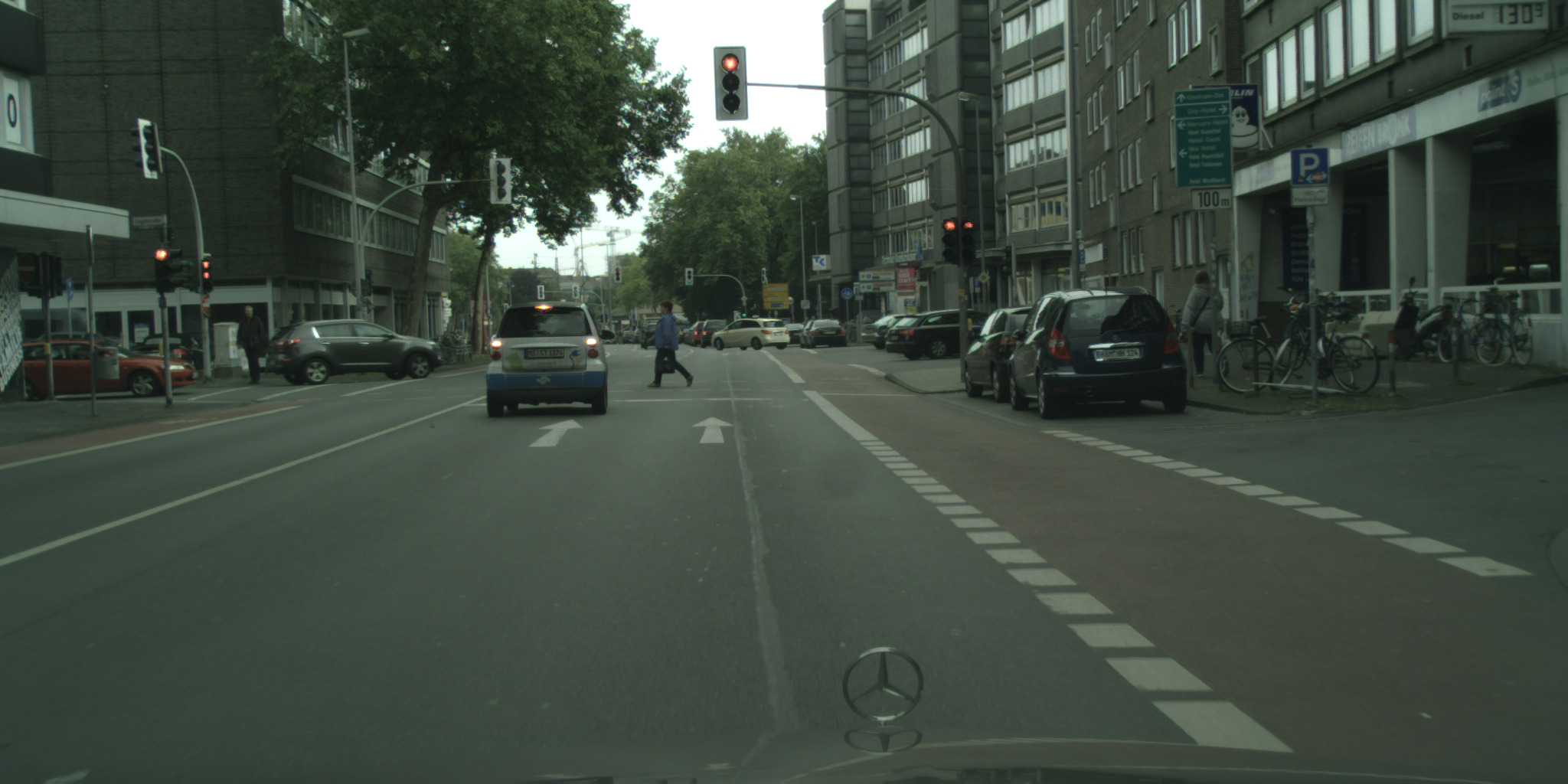}
\end{subfigure}\hfill
\begin{subfigure}{.332\linewidth}
  \centering
  \includegraphics[trim={0 150 0 150},clip,width=\linewidth]{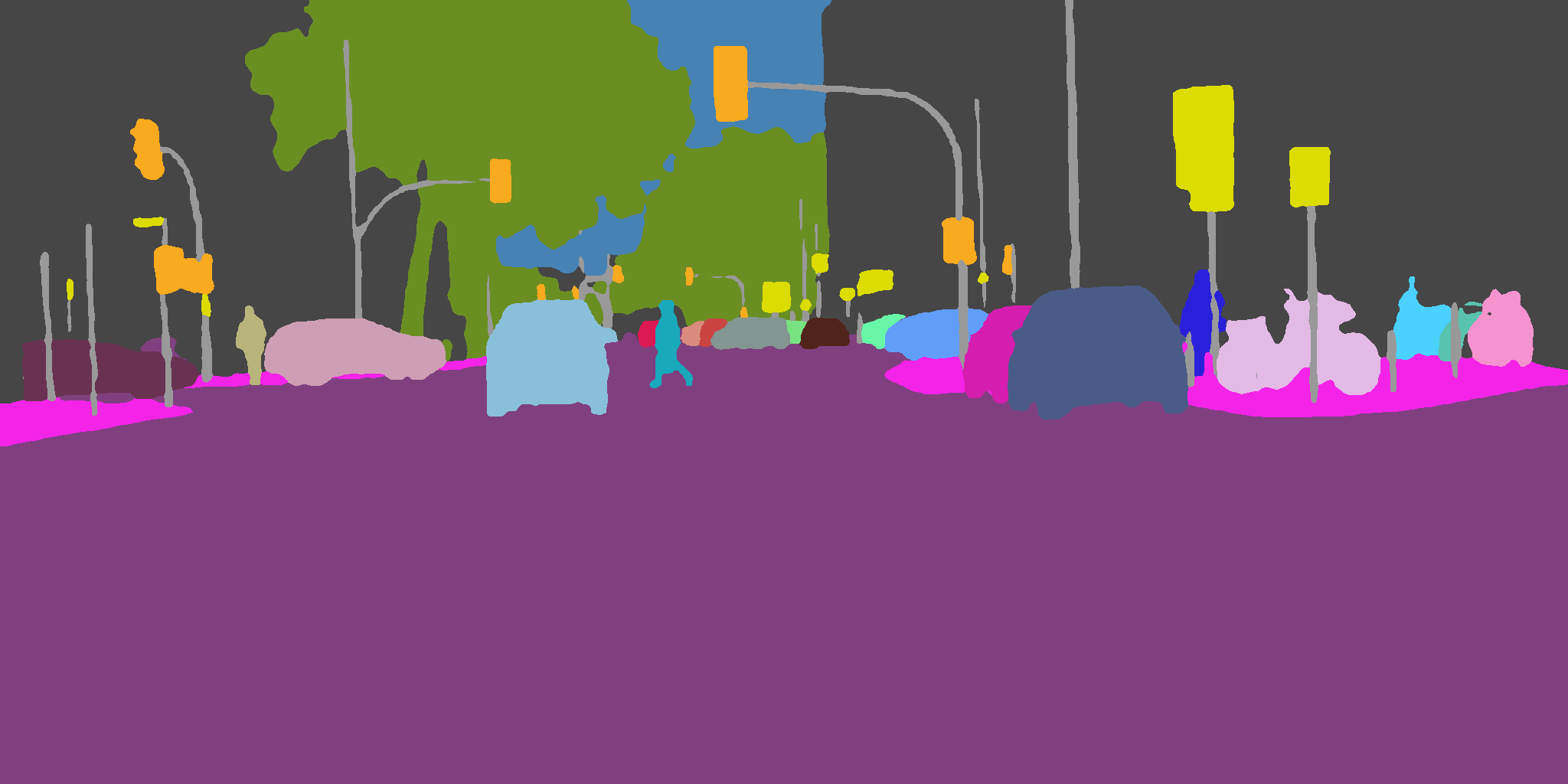}
\end{subfigure}\hfill
\begin{subfigure}{.332\linewidth}
  \centering
  \includegraphics[trim={0 150 0 150},clip,width=\linewidth]{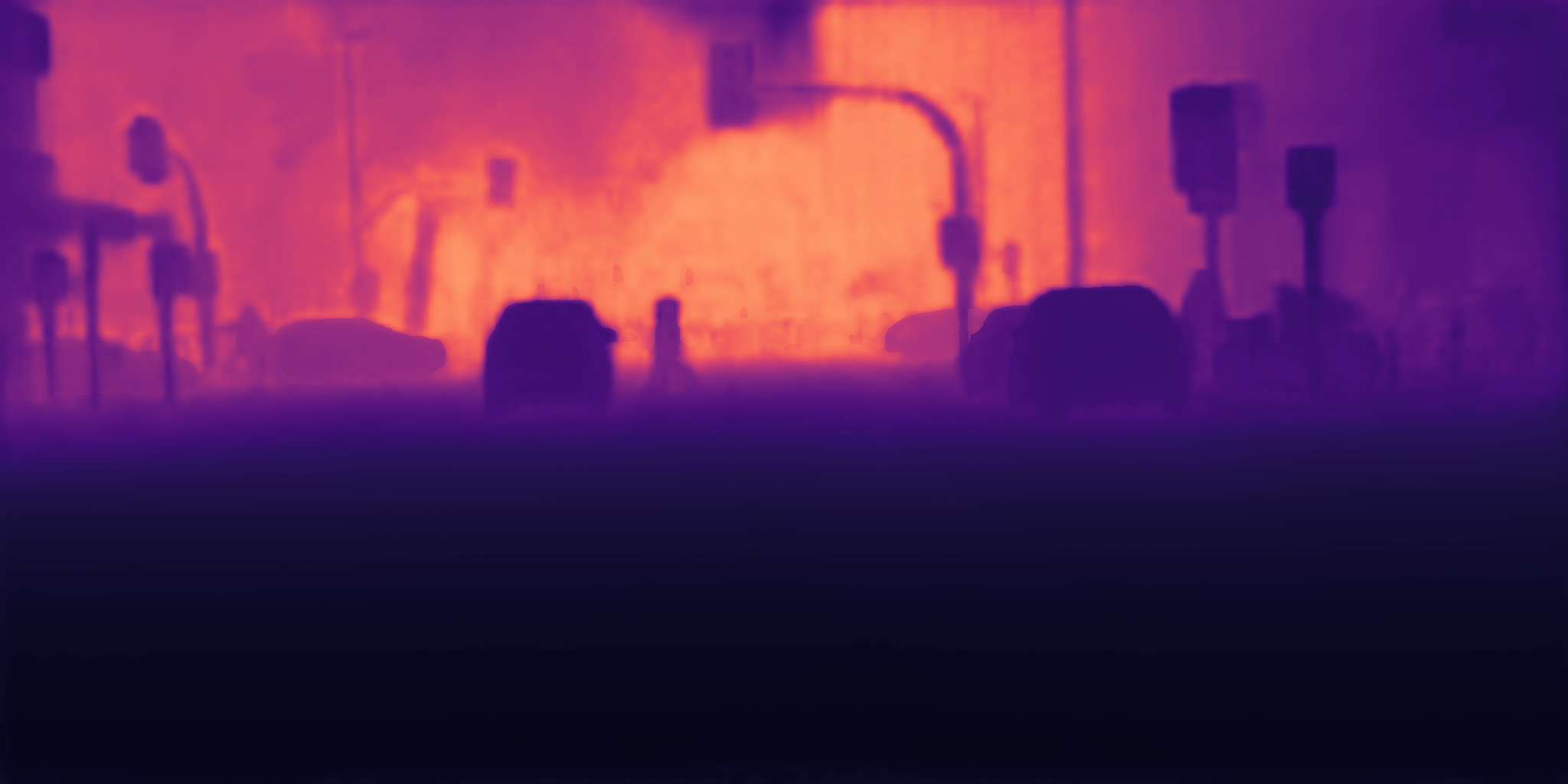}
\end{subfigure}\\

\begin{subfigure}{.332\linewidth}
  \centering
  \includegraphics[trim={0 150 0 150},clip,width=\linewidth]{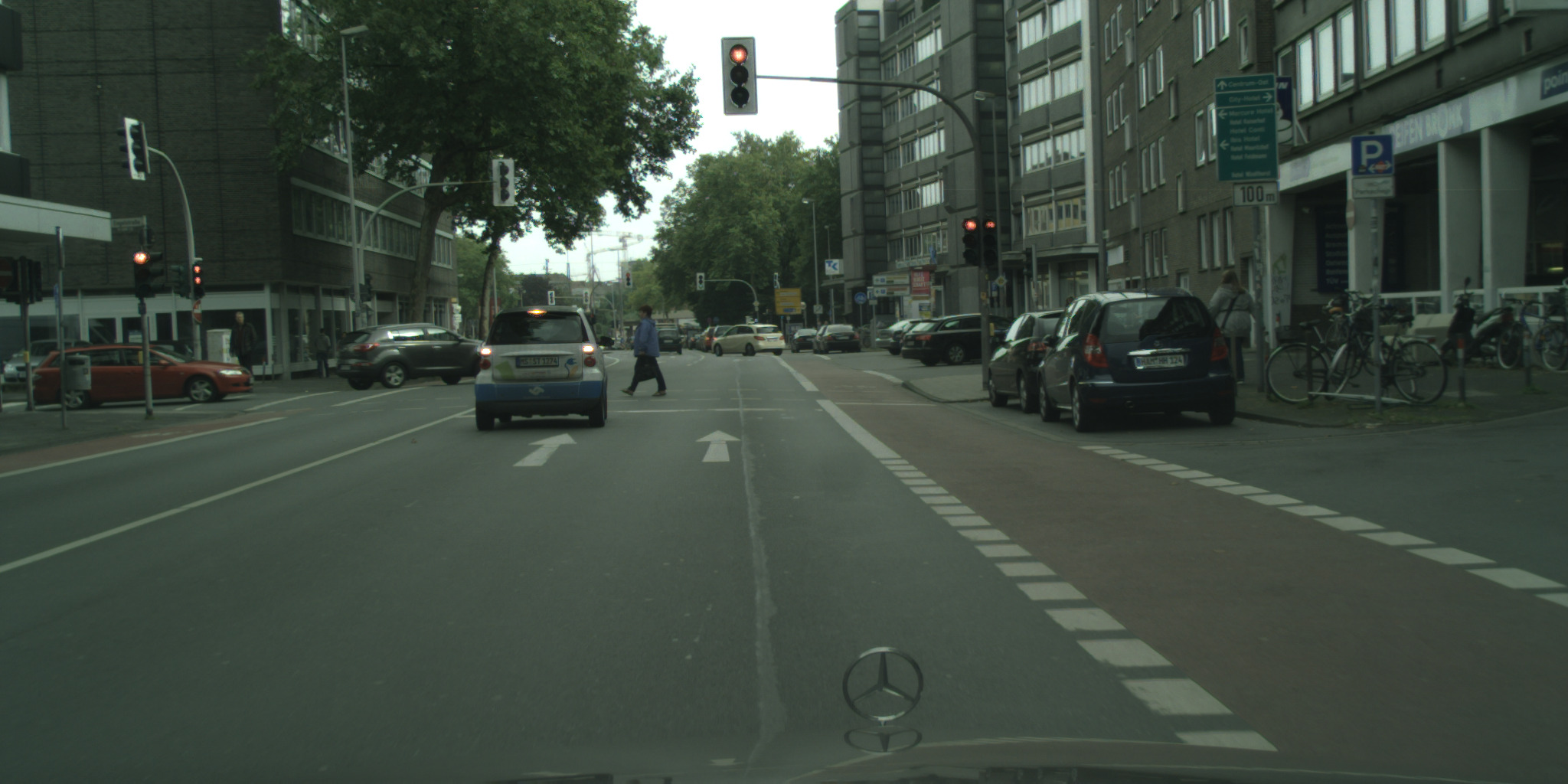}
\end{subfigure}\hfill
\begin{subfigure}{.332\linewidth}
  \centering
  \includegraphics[trim={0 150 0 150},clip,width=\linewidth]{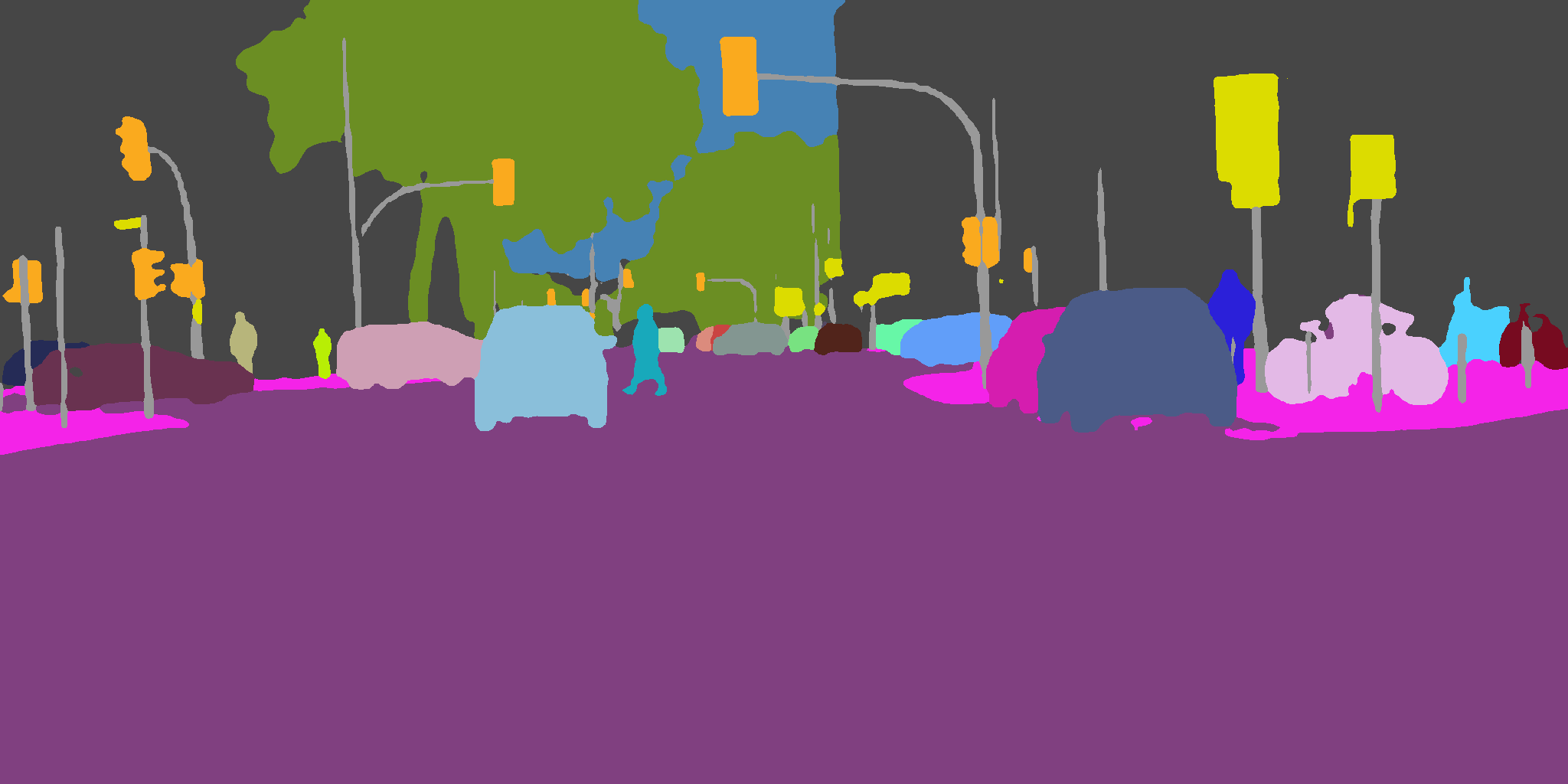}
\end{subfigure}\hfill
\begin{subfigure}{.332\linewidth}
  \centering
  \includegraphics[trim={0 150 0 150},clip,width=\linewidth]{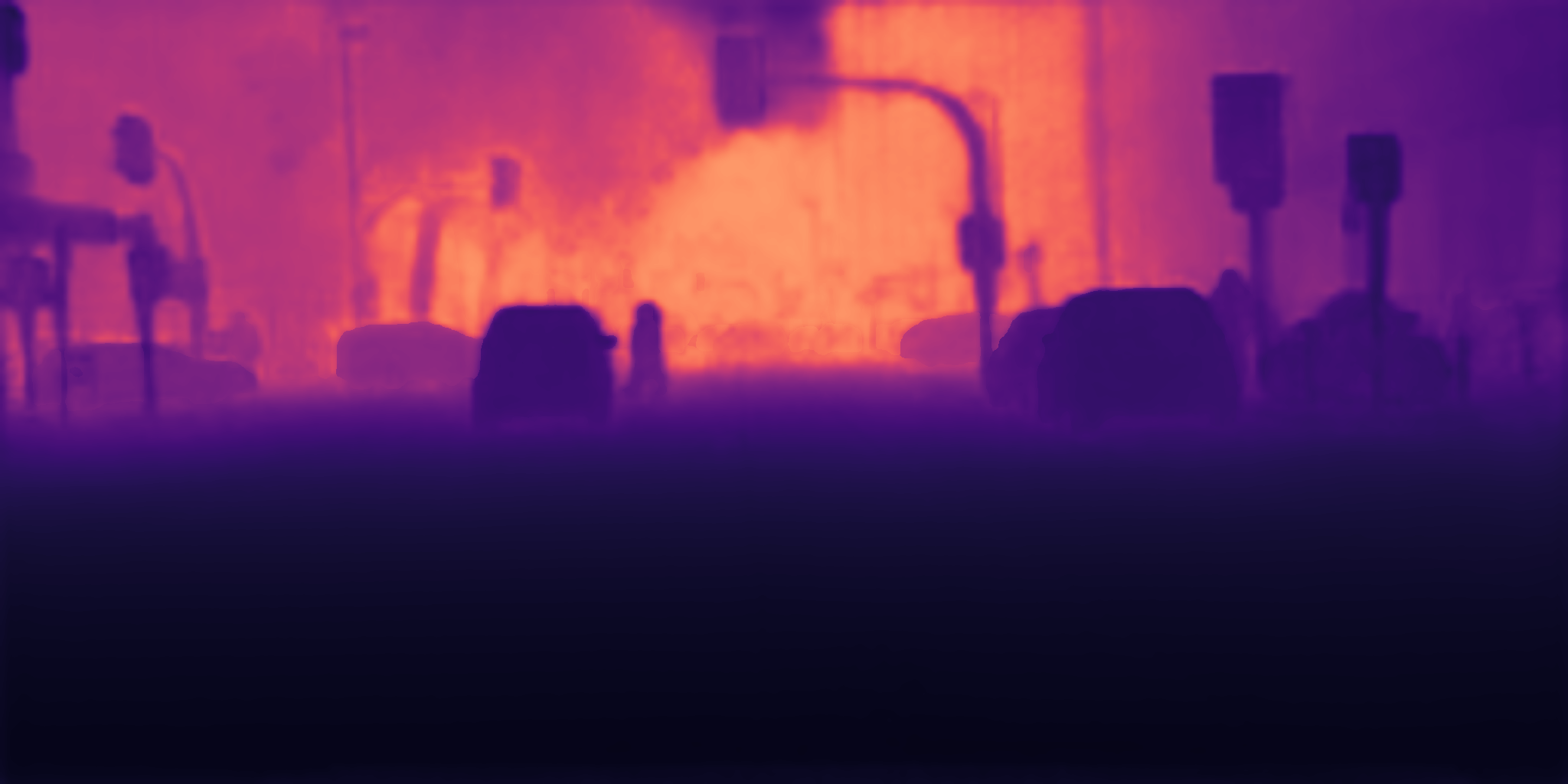}
\end{subfigure}\\

\begin{subfigure}{.332\linewidth}
  \centering
  \includegraphics[trim={0 150 0 150},clip,width=\linewidth]{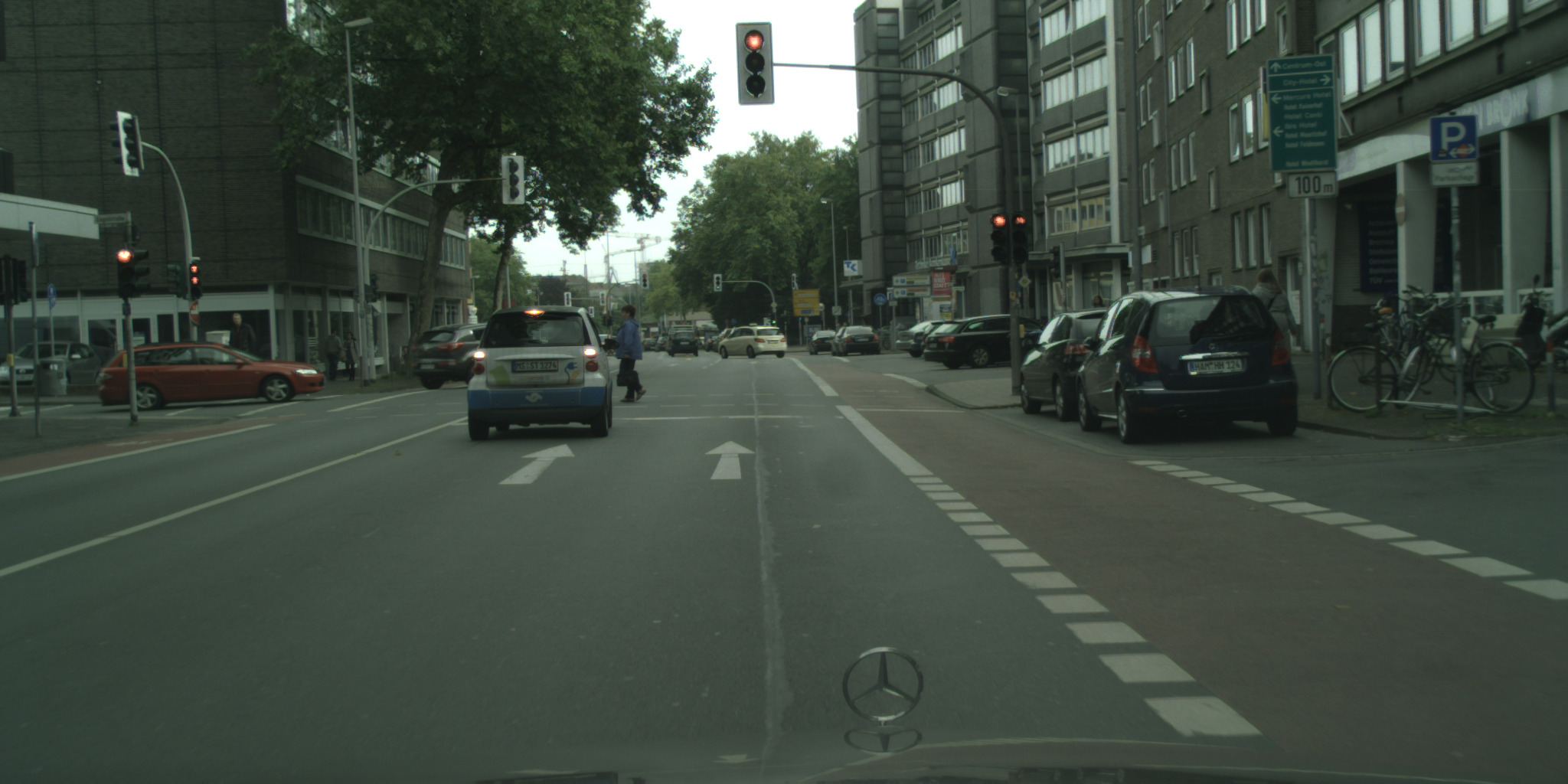}
\end{subfigure}\hfill
\begin{subfigure}{.332\linewidth}
  \centering
  \includegraphics[trim={0 150 0 150},clip,width=\linewidth]{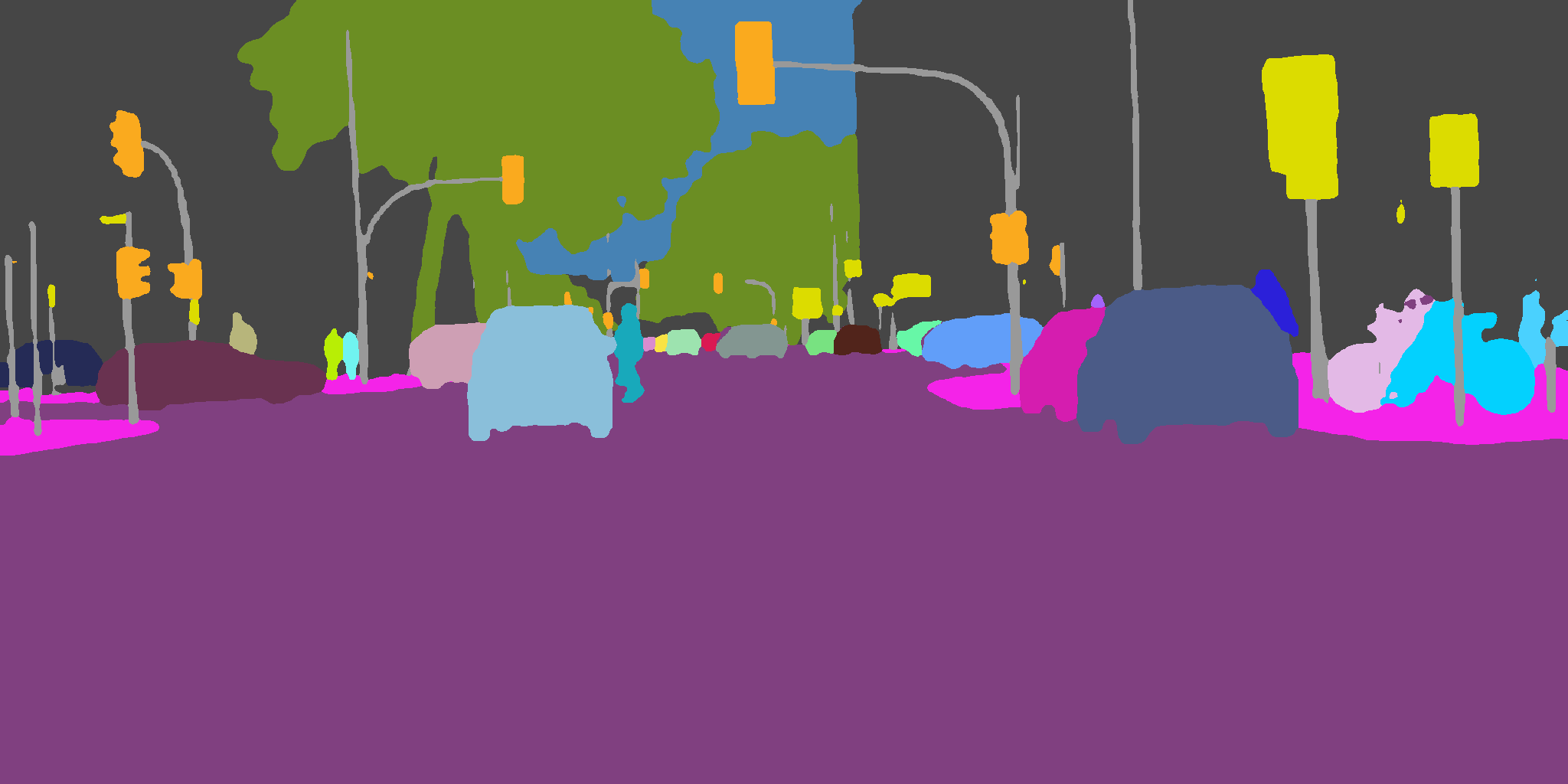}
\end{subfigure}\hfill
\begin{subfigure}{.332\linewidth}
  \centering
  \includegraphics[trim={0 150 0 150},clip,width=\linewidth]{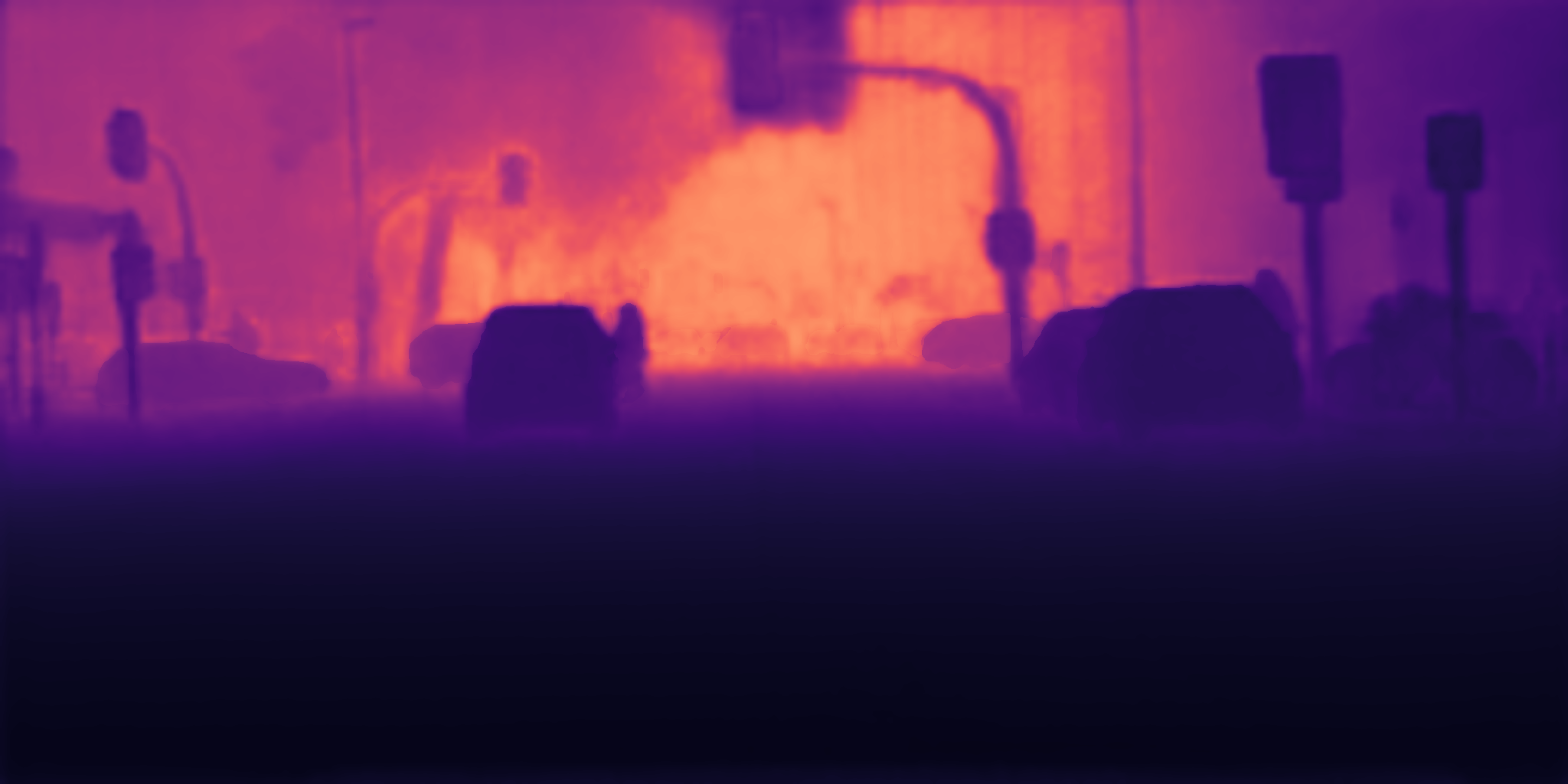}
\end{subfigure}\\

\caption{Prediction visualizations on Cityscapes-DVPS. From left to right: input images, temporally consistent panoptic segmentation (TCPS), and depth predictions. Color change of the same instance of TCPS indicates an id switch.}
\label{fig:cs_1}
\end{figure*}

\begin{figure*}[!t]
\centering
\begin{subfigure}{.332\linewidth}
  \centering
  \includegraphics[trim={150 0 150 125},clip,width=\linewidth]{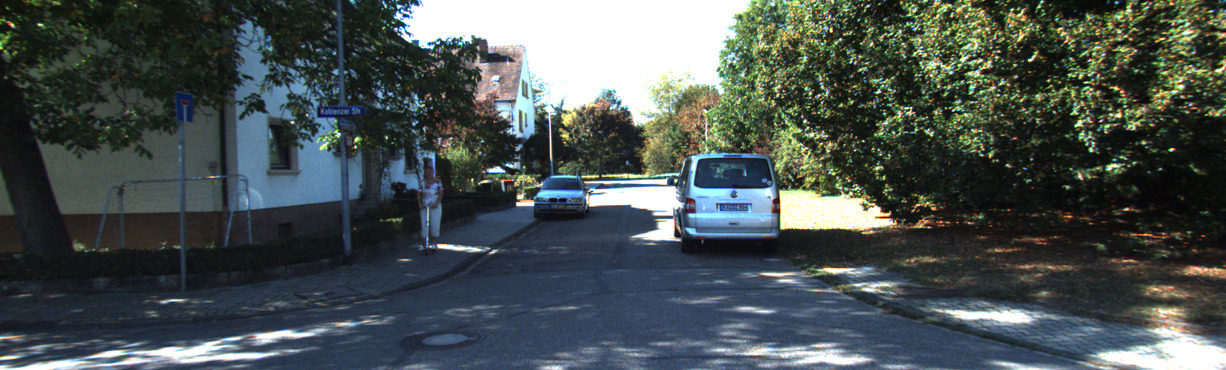}
\end{subfigure}\hfill
\begin{subfigure}{.332\linewidth}
  \centering
  \includegraphics[trim={150 0 150 125},clip,width=\linewidth]{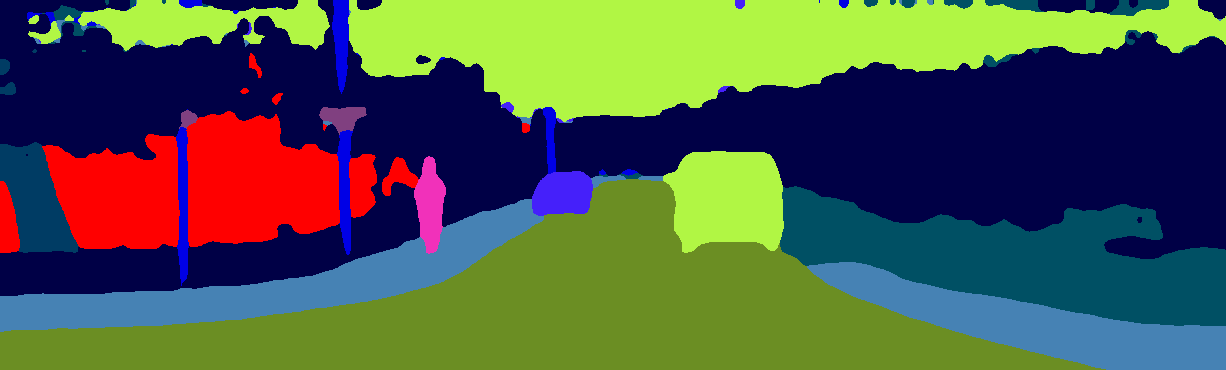}
\end{subfigure}\hfill
\begin{subfigure}{.332\linewidth}
  \centering
  \includegraphics[trim={150 0 150 125},clip,width=\linewidth]{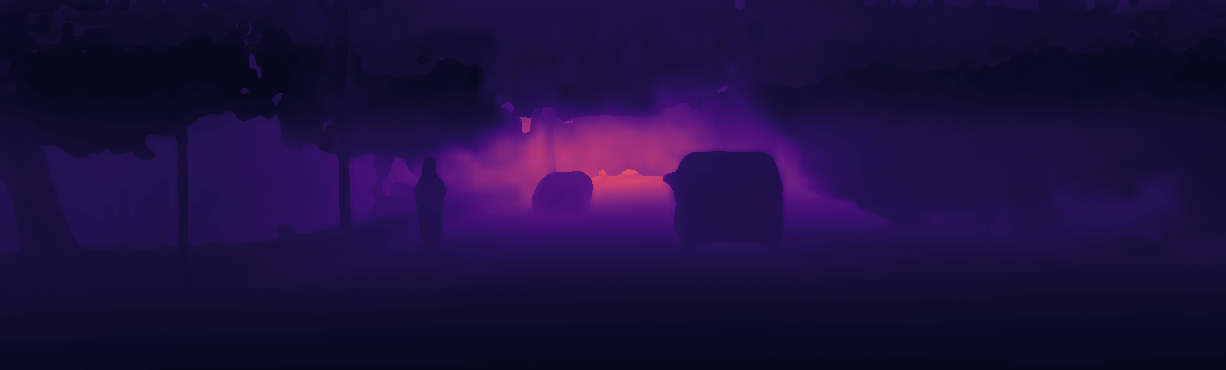}
\end{subfigure}\\

\begin{subfigure}{.332\linewidth}
  \centering
  \includegraphics[trim={150 0 150 125},clip,width=\linewidth]{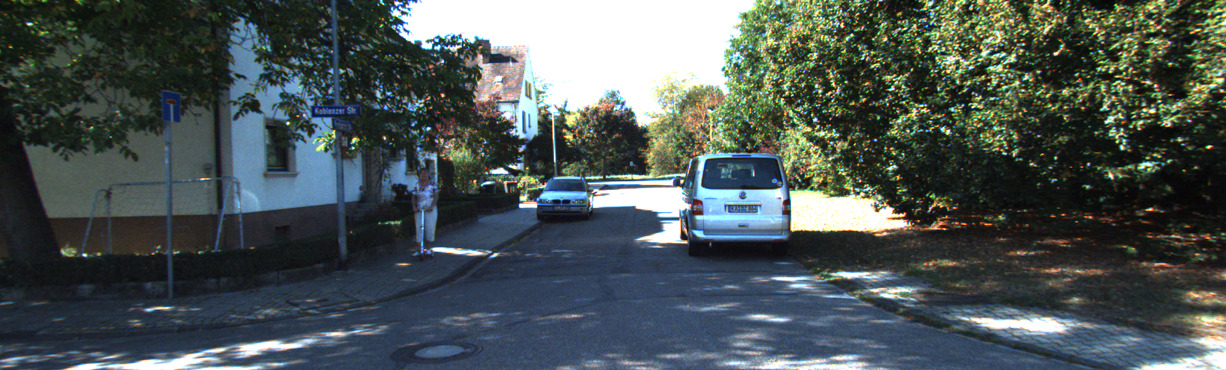}
\end{subfigure}\hfill
\begin{subfigure}{.332\linewidth}
  \centering
  \includegraphics[trim={150 0 150 125},clip,width=\linewidth]{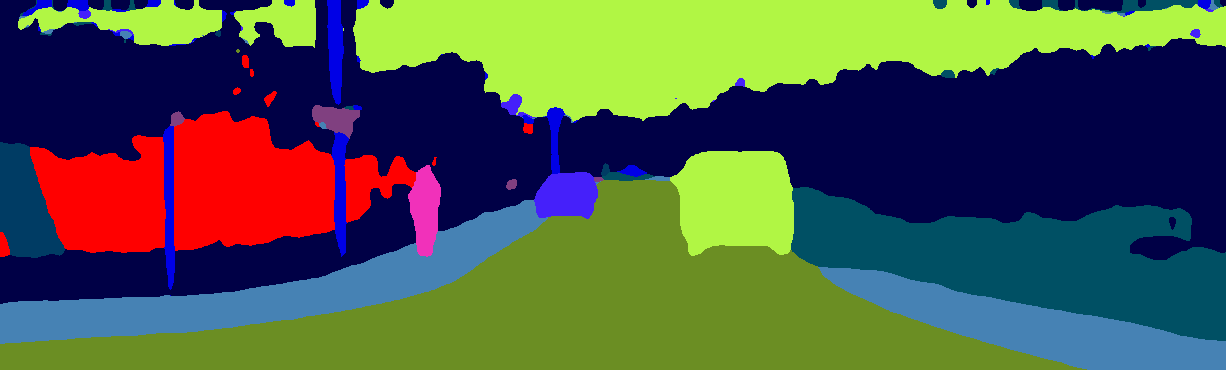}
\end{subfigure}\hfill
\begin{subfigure}{.332\linewidth}
  \centering
  \includegraphics[trim={150 0 150 125},clip,width=\linewidth]{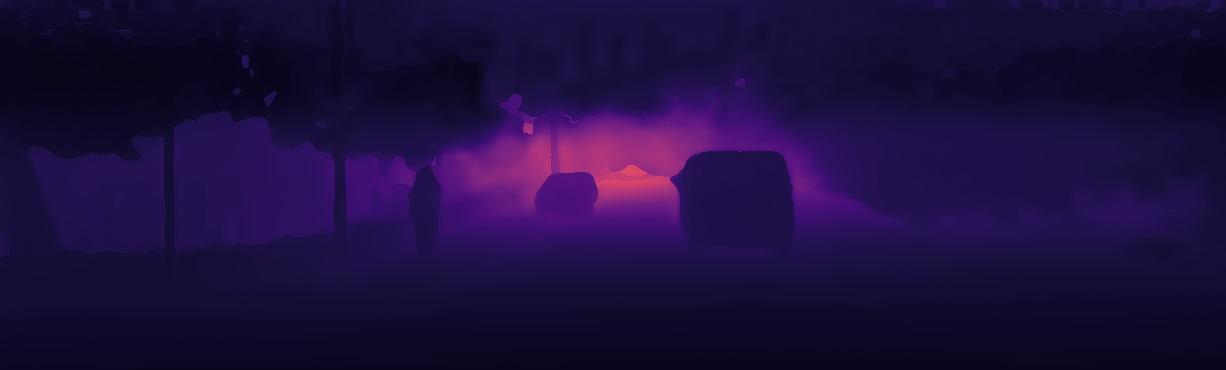}
\end{subfigure}\\

\begin{subfigure}{.332\linewidth}
  \centering
  \includegraphics[trim={150 0 150 125},clip,width=\linewidth]{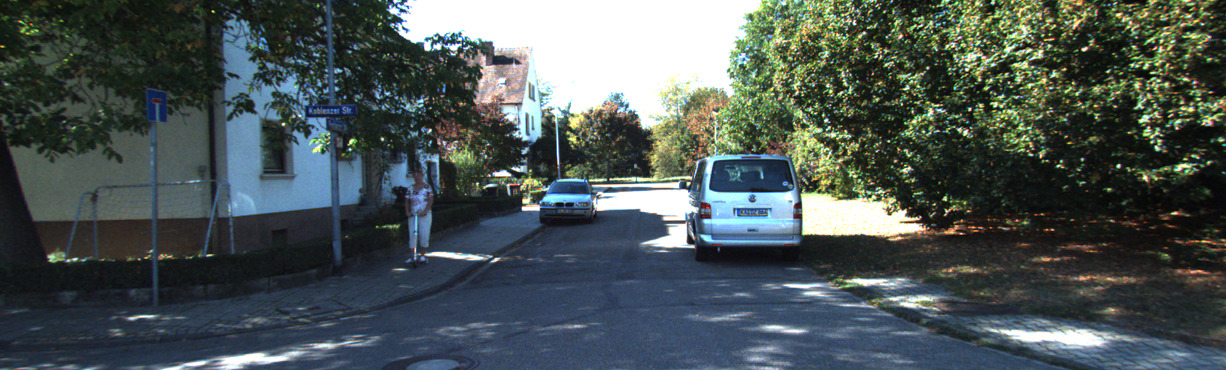}
\end{subfigure}\hfill
\begin{subfigure}{.332\linewidth}
  \centering
  \includegraphics[trim={150 0 150 125},clip,width=\linewidth]{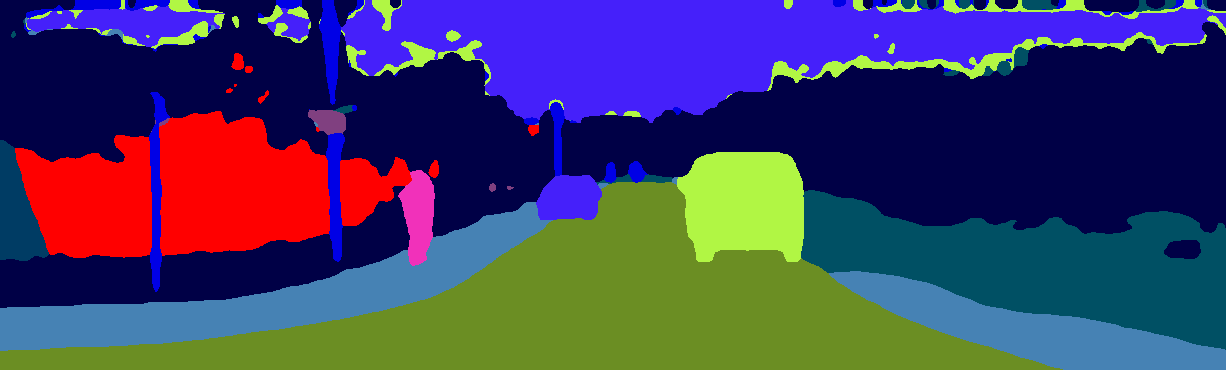}
\end{subfigure}\hfill
\begin{subfigure}{.332\linewidth}
  \centering
  \includegraphics[trim={150 0 150 125},clip,width=\linewidth]{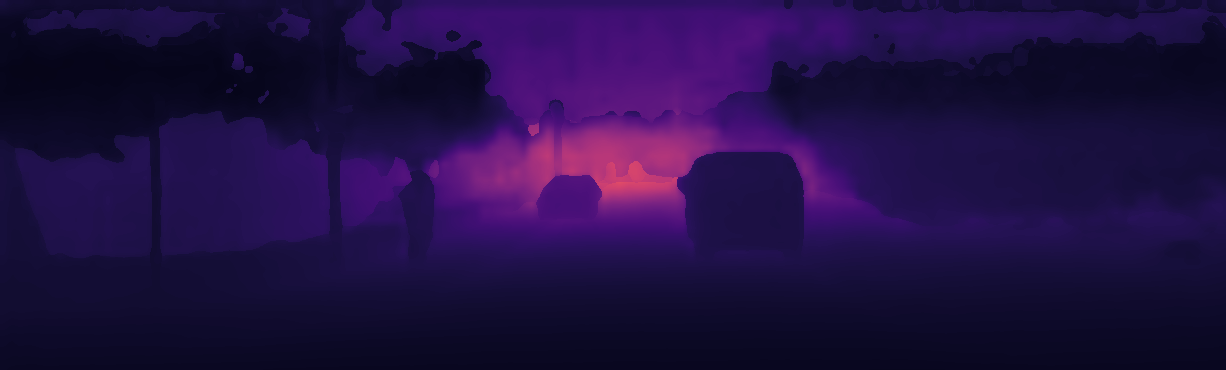}
\end{subfigure}\\

\begin{subfigure}{.332\linewidth}
  \centering
  \includegraphics[trim={150 0 150 125},clip,width=\linewidth]{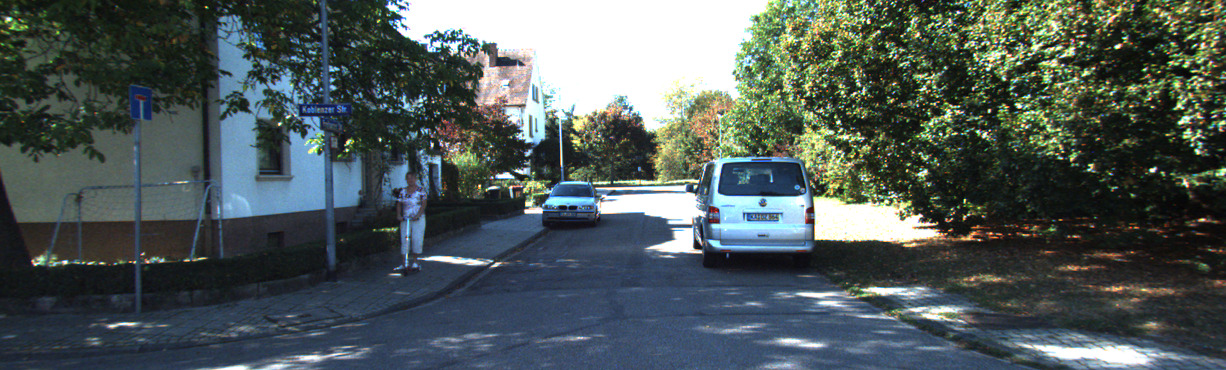}
\end{subfigure}\hfill
\begin{subfigure}{.332\linewidth}
  \centering
  \includegraphics[trim={150 0 150 125},clip,width=\linewidth]{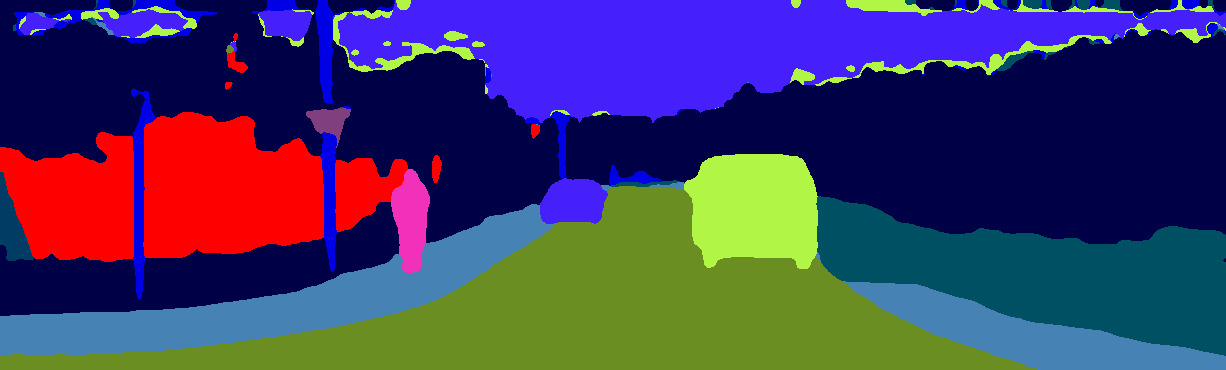}
\end{subfigure}\hfill
\begin{subfigure}{.332\linewidth}
  \centering
  \includegraphics[trim={150 0 150 125},clip,width=\linewidth]{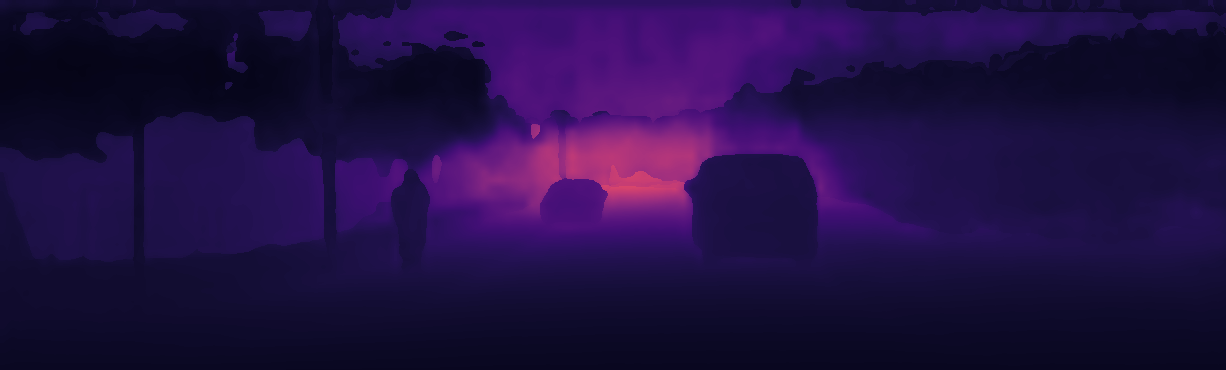}
\end{subfigure}\\

\begin{subfigure}{.332\linewidth}
  \centering
  \includegraphics[trim={150 0 150 125},clip,width=\linewidth]{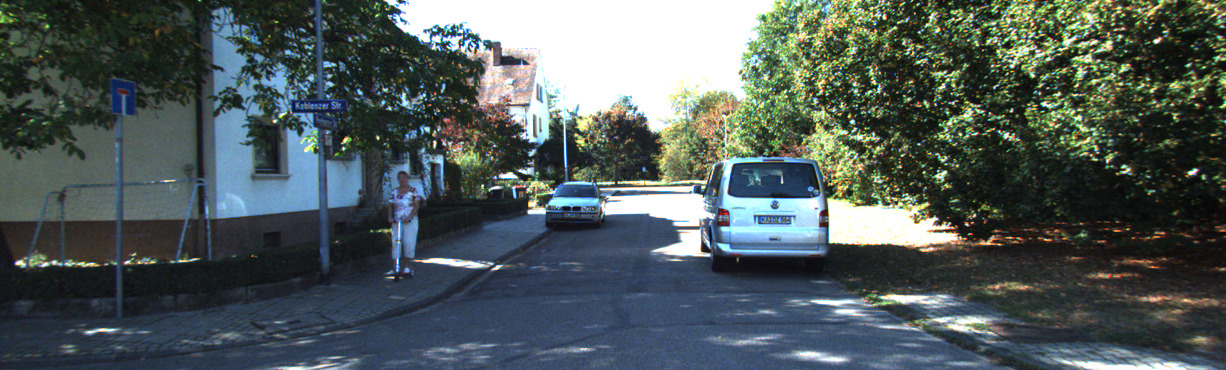}
\end{subfigure}\hfill
\begin{subfigure}{.332\linewidth}
  \centering
  \includegraphics[trim={150 0 150 125},clip,width=\linewidth]{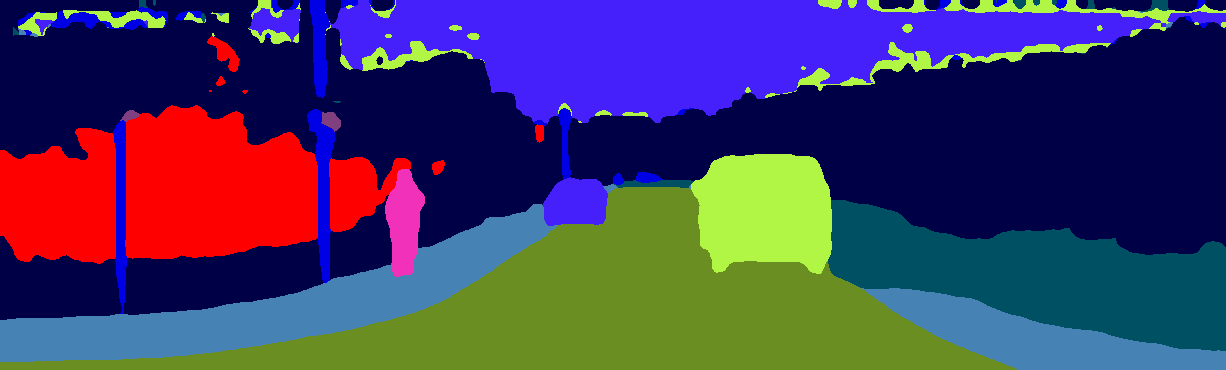}
\end{subfigure}\hfill
\begin{subfigure}{.332\linewidth}
  \centering
  \includegraphics[trim={150 0 150 125},clip,width=\linewidth]{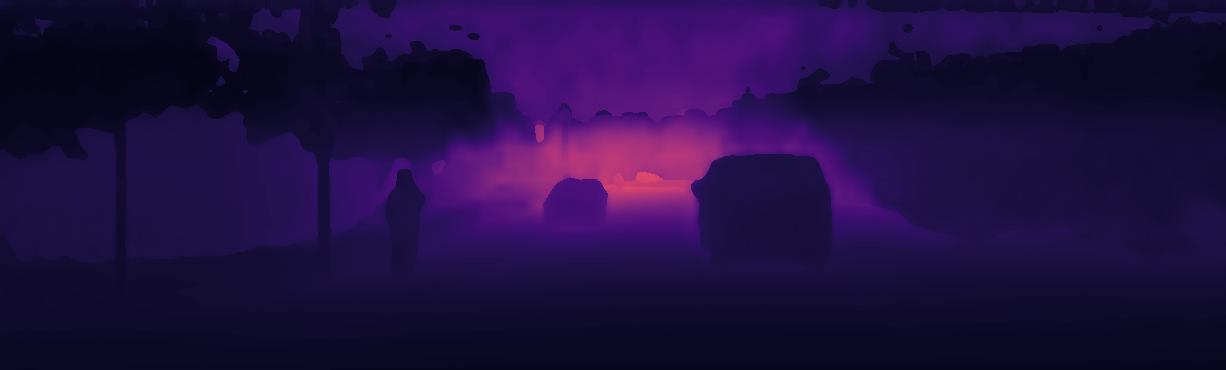}
\end{subfigure}\\

\begin{subfigure}{.332\linewidth}
  \centering
  \includegraphics[trim={150 0 150 125},clip,width=\linewidth]{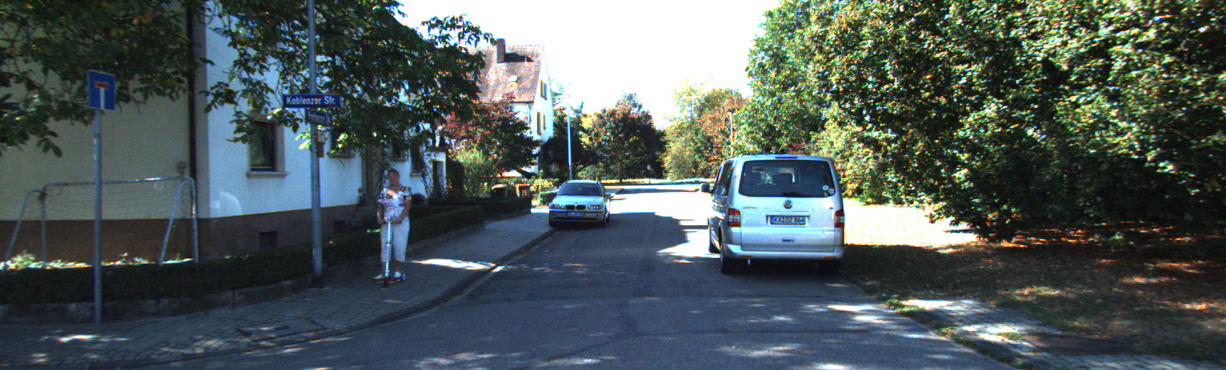}
\end{subfigure}\hfill
\begin{subfigure}{.332\linewidth}
  \centering
  \includegraphics[trim={150 0 150 125},clip,width=\linewidth]{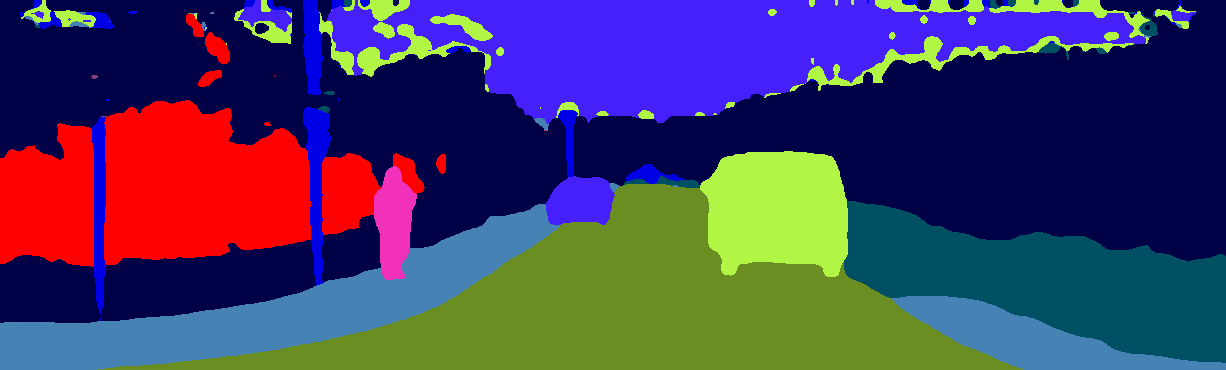}
\end{subfigure}\hfill
\begin{subfigure}{.332\linewidth}
  \centering
  \includegraphics[trim={150 0 150 125},clip,width=\linewidth]{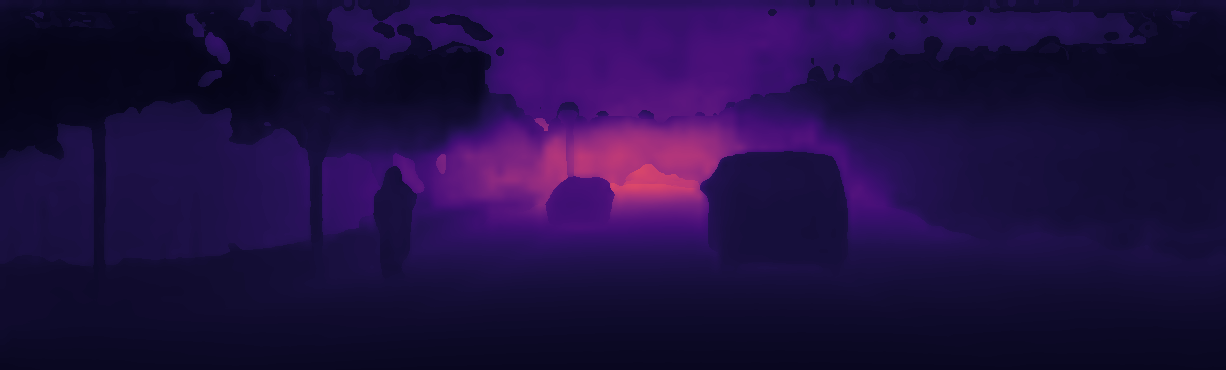}
\end{subfigure}\\

\begin{subfigure}{.332\linewidth}
  \centering
  \includegraphics[trim={150 0 150 125},clip,width=\linewidth]{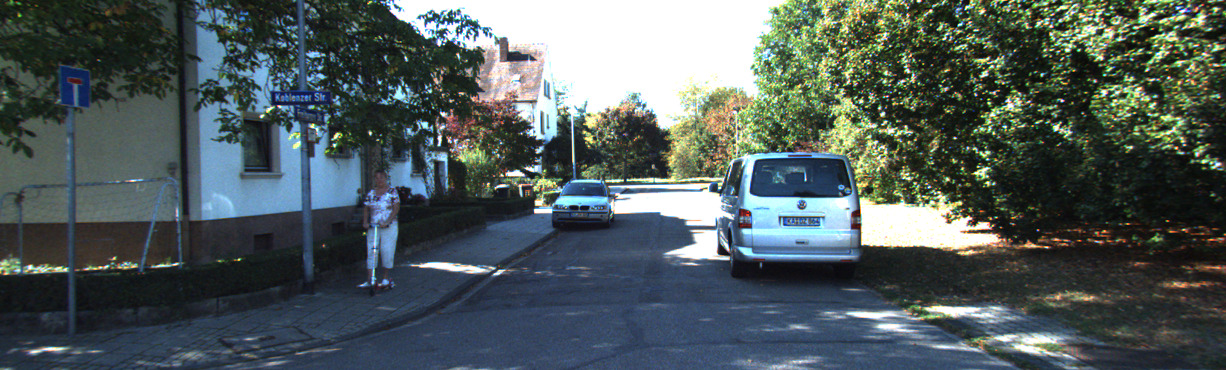}
\end{subfigure}\hfill
\begin{subfigure}{.332\linewidth}
  \centering
  \includegraphics[trim={150 0 150 125},clip,width=\linewidth]{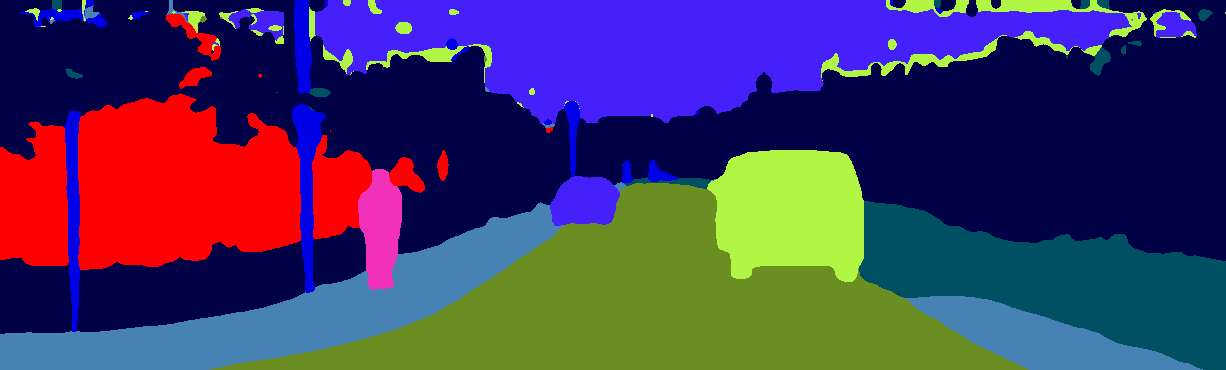}
\end{subfigure}\hfill
\begin{subfigure}{.332\linewidth}
  \centering
  \includegraphics[trim={150 0 150 125},clip,width=\linewidth]{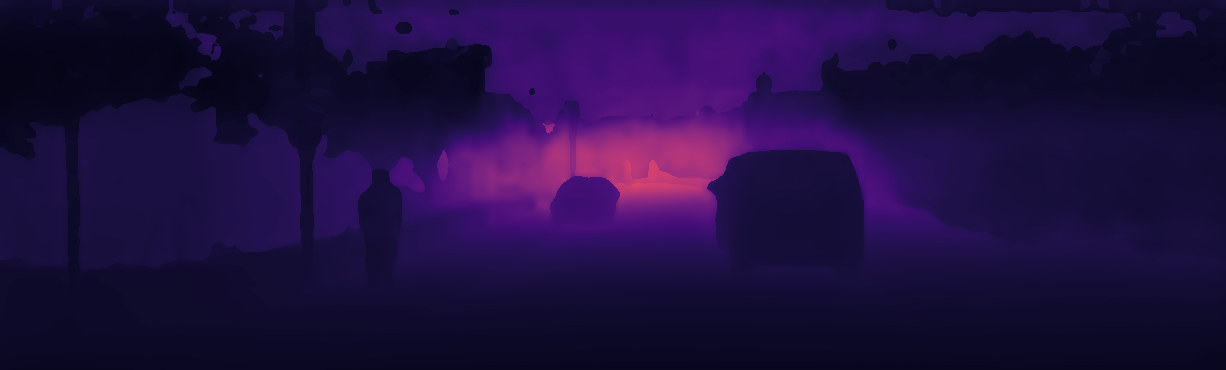}
\end{subfigure}\\

\begin{subfigure}{.332\linewidth}
  \centering
  \includegraphics[trim={150 0 150 125},clip,width=\linewidth]{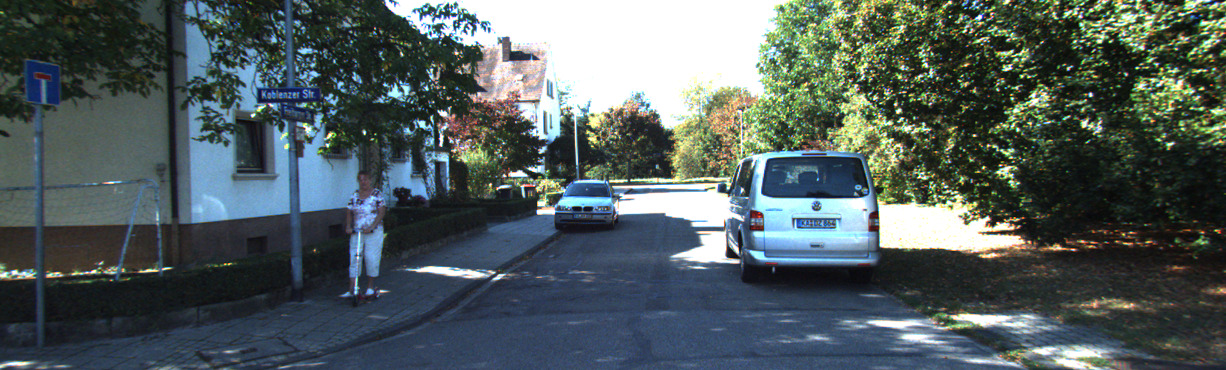}
\end{subfigure}\hfill
\begin{subfigure}{.332\linewidth}
  \centering
  \includegraphics[trim={150 0 150 125},clip,width=\linewidth]{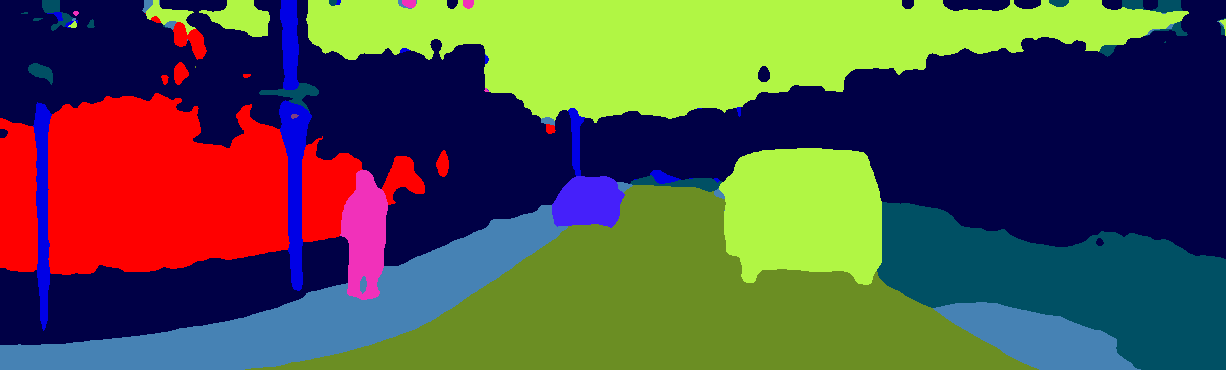}
\end{subfigure}\hfill
\begin{subfigure}{.332\linewidth}
  \centering
  \includegraphics[trim={150 0 150 125},clip,width=\linewidth]{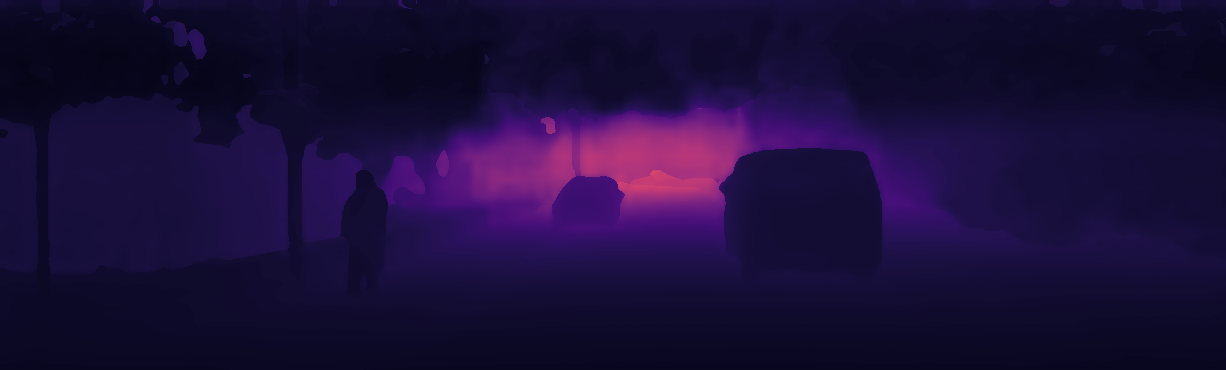}
\end{subfigure}\\

\begin{subfigure}{.332\linewidth}
  \centering
  \includegraphics[trim={150 0 150 125},clip,width=\linewidth]{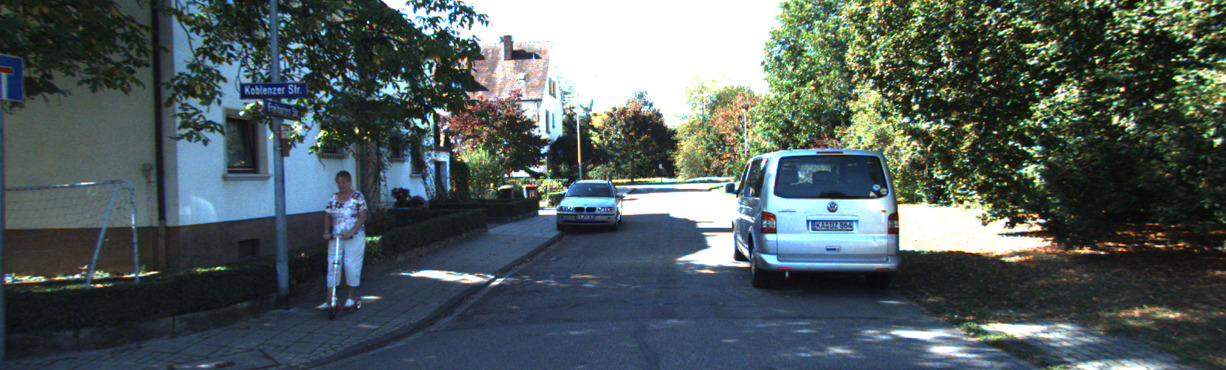}
\end{subfigure}\hfill
\begin{subfigure}{.332\linewidth}
  \centering
  \includegraphics[trim={150 0 150 125},clip,width=\linewidth]{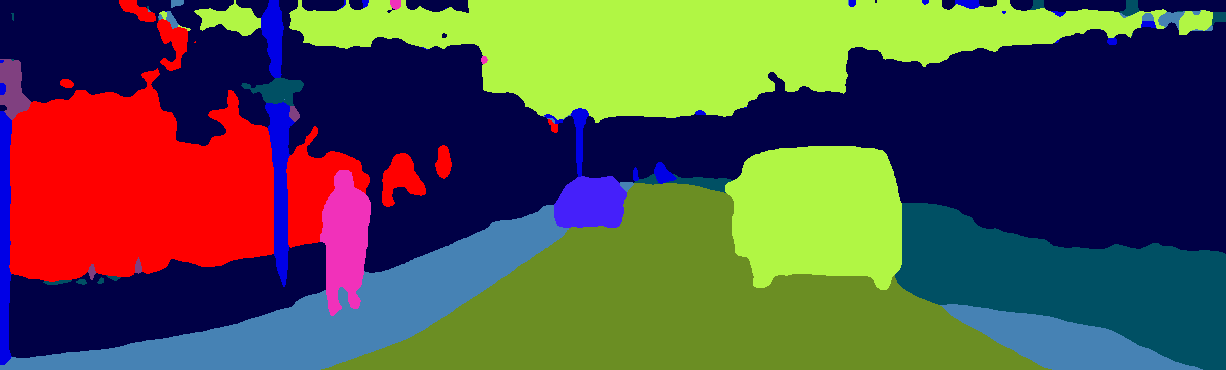}
\end{subfigure}\hfill
\begin{subfigure}{.332\linewidth}
  \centering
  \includegraphics[trim={150 0 150 125},clip,width=\linewidth]{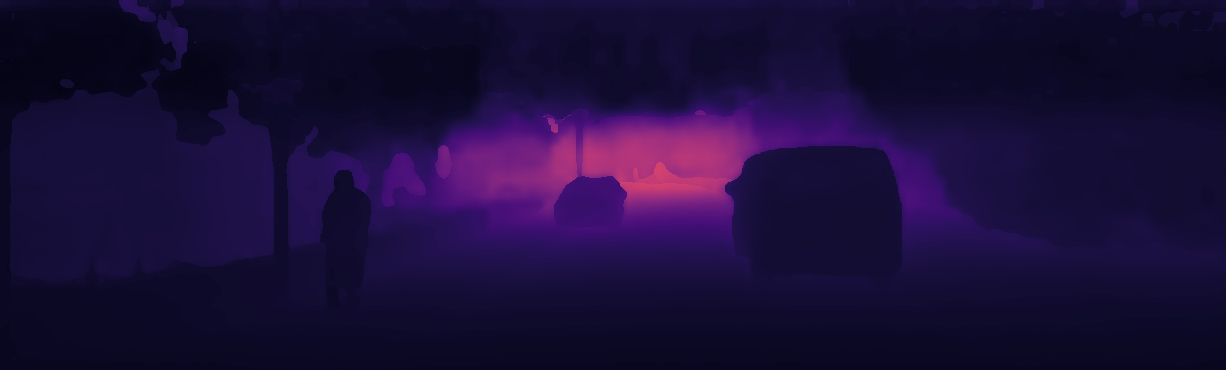}
\end{subfigure}\\

\begin{subfigure}{.332\linewidth}
  \centering
  \includegraphics[trim={150 0 150 125},clip,width=\linewidth]{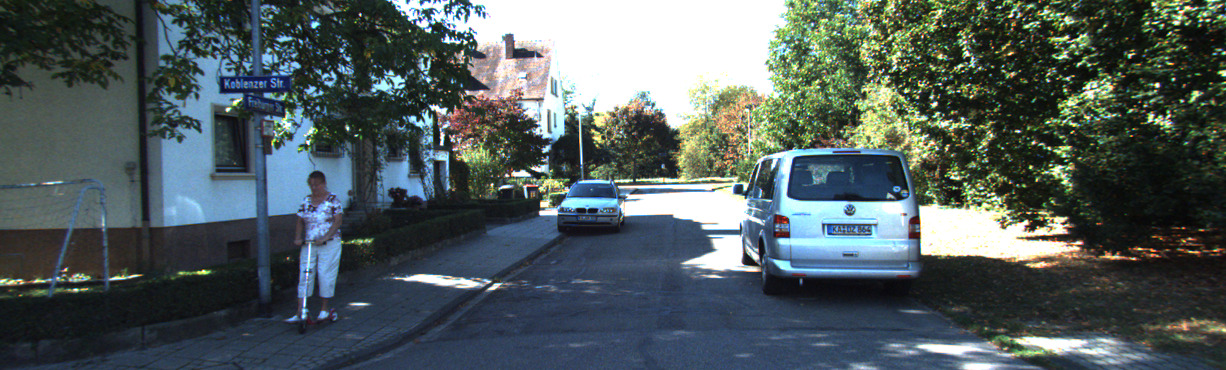}
\end{subfigure}\hfill
\begin{subfigure}{.332\linewidth}
  \centering
  \includegraphics[trim={150 0 150 125},clip,width=\linewidth]{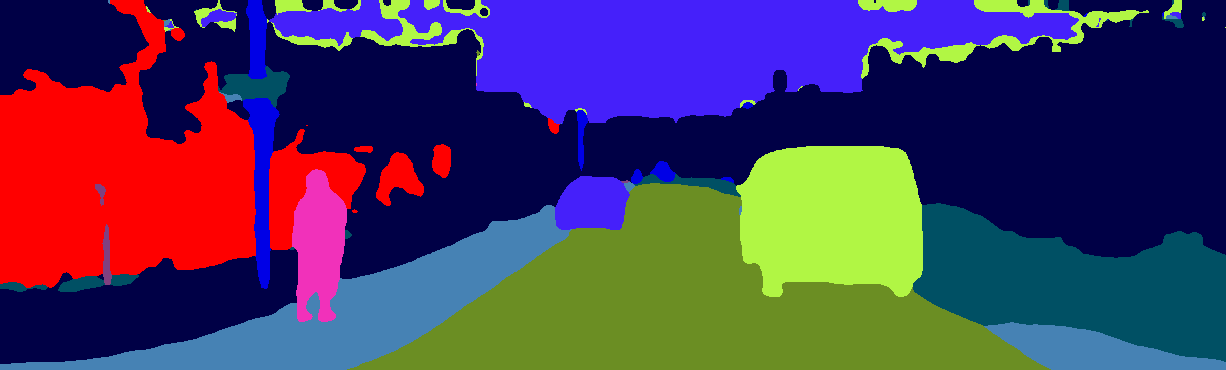}
\end{subfigure}\hfill
\begin{subfigure}{.332\linewidth}
  \centering
  \includegraphics[trim={150 0 150 125},clip,width=\linewidth]{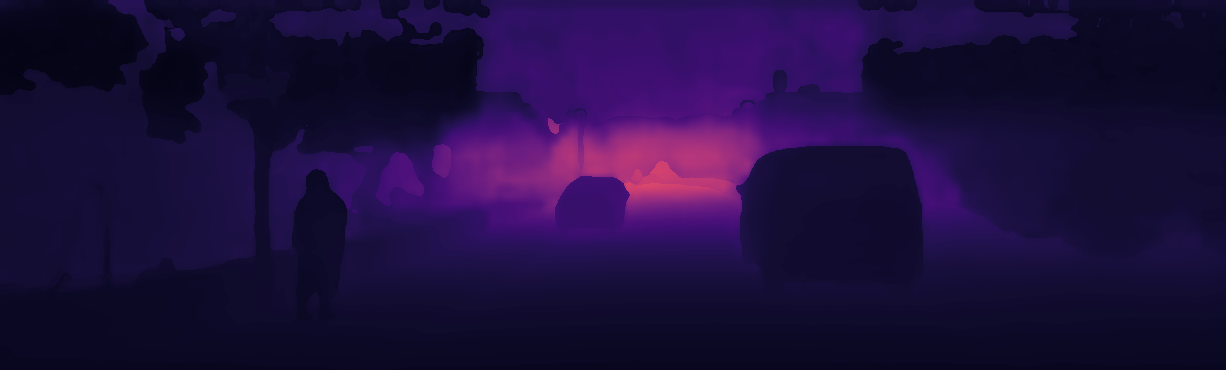}
\end{subfigure}\\

\begin{subfigure}{.332\linewidth}
  \centering
  \includegraphics[trim={150 0 150 125},clip,width=\linewidth]{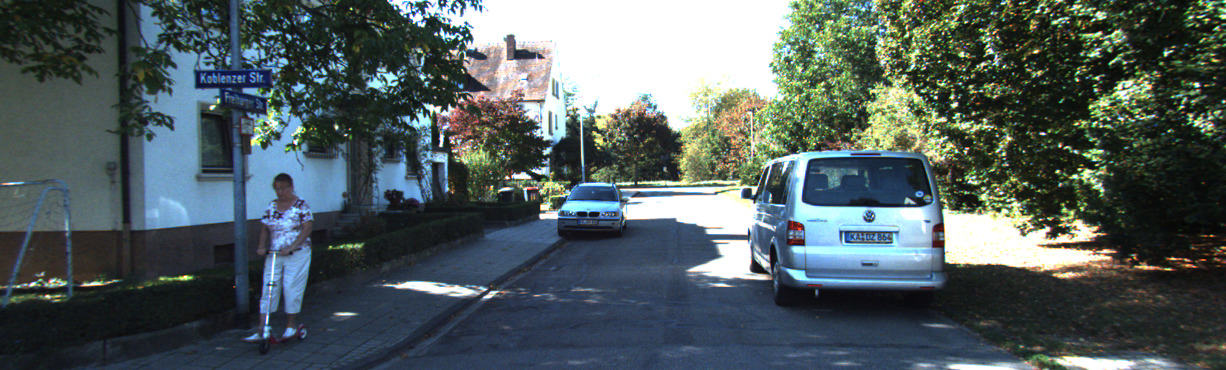}
\end{subfigure}\hfill
\begin{subfigure}{.332\linewidth}
  \centering
  \includegraphics[trim={150 0 150 125},clip,width=\linewidth]{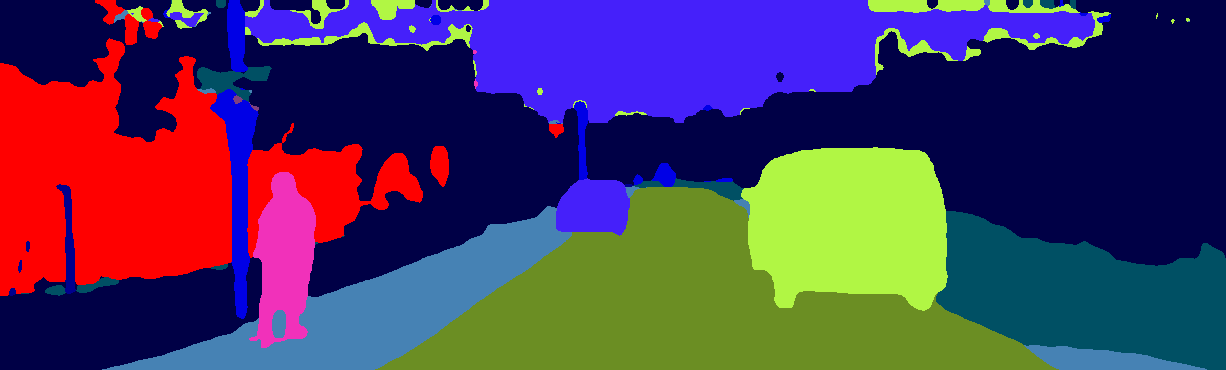}
\end{subfigure}\hfill
\begin{subfigure}{.332\linewidth}
  \centering
  \includegraphics[trim={150 0 150 125},clip,width=\linewidth]{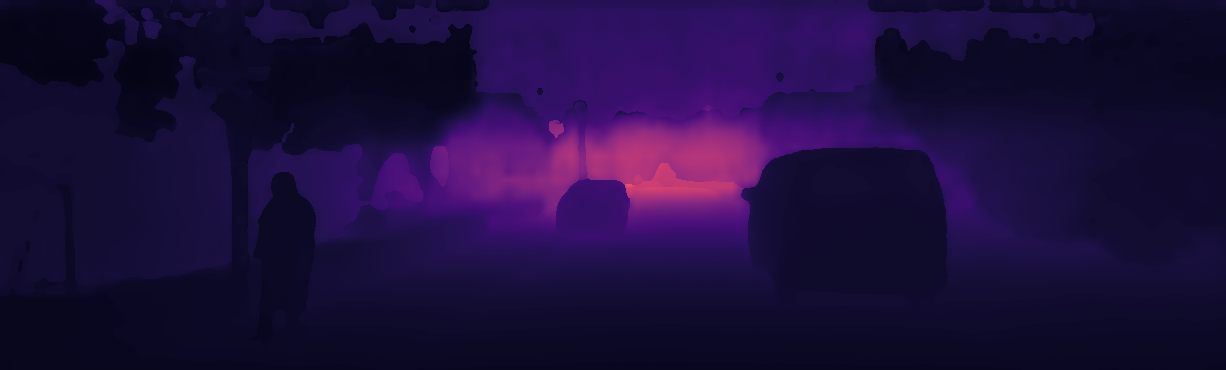}
\end{subfigure}\\

\begin{subfigure}{.332\linewidth}
  \centering
  \includegraphics[trim={150 0 150 125},clip,width=\linewidth]{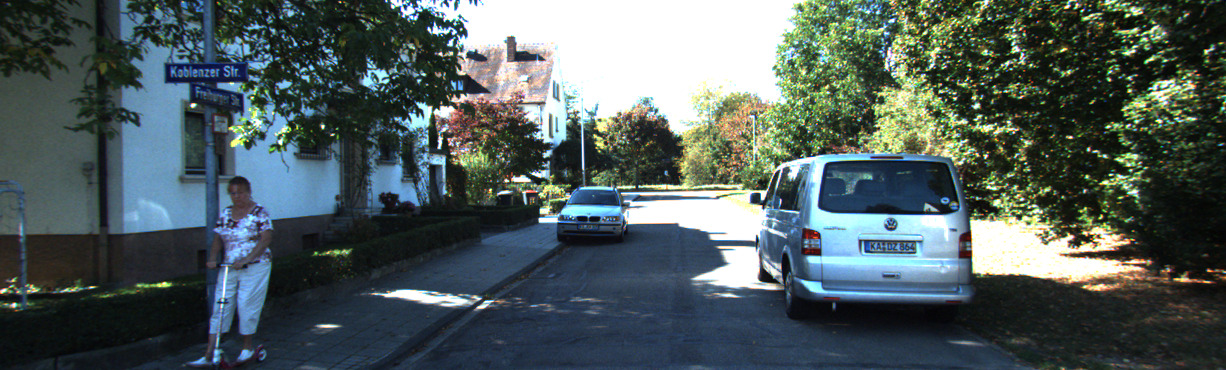}
\end{subfigure}\hfill
\begin{subfigure}{.332\linewidth}
  \centering
  \includegraphics[trim={150 0 150 125},clip,width=\linewidth]{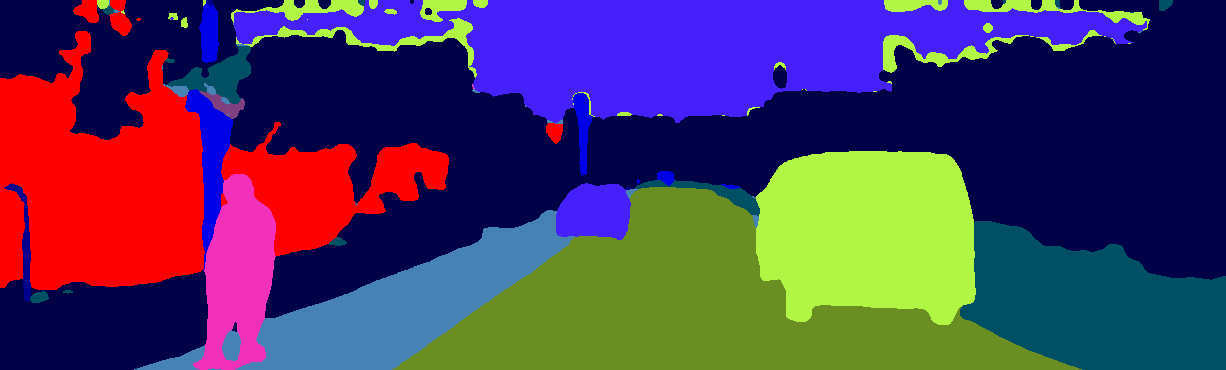}
\end{subfigure}\hfill
\begin{subfigure}{.332\linewidth}
  \centering
  \includegraphics[trim={150 0 150 125},clip,width=\linewidth]{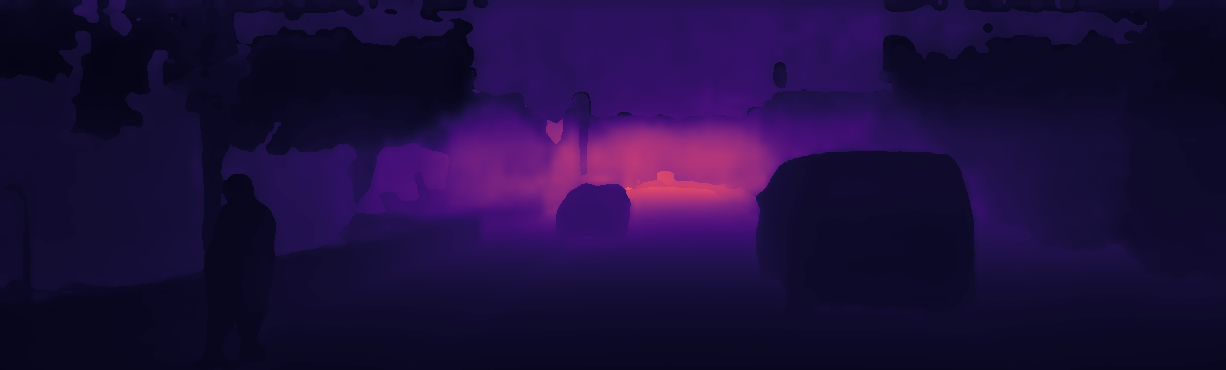}
\end{subfigure}\\

\begin{subfigure}{.332\linewidth}
  \centering
  \includegraphics[trim={150 0 150 125},clip,width=\linewidth]{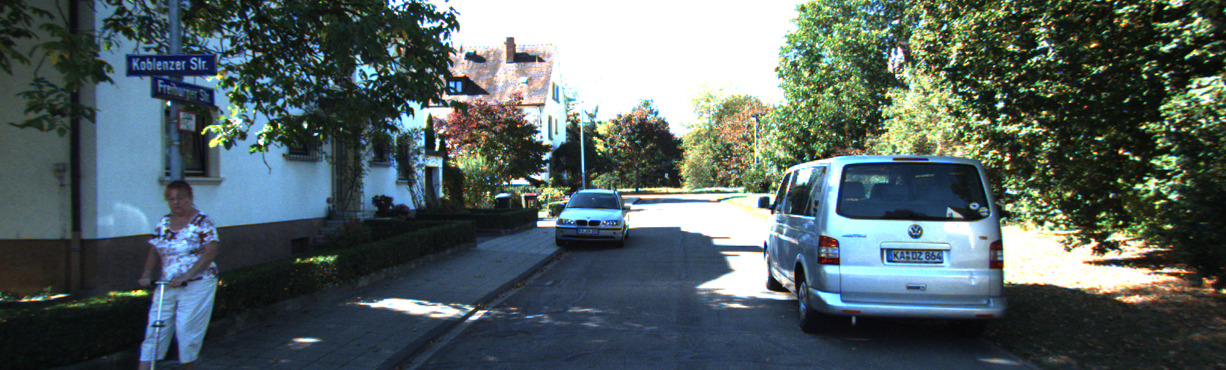}
\end{subfigure}\hfill
\begin{subfigure}{.332\linewidth}
  \centering
  \includegraphics[trim={150 0 150 125},clip,width=\linewidth]{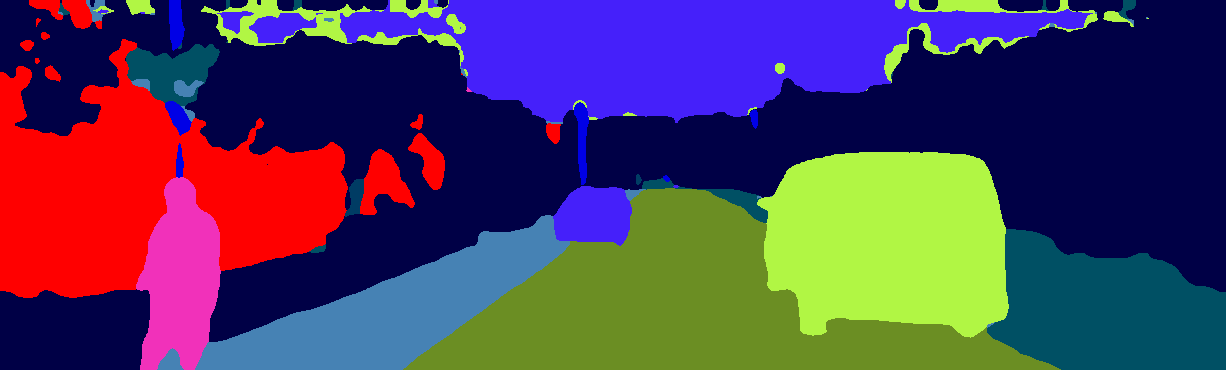}
\end{subfigure}\hfill
\begin{subfigure}{.332\linewidth}
  \centering
  \includegraphics[trim={150 0 150 125},clip,width=\linewidth]{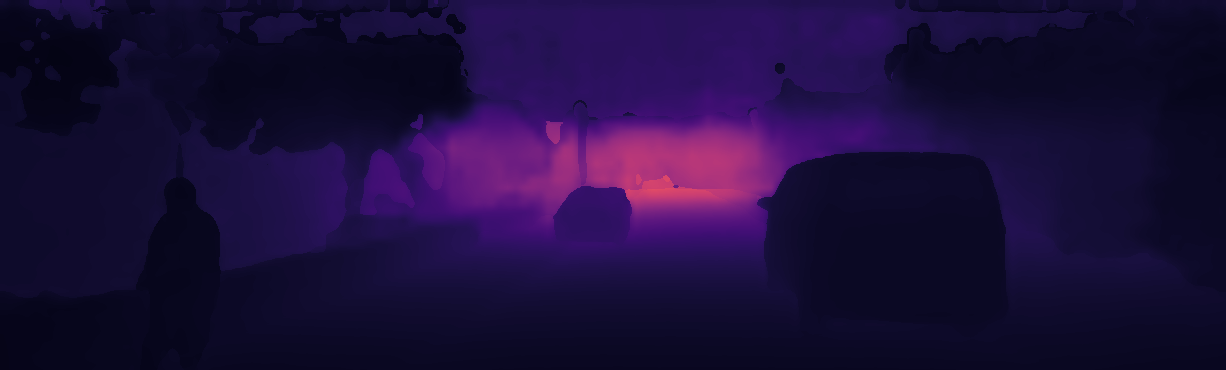}
\end{subfigure}\\

\begin{subfigure}{.332\linewidth}
  \centering
  \includegraphics[trim={150 0 150 125},clip,width=\linewidth]{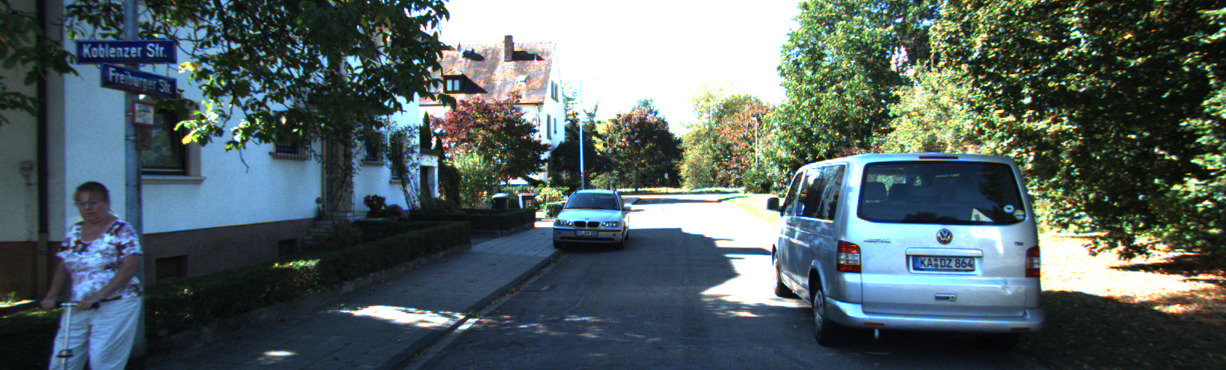}
\end{subfigure}\hfill
\begin{subfigure}{.332\linewidth}
  \centering
  \includegraphics[trim={150 0 150 125},clip,width=\linewidth]{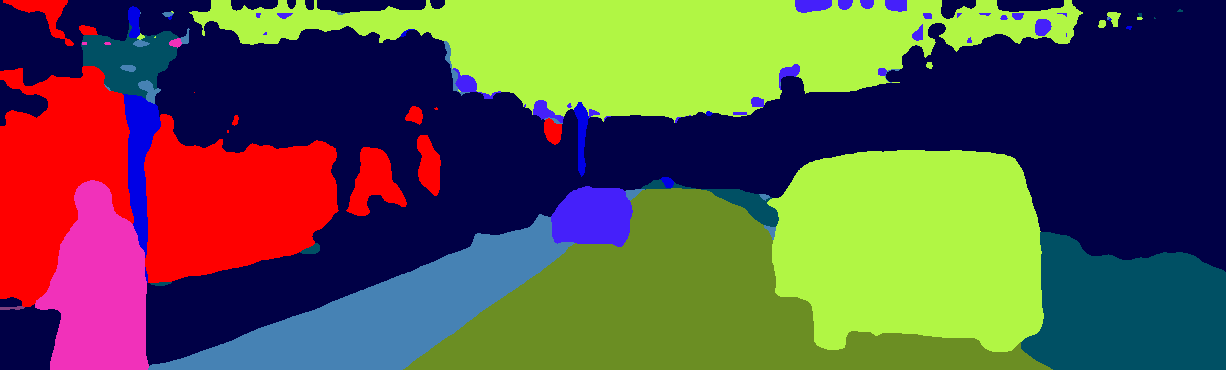}
\end{subfigure}\hfill
\begin{subfigure}{.332\linewidth}
  \centering
  \includegraphics[trim={150 0 150 125},clip,width=\linewidth]{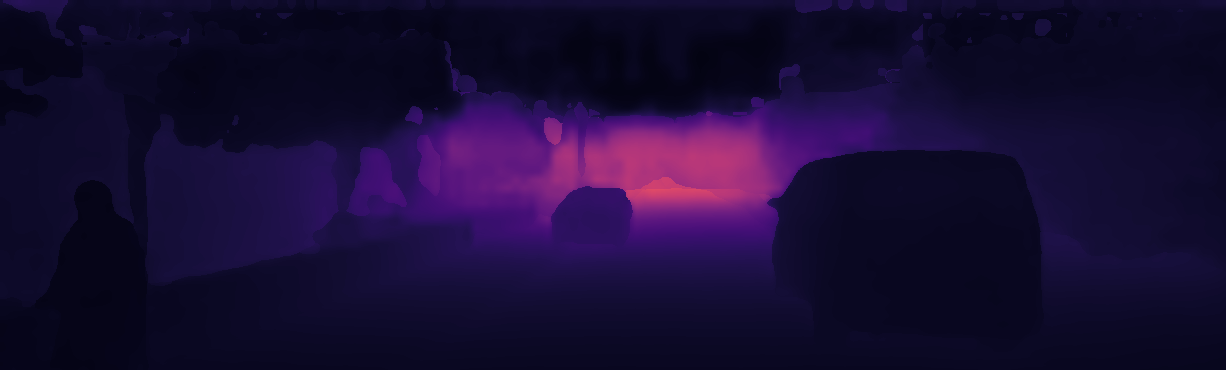}
\end{subfigure}\\

\begin{subfigure}{.332\linewidth}
  \centering
  \includegraphics[trim={150 0 150 125},clip,width=\linewidth]{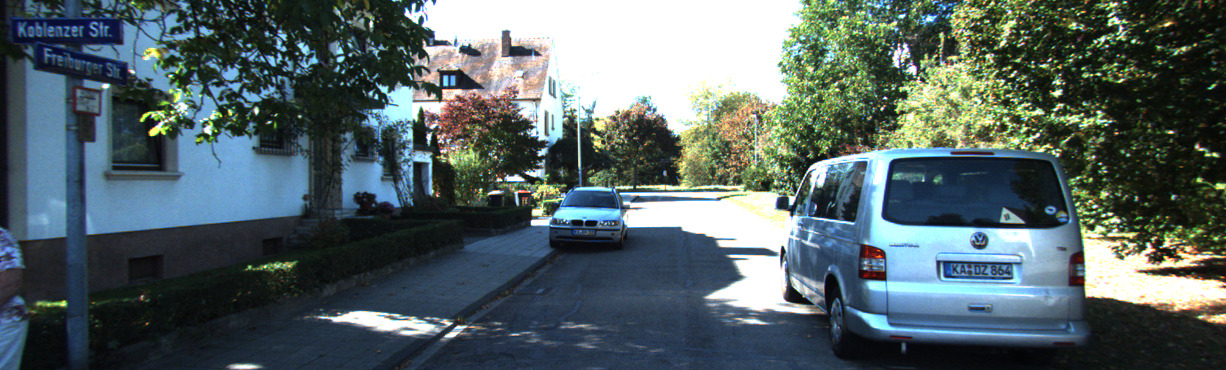}
\end{subfigure}\hfill
\begin{subfigure}{.332\linewidth}
  \centering
  \includegraphics[trim={150 0 150 125},clip,width=\linewidth]{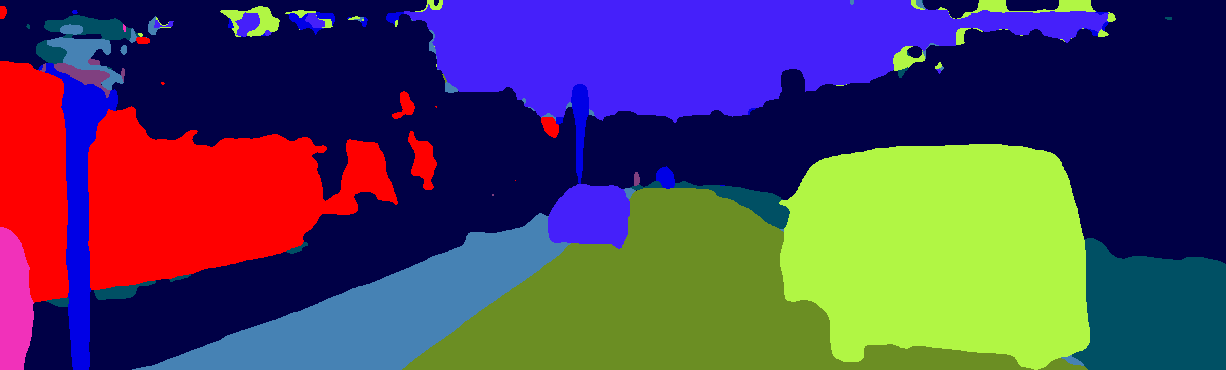}
\end{subfigure}\hfill
\begin{subfigure}{.332\linewidth}
  \centering
  \includegraphics[trim={150 0 150 125},clip,width=\linewidth]{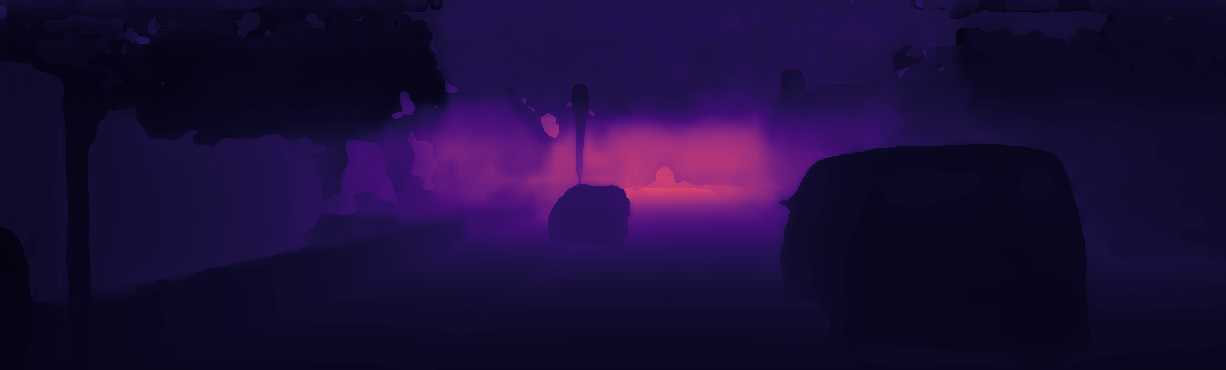}
\end{subfigure}\\

\begin{subfigure}{.332\linewidth}
  \centering
  \includegraphics[trim={150 0 150 125},clip,width=\linewidth]{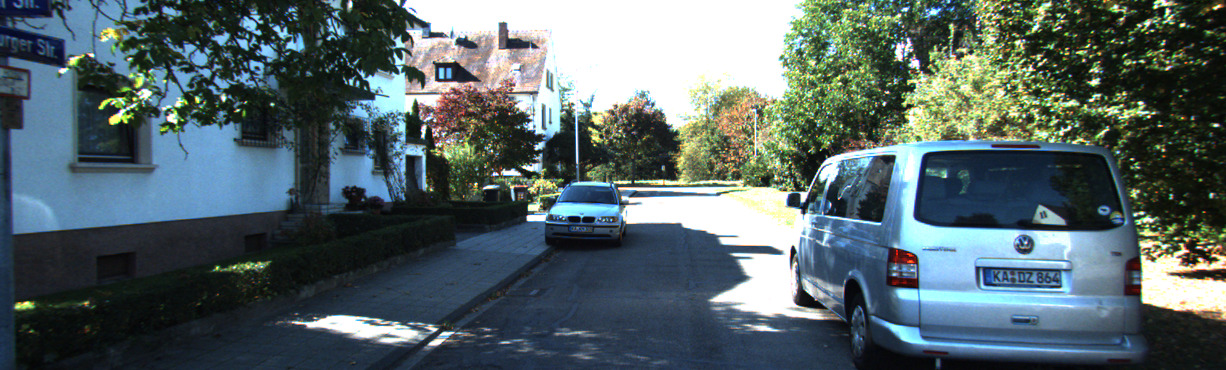}
\end{subfigure}\hfill
\begin{subfigure}{.332\linewidth}
  \centering
  \includegraphics[trim={150 0 150 125},clip,width=\linewidth]{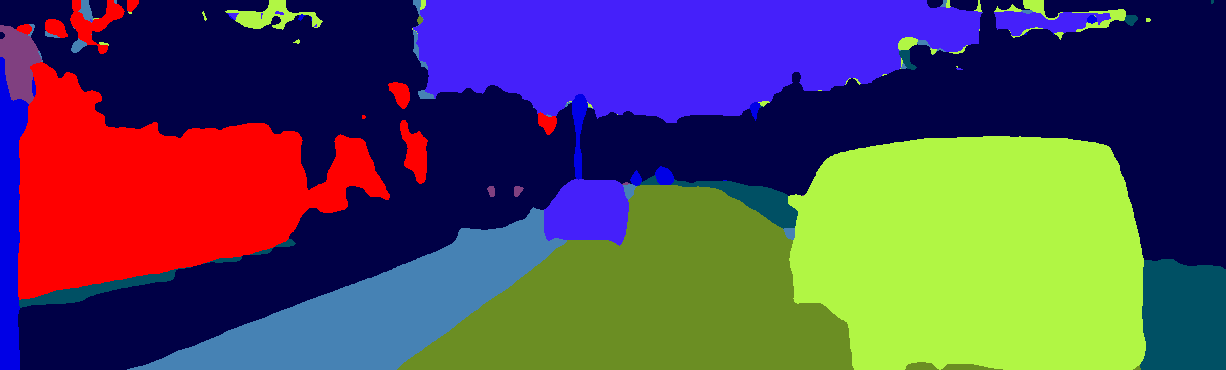}
\end{subfigure}\hfill
\begin{subfigure}{.332\linewidth}
  \centering
  \includegraphics[trim={150 0 150 125},clip,width=\linewidth]{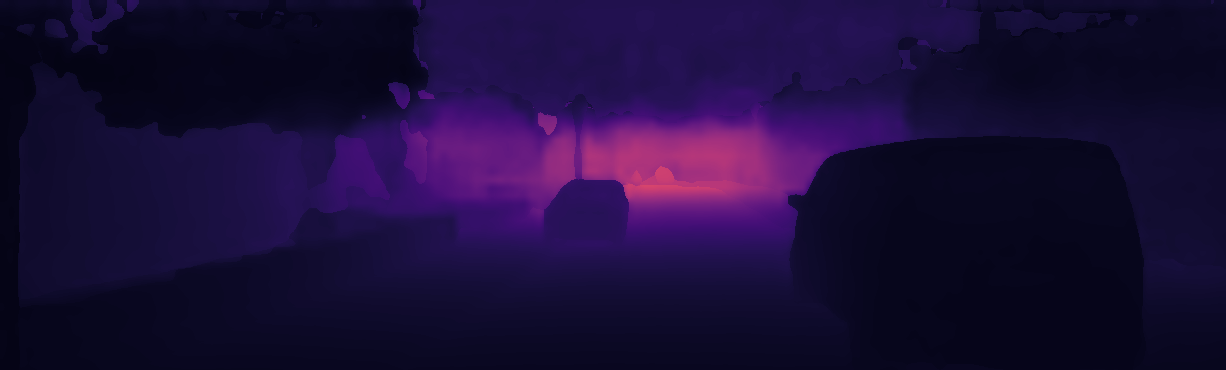}
\end{subfigure}\\

\caption{Prediction visualizations on SemKITTI-DVPS. From left to right: input images, temporally consistent panoptic segmentation (TCPS), and depth predictions. Color change of the same instance of TCPS indicates an id switch.}
\label{fig:sk_1}
\end{figure*}

We show some of the visualization results from the Cityscapes-DVPS and SemKITTI-DVPS datasets along with PolyphonicFormer (Swin-B backbone) predictions in Figure~\ref{fig:cs_1} and Figure~\ref{fig:sk_1}.

%
%
\bibliographystyle{splncs04}
\bibliography{egbib}

\end{document}